\documentclass[lettersize,journal]{IEEEtran}
\usepackage{pifont} % 引入 pifont 包

\usepackage{amsmath,amsfonts}
\usepackage{algorithmic}
\usepackage{algorithm}
\usepackage{array}
\usepackage[caption=false,font=normalsize,labelfont=sf,textfont=sf]{subfig}
\usepackage{textcomp}
\usepackage{stfloats}
\usepackage{bm}
\usepackage{url}
\usepackage{verbatim}
\usepackage{graphicx}
\usepackage{cite}
\usepackage{hyperref}       % hyperlinks
\usepackage{booktabs}       % professional-quality table
\usepackage{nicefrac}       % compact symbols for 1/2, etc.
\usepackage{microtype}      % microtypography
\usepackage{xcolor}         % colors
\usepackage{multirow}
\usepackage{caption}
\usepackage{epsfig}
\usepackage{amsmath}
\usepackage{amssymb}
\usepackage{amsthm}
\usepackage{subfloat}
\usepackage{times}
\usepackage{tikz} 
\usepackage{algorithm}
\usepackage{algorithmic}

\usepackage{pgfplots}
\pgfplotsset{compat=1.17}
\usepackage{makecell}
\usepackage{threeparttable}
\newtheorem{lemma}{Lemma}
\newtheorem{remark}{Remark}
\newtheorem{definition}{Definition}
\newtheorem{theorem}{Theorem}%[section]

\usepackage{xcolor}
\usepackage{colortbl}
\begin{document}

\title{Tensor Decomposition Structure Search Framework from an Interaction Perspective}
\author{ Ting-Wei Zhou, Xi-Le Zhao*, Sheng Liu, Wei-Hao Wu, Yu-Bang Zheng, Deyu Meng \IEEEmembership{}

        % <-this % stops a space
\thanks{*Corresponding author: Xi-Le Zhao.}
\thanks{Ting-Wei Zhou, Xi-Le Zhao, Sheng Liu, and Wei-Hao Wu are with the School of Mathematical Sciences/Research Center for Image and Vision Computing, University of Electronic Science and Technology of China, Chengdu, Sichuan 611731, China (e-mail: zhoutingwei2021@163.com; xlzhao122003@163.com; liusheng16@163.com; weihaowu99@163.com).}
\thanks{ Yu-Bang Zheng is with the School of Information Science and Technology, Southwest Jiaotong University, Chengdu 611756, China (e-mail: zhengyubang@163.com)}
\thanks{Deyu Meng is with the School of Mathematics and Statistics and Ministry	of Education Key Lab of Intelligent Networks and Network Security, Xi’an	Jiaotong University, Xi’an, Shaanxi 710049, China, and also with Pazhou	Laboratory (Huangpu), Guangzhou, Guangdong 510335, China (e-mail: dymeng@mail.xjtu.edu.cn).}}
\markboth{}%
{Shell \MakeLowercase{\textit{et al.}}: A Sample Article Using IEEEtran.cls for IEEE Journals}

\IEEEpubid{}

\maketitle

\begin{abstract}
Recently, tensor decompositions have attracted increasing attention. Fundamentally, different interactions among factors induce distinct tensor decomposition structures (i.e., tensor decomposition). Identifying an appropriate interaction-induced tensor decomposition structure for given data is a fundamental yet challenging problem in tensor modeling, and remains largely under-explored. Existing tensor decomposition structure search methods are typically restricted to a predefined interaction family, such as tensor contraction. To address this problem, we suggest a tensor decomposition structure search framework (I-TSS) from an interaction perspective that can identify either a single structure beyond a predefined interaction family or a mixture of structures involving heterogeneous interaction families. Specifically, we first systematically review existing tensor decomposition structures from an interaction perspective and construct a heterogeneous interaction-induced candidate structure set. Based on it, we propose a unified energy-based rank estimation scheme for heterogeneous interaction-induced candidate structure set. We then introduce the top-$k$ gating mechanism with learnable gating scores to dynamically select a suitable single-candidate structure or combine suitable multi-candidate structures, thereby identifying the tensor decomposition structure in a data-adaptive manner. Theoretically, we derive an approximation error bound for I-TSS, thereby establishing its approximation capability. Extensive experiments on synthetic and real-world datasets demonstrate that I-TSS consistently outperforms state-of-the-art tensor decomposition methods.

\end{abstract}

\begin{IEEEkeywords}
Tensor decomposition, tensor decomposition structure search, multi-dimensional data recovery.
\end{IEEEkeywords}

\section{Introduction}
Recent technological advancements have led to a rapid surge in diverse types of multi-dimensional data \cite{9791317,10556743,10026252,11386834,11020636} are increasingly emerging. Mathematically, tensor decomposition is well-suited for representing multi-dimensional data. Due to its capability to preserve the intrinsic multi-dimensional correlations of data, tensor decomposition has garnered substantial attention across multiple domains, including data analysis \cite{Han023, bay, Zhao2014BayesianCF}, machine learning \cite{11083646,11231345}, and computer vision \cite{11203237,9007523}.

Tensor decomposition, which decomposes the target tensor into a series of factors (e.g., matrix and tensor) with the interaction (i.e., algebraic operation) between them, reveals the intrinsic information behind multi-dimensional
data. Different interactions, including outer product, mode-$n$ product, tensor contraction, and tensor-tensor product (t-product) et al. \cite{cp,tucker,3524938,doi:10.1137/110837711}, can induce different tensor decomposition structures (i.e, tensor decompositions).  
The most classical tensor decomposition methods are outer product-induced CANDECOMP/PARAFAC (CP) decomposition \cite{cp} and mode-$n$ product-induced Tucker decomposition \cite{tucker}. It has been demonstrated that CP/Tucker models have been widely adopted to enhance the efficiency of modern algorithms \cite{Zhao2014BayesianCF, 7891546, dtt, 7902201} due to their powerful feature extraction and compression capabilities. A separate category of tensor decomposition is tensor contraction-induced tensor network decomposition \cite{3524938}, which encompasses tensor train (TT) decomposition \cite{doi:10.1137/090752286}, tensor ring (TR) decomposition \cite{Zhao2016TensorRD}, and fully-connected tensor network (FCTN) decomposition \cite{fctn}. These more sophisticated tensor network decompositions have been empirically validated to exhibit robust and generalized capabilities for modeling multi-dimensional data \cite{11184870, 7859390, 8237869, 10530915}.
\begin{table}[!t]\footnotesize
	\centering

	%	\caption{Quantitative comparison of the proposed DTR
		%		with different number of parameters.}
	%	\setlength{\tabcolsep}{0.4mm}{
		%		\begin{tabular}{ccccccc}
			%			\toprule
			%			\multicolumn{2}{c}{Characteristic\textbackslash Method}&\multicolumn{1}{c}{Tucker}&\multicolumn{1}{ c }{FCTN}&\multicolumn{1}{ c }{T-SVD}&\multicolumn{1}{ c }{SVDinsTN}&\multicolumn{1}{c}{I-TSS}\\
			%			\midrule
			%			\multicolumn{1}{c}{\multirow{3}{*}{Search ability}}&Tucker Family (CP, Tucker)&\cmark&&&&\cmark\\
			%			&FCTN Family (TT, TR, FCTN)&&\cmark&&\cmark&\cmark \\
			%			&T-SVD&&&\cmark&&\cmark \\
			%			\midrule
			%			\multicolumn{2}{c}{Data-Adaptability}&&&&\cmark&\cmark\\
			%			\bottomrule
			%	\end{tabular}}\label{resultnp}
	\caption{Comparison between classic tensor decomposition structure search methods and our I-TSS in terms of search capability and data adaptability.}
	\setlength{\tabcolsep}{0.5mm}{
		\begin{tabular}{ccccccccc}
			
			\toprule
			\multicolumn{1}{c}{Characteristic}&\multicolumn{7}{c}{\textbf{Search Capability}}&\multicolumn{1}{c}{\multirow{2.5}{*}{\textbf{Data Adaptability}}}\\
			\cmidrule{1-8}
			
			\multicolumn{1}{c}{Method} &CP & Tucker & TT & TR & FCTN & T-SVD & Mixture&\\
			\midrule
			Tucker & \multicolumn{1}{c}{\checkmark}&\multicolumn{1}{c}{\checkmark} & & & &&&\\
			FCTN &  &  & \multicolumn{1}{c}{\checkmark} &\multicolumn{1}{c}{\checkmark}  &\multicolumn{1}{c}{\checkmark} &&&\\
			T-SVD & &  &  &  &  &\multicolumn{1}{c}{\checkmark} &&\\
			SVDinsTN  &&  & \multicolumn{1}{c}{\checkmark} &\multicolumn{1}{c}{\checkmark}  &\multicolumn{1}{c}{\checkmark} &&&\checkmark\\
			I-TSS  &\multicolumn{1}{c}{\checkmark}& \multicolumn{1}{c}{\checkmark} &\multicolumn{1}{c}{\checkmark} & \multicolumn{1}{c}{\checkmark} & \multicolumn{1}{c}{\checkmark}&\multicolumn{1}{c}{\checkmark}&\multicolumn{1}{c}{\checkmark}&\checkmark\\
			%	\midrule
			%	\multicolumn{2}{c}{\textbf{Data-Adaptability}} &  &  &  &\multicolumn{1}{c}{\checkmark}  &\multicolumn{1}{c}{\checkmark} \\
			\bottomrule
	\end{tabular}}
	\label{mm}
\end{table}Most recently, t-product-induced tensor singular value decomposition (T-SVD) \cite{doi:10.1137/110837711}  has gained significant attention, mainly due to its close theoretical connection to the matrix singular value decomposition (SVD); see for example \cite{8740980, 9369083, 9064895, 10665981}. However, for given data, how to identify an appropriate interaction-induced tensor decomposition structure is a fundamental yet challenging problem in tensor modeling, and remains largely under-explored. 

Tensor decomposition structure search aims to discover a suitable tensor decomposition structure for compactly representing given data. The choice of tensor decomposition structure exerts a profound impact on its performance in practical applications.
Recently, several pioneer efforts have been proposed \cite{3618408, Hashemizadeh, 3524938, pmlr-v162-li22y, pmlr-v202-li23ar, cis, 10.1109, NieWT21,9321501}. Most of these approaches adopt a ``sampling-evaluation'' framework, which necessitates sampling a large pool of candidate structures and performing numerous repeated structure evaluations. Apart from heuristic approaches, the regularity-based method has also been explored for structure optimization \cite{10655696}. 
Despite these great efforts, these methods are typically restricted to a predefined interaction family, such as tensor contraction; as shown in Table~\ref{mm}.

	\begin{figure*}[!t]
	\centering
	\includegraphics[width=\hsize]{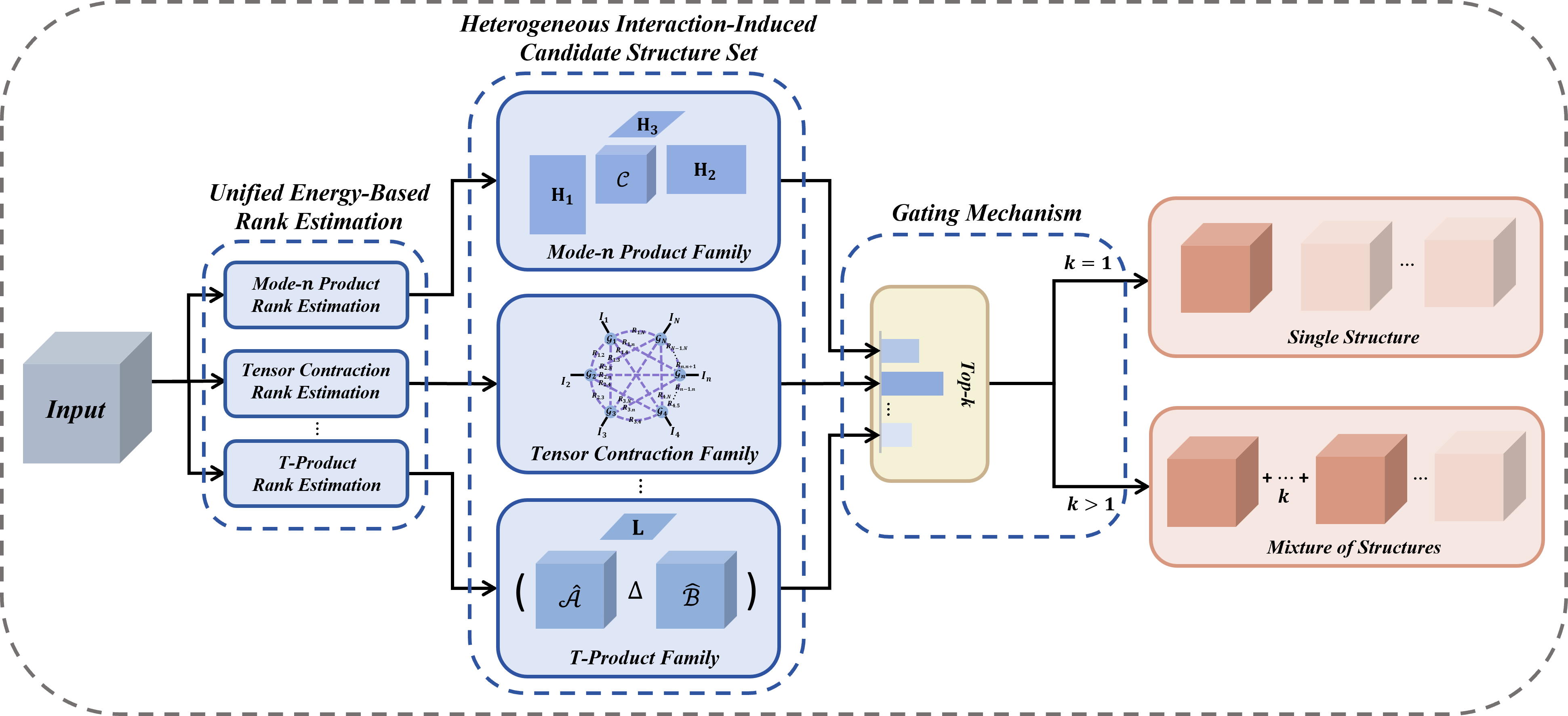}	
	
	\caption{Diagram of our I-TSS. First, we perform the energy-based rank estimation scheme for the input data. Then, the estimated ranks are fed into the heterogeneous interaction-induced candidate structure set. Finally, the top-$k$ gating mechanism is introduced to dynamically select a suitable single-candidate structure or combine suitable multi-candidate structures, leading to either a single structure ($k=1$) or mixture of structures ($k>1$).
	}
	\label{fig2}
\end{figure*}

To address these issues, we suggest tensor decomposition structure search framework (I-TSS) from an interaction perspective; see Fig.~\ref{fig2}. Specifically, we first systematically review existing tensor decomposition structures from an interaction perspective and construct a heterogeneous interaction-induced candidate structure set. Based on it, we propose a unified energy-based rank estimation scheme for heterogeneous interaction-induced candidate structure set. We then introduce the top-$k$ gating mechanism with learnable gating scores to dynamically select a suitable single-candidate structure or combine suitable multi-candidate structures, thereby identifying the tensor decomposition structure in a data-adaptive manner. Theoretically, we derive an approximation error bound for I-TSS, thereby establishing its approximation capability. 
In summary, the main contributions of this work are as follows:

\begin{itemize}
	\item To address the challenging and relatively under-explored problem of identifying an appropriate interaction-induced tensor decomposition structure for given data, we suggest a tensor decomposition structure search framework from an interaction perspective that can identify either a single structure beyond a predefined interaction family or a mixture of structures involving heterogeneous interaction families.
	\item To construct a heterogeneous interaction-induced candidate structure set, we first systematically review exiting tensor decomposition structures from an interaction perspective. Based on it, we suggest a unified energy-based rank estimation scheme for heterogeneous interaction-induced candidate structure set. 
	\item To identify the tensor decomposition structure in a data-adaptive manner, we then introduce the top-$k$ gating mechanism with learnable gating scores to dynamically select a suitable single-candidate structure or combine suitable multi-candidate structures. Theoretically, we derive an approximation error bound for I-TSS, thereby establishing its approximation capability.
	\item Extensive experiments on both synthetic datasets and real-world datasets consistently demonstrate that the proposed I-TSS outperforms the state-of-the-art tensor decomposition-based methods.

\end{itemize}

The remainder of this paper is organized as follows:
Section \ref{section:3} provides a comprehensive review of related work on tensor decomposition and tensor decomposition structure search. Section \ref{section:2} introduces the necessary notations and preliminary concepts for tensor decomposition.
Section \ref{section:4} details the proposed I-TSS and presents an unsupervised multi-dimensional
data recovery model.
Section \ref{section:6} reports the results of extensive experiments on synthetic and real-world datasets.
Section \ref{section:7} discusses the experimental findings and insights.
Finally, Section \ref{section:8} concludes this work and summarizes its main contributions.

	\section{Related work}
\label{section:3}
\subsection{Tensor Decomposition}
Most real-world data (e.g., multispectral images (MSIs), color videos, and light field data) inherently exhibit low-rank structures. Consequently, low-rank decomposition of matrices and tensors has become a widely studied topic in data processing. Unlike matrices, the rank definition of high-order tensors lacks a unified formulation. So far, many tensor ranks based on different interactions (e.g., outer product, mode-$n$ product, tensor contraction, and t-product) have been proposed.
The most classical tensor ranks include the outer product-based CP rank (defined as the minimum number of rank-one tensors required for decomposition) \cite{cp} and the mode-$n$ product-based Tucker rank (defined by the rank of tensor unfolding matrices) \cite{tucker}. It has been demonstrated that solving low-CP-rank or low-Tucker-rank optimization problems yields effective low-rank representation of multi-dimensional data \cite{Zhao2014BayesianCF, 7891546}. Moreover, low-Tucker/CP-rank models have been widely adopted to enhance the efficiency of modern deep learning algorithms \cite{dtt, 7902201} due to their powerful feature extraction and compression capabilities. 
Another category of tensor contraction-based tensor ranks is derived from tensor network decomposition \cite{3524938}, such as TT rank \cite{doi:10.1137/090752286}, TR rank \cite{Zhao2016TensorRD}, and FCTN rank \cite{fctn}. These relatively complex rank-based decompositions have powerful representation capability for multi-dimensional data modeling \cite{11184870, 7859390, 8237869, 10530915}.
More recently, the t-product-based tensor tubal-rank \cite{doi:10.1137/110837711}, which is rooted in T-SVD, has garnered significant attention, primarily due to its close connection to the matrix SVD definition. Additionally, the effectiveness of tubal-rank minimization has been validated across various signal processing tasks \cite{8740980, 9369083, 9064895, 10665981}. 

However, for given data, how to identify an appropriate interaction-induced tensor decomposition structure is a fundamental yet challenging problem in tensor modeling, and remains largely under-explored.
\subsection{Tensor Decomposition Structure Search}
Tensor decomposition structure search aims to discover a suitable tensor decomposition structure for compactly representing the given data.
Pioneer approaches \cite{3618408, Hashemizadeh, 3524938, pmlr-v162-li22y, pmlr-v202-li23ar, cis, 10.1109, NieWT21,9321501} adopt a ``sampling-evaluation'' framework, which relies on heuristic search algorithms to sample candidate structures, then evaluate each candidate structure individually. For instance, Hashemizadeh \emph{et al.} \cite{Hashemizadeh} proposed a greedy algorithm-based method that starts from a rank-one tensor and sequentially identifies the most promising tensor network edges to incrementally increase rank. Li \emph{et al.} \cite{3524938} developed a genetic meta-algorithm to search for optimal structure in Hamming space; they also proposed a local enumeration algorithm \cite{pmlr-v202-li23ar} to alternately update each structure-related variable. Apart from heuristic approaches, the regularity-based method has also been explored for structure search. Zheng \emph{et al.} \cite{10655696} proposed the SVD-inspired tensor network decomposition (SVDinsTN) to address tensor network structure search via regularized modeling, which greatly reduces computational costs. 

Despite these great efforts, these methods are typically restricted to a predefined interaction family, such as tensor contraction.
%\subsection{Mixture of Experts} 
%Mixture of experts (MoE) \cite{6797059, 716791} is an ensemble learning paradigm that trains specialized ``expert'' models to handle subtasks of a broader predictive modeling problem. The MoE framework operates in four core steps: (i) decompose the target task into distinct subtasks; (ii) develop a dedicated expert model for each subtask; (iii) employ a gating mechanism to determine the relevance of each expert to individual input samples; and (iv) aggregate expert outputs weighted by the gating signals to generate the final prediction. 
%
%Based on MoE framework and domain knowledge of tensor decomposition, we suggest a mixture of tensor experts (I-TSS) framework, which allows us to dynamically select and activate suitable tensor decompositions for the given data. This framework not only allows to select suitable tensor decomposition, but also supports to discover appropriate combination of tensor decompositions, which beyond the state-of-the-art tensor decomposition search methods.

\section{Notations and preliminaries}
\label{section:2}

\subsection{Notations}
Scalars, vectors, matrices, and tensors are denoted by $x$, $\bm{x}$, $\mathbf{X}$, and $\mathcal{X}$, respectively. For a vector $\bm{x} \in \mathbb{R}^{I}$, we use $\bm{x}(n) \in \mathbb{R}$ to denote the $n$-th element of $\bm{x}$. Meanwhile, for an $N$th-order tensor $\mathcal{X} \in \mathbb{R}^{I_{1} \times I_{2}\cdots \times I_{N}}$, we use $\mathcal{X}{(i_1,i_2,\cdots,i_N)}$ to denote the $(i_1,i_2,\cdots,i_N)$-th element of $\mathcal{X}$. For a third-order tensor $\mathcal{X} \in \mathbb{R}^{I_{1} \times I_{2} \times I_{3}}$, we use $\mathbf{X}^{(n)} \in \mathbb{R}^{I_{1} \times I_{2}}$ to denote the $n$-th frontal slice of $\mathbf{X}$. The symbol $\odot$ denotes the Hadamard product (i.e., point-wise multiplication). The symbol $\triangle$ denotes the face-wise product between two third-order tensors \cite{doi:10.1137/110837711}. The rank of matrix $\mathbf{X}$ is denoted as $\mathrm{rank}(\mathbf{X})$. The Frobenius norm of tensor $\mathcal{X}$ is defined as $\|\mathcal{X}\|_F=\sqrt{\sum_{i_1,i_2,\cdots,i_N}\left|\mathcal{X}(i_1,i_2,\cdots,i_N)\right|^2}$.   

\subsection{Preliminaries} 
%\begin{definition}[Unfolding Matrix] For an Nth-order tensor $\mathcal{X}\in \mathbb{R}^{I_1\times I_2\times \cdots \times I_N}$, its single-mode unfolding matrix is $\mathcal{X}_{<k>} \in \mathbb{R}^{I_k \times \prod_{m=1}^{N, \ne k} I_m}$ and its double-mode unfolding matrix is $\mathcal{X}_{<k_1k_2>} \in \mathbb{R}^{I_{k_1}I_{k_2} \times \prod_{m=1}^{N,\ne k_1\ne k_2} I_m}$.
%	\label{unfold}
%\end{definition}

\begin{definition}[Generalized Unfolding \cite{fctn}] Given a permutation $\mathbf{m}$ of the vector $(1, 2, \cdots ,N)$, the generalized unfolding of tensor $\mathcal{X}\in \mathbb{R}^{I_1\times I_2\times \cdots \times I_N}$ corresponding to vector $\mathbf{m}$ is defined as the matrix
\begin{equation}
		\mathbf{X}_{[\mathbf{m}(1:q);\mathbf{m}(q+1:N)]}=\mathrm{reshape}\footnote{By following the notation of MATLAB commands.}\left({\mathcal{X}}^{\mathbf{m}},\prod_{i=1}^{q}I_{\mathbf{m}(i)},\prod_{i=q+1}^{N}I_{\mathbf{m}(i)}\right),
\end{equation}
	where $\mathcal{X}^{\mathbf{m}}$ is formed by permuting the modes of tensor $\mathcal{X}$ according to the order of vector $\mathbf{m}$.
	
	For example, the traditional mode-$n$ unfolding of $\mathcal{X}$ is $\mathbf{X}_{[n;1,2,\cdots,n-1,n+1,\cdots,N]}$, which is denoted by $\mathbf{X}_{<n>}$. 
%	Similarly, mode-$n_1n_2$ unfolding of $\mathcal{X}$ is $\mathbf{X}_{[n_1,n_2;1,2,\cdots,n_1-1,n_1+1,\cdots,n_2-1,n_2+1,\cdots,N]}$, which is denoted by $\mathbf{X}_{<n_1n_2>}$.
	\label{unfold}
\end{definition}
\subsubsection{Interaction}
%$\mathbf{Factors\ Interaction}$
Interaction serves as a key component of tensor decomposition structure, with different categories of it inducing distinct tensor decomposition structures. We will first introduce different interactions, including outer product, mode-$n$ product, tensor contraction, and t-product.
\begin{definition}[Outer Product \cite{sr}] The outer product between vectors $\bm{a}_n \in \mathbb{R}^{I_n} (n=1,2,\cdots,N)$ results in an Nth-order tensor $\mathcal{X}\in \mathbb{R}^{I_1\times I_2\times \cdots \times I_N} $, defined as
	\begin{equation}
	\mathcal{X} = \bm{a}_1 \circ \bm{a}_2 \circ \cdots \circ \bm{a}_N.
	\end{equation}
		with elements as
		\begin{equation}
			\begin{aligned}
					\mathcal{X}{(i_{1},i_{2},\cdots,i_N)}=&\bm{a}_1(i_1)\bm{a}_2(i_2)\cdots\bm{a}_N(i_N).
				\end{aligned}
			\end{equation}
			
%Similarly, extending to high-order space, the outer product for vectors $\bm{a}_n \in \mathbb{R}^{I_n} (n=1,2,\cdots,N)$ will result in a high-order tensor $\mathcal{X}\in \mathbb{R}^{I_1\times I_2\times \cdots \times I_N} $ as follows:
%	\begin{equation}
%	\mathcal{X} = \bm{a}_1 \circ \bm{a}_2 \circ \cdots \circ \bm{a}_N.
%\end{equation}
\end{definition}
\begin{definition}[Mode-$n$ Product \cite{sr}] The mode-$n$ product between an Nth-order tensor $\mathcal{X}\!\in\mathbb{R}^{ I_1\times I_2\times \cdots I_{n-1}\times I_n\times I_{n+1} \times \cdots \times I_N}$ and a matrix $\mathbf{A} \in \mathbb{R}^{J \times I_n}$ results in an Nth-order tensor $\mathcal{Y} \in \mathbb{R}^{I_1\times I_2\times \cdots I_{n-1}\times J\times I_{n+1} \times \cdots \times I_N} $, defined as
	\begin{equation}
		\begin{aligned}
			&\left(\mathcal{X} \times_n \mathbf{A}\right)(i_1, i_2, \cdots, j, \cdots, i_N)\\
			&=\sum_{i_n=1}^{I_n} \mathcal{X}(i_1, i_2, \cdots, i_n, \cdots, i_N) \mathbf{A}(j, i_n).
		\end{aligned}
	\end{equation}
\end{definition}

According to the above-mentioned two definitions, we have
\begin{equation}
	\mathcal{Y}=\mathcal{X} \times_n \mathbf{A} \Leftrightarrow \mathbf{Y}_{<n>}=\mathbf{A} \mathbf{X}_{<n>}.
\end{equation}

\begin{definition}[Tensor Contraction \cite{fctn}] Let $\mathcal{A}\in \mathbb{R}^{J_1\times J_2\times \cdots \times J_M}$ and $\mathcal{B}\in \mathbb{R}^{I_1\times I_2\times \cdots \times I_N}$ share the same
size in $p$ modes (i.e., $J_{\mathbf{m}(q)} = I_{\mathbf{n}(q)}$ for $q = 1$, $2$, $\cdots$, $p$). The tensor contraction between $\mathcal{A}$ and $\mathcal{B}$ results in an $(M +N - 2p)$th-order tensor $\mathcal{X}$, defined as $\mathcal{X}=\mathcal{A}\times_{\mathbf{m}({1}:p)}^{\mathbf{n}({1}:p)}\mathcal{B}\Leftrightarrow$ $\mathbf{X}_{[1:M-p;M-p+1:M+N-2p]}=\mathbf{A}_{[\mathbf{m}({p+1:M});\mathbf{m}({1:p})]}\mathbf{B}_{[\mathbf{n}({1:p});\mathbf{n}({p+1:N})]}$.
\end{definition}

\begin{definition}
	[T-Product \cite{34234014}] The invertible transform-based t-product
	between $\mathcal{A} \in \mathbb{R}^{I_{1} \times I_{2} \times I_{3}}$ and $\mathcal{B} \in \mathbb{R}^{I_{2} \times I_{4} \times I_{3}}$ is defined as
	$\mathcal{A} *_M \mathcal{B}=((\mathcal{A}\times_3 \mathbf{M})\triangle(\mathcal{B}\times_3\mathbf{M}))\times_3 \mathbf{M}^{\mathtt{-1}} \in \mathbb{R}^{I_{1} \times I_{4} \times I_{3}}$, where
	$\mathbf{M} \in \mathbb{R}^{I_{3} \times I_{3}}$ is the invertible matrix, and $\mathbf{M}^{\mathtt{-1}}$ is the inverse matrix of $\mathbf{M}$.
	\label{tp}
\end{definition}

\subsubsection{Tensor Decomposition} Based on the above interactions, different tensor decomposition structures are induced as follows.
\begin{definition}[CP Decomposition \cite{cp}] For an Nth-order tensor $\mathcal{X}\in$
	$\mathbb{R}^{I_1\times I_2\times \cdots \times I_N}$, the CP decomposition is defined as
	
	\begin{equation}
		\mathcal{X}=\sum_{r=1}^{R}\bm{h}_r^{1}\circ\bm{h}_r^{2}\circ\cdots\circ\bm{h}_r^{N},
	\end{equation}
%	with elements as
%	\begin{equation}
%	\begin{aligned}
%		\mathcal{X}{(i_{1},i_{2},\cdots,i_{N})}=&\sum_{r=1}^{R}\bm{h}_r^{1}(i_1)\bm{h}_r^{2}(i_2)\cdots\bm{h}_r^{N}(i_N),
%	\end{aligned}
%	\end{equation}
	where $\mathbf{H}_{n}=[\bm{h}_1^{n}, \bm{h}_2^{n}, \cdots, \bm{h}_R^{n}]\in \mathbb{R}^{I_{n}\times R} ( n=1,2,\cdots,N)$ is the factor matrix. The CP rank is defined as $R$. 
	\label{CP}
\end{definition}

\begin{definition}[Tucker Decomposition \cite{tucker}] For an Nth-order tensor $\mathcal{X}\in$
	$\mathbb{R}^{I_1\times I_2\times \cdots \times I_N}$, the Tucker decomposition is defined as
	
	\begin{equation}
		\mathcal{X}=\mathcal{C}\times_{1}\mathbf{H}_{1}\times_2 \mathbf{H}_{2}\cdots\times_N \mathbf{H}_{N},
	\end{equation}
%		with elements as
%	\begin{equation}
%		\begin{aligned}
%			\mathcal{X}{(i_{1},i_{2},\cdots,i_{N})}=&\sum_{r_1=1}^{R_1}\sum_{r_2=1}^{R_2}\cdots\sum_{r_N=1}^{R_N}\mathcal{C}(r_1,r_2,\cdots, r_N)\\& \mathbf{H}_{1}(i_1,r_1)\mathbf{H}_{2}(i_2,r_2)\cdots\mathbf{H}_{N}(i_N,r_N),
%		\end{aligned}
%	\end{equation}
	where $\mathbf{H}_{n}\in \mathbb{R}^{I_{n}\times R_{n}}( n=1,2,\cdots,N)$ is the factor matrix and $\mathcal{C}\in\mathbb{R}^{R_{1}\times R_{2}\times\cdots\times R_{N}}$ is the Nth-order core tensor. The Tucker rank is defined as the vector $[R_1,R_2,\cdots,R_N]$. 
	\label{Tucker}
\end{definition}

\begin{definition}[TT Decomposition \cite{doi:10.1137/090752286,ZHENG2025107808}] For an Nth-order tensor $\mathcal{X}\in$
	$\mathbb{R}^{I_1\times I_2\times \cdots \times I_N}$, the TT decomposition is defined as
	\begin{equation}
%	\mathcal{X}=\sum_{r_1=1}^{R_1}\cdots\sum_{r_N=1}^{R_N}\mathcal{G}_{1}(r_1,:,r_2)\circ\cdots\circ\mathcal{G}_{N}(r_N,:,r_1),
	\mathcal{X}=\mathbf{G}_{1}\times_{2}^{1}\mathcal{G}_{2}\cdots\times_{N}^{1}\mathbf{G}_{N},
	\end{equation}
%	with elements as
%	\begin{equation}
%		\begin{aligned}
%			\mathcal{X}{(i_{1},i_{2},\cdots,i_{N})}=&\sum_{r_1=1}^{R_1}\sum_{r_2=1}^{R_2}\cdots\sum_{r_{N-1}=1}^{R_{N-1}} \\&\mathbf{G}_{1}(i_{1},r_{1}) \mathcal{G}_{2}(r_{1},i_{2},r_{2})\cdots\\&\mathbf{G}_N(r_{N-1},i_{N}),
%		\end{aligned}
%	\end{equation}
	where $\mathcal{G}_{n}\in\mathbb{R}^{R_{n-1}\times I_{n}\times R_{n}}( n=2,\cdots,N-1)$ is the third-order factor tensor. $\mathbf{G}_{1}\in\mathbb{R}^{I_{1}\times R_{1}}$ and $\mathbf{G}_{N}\in\mathbb{R}^{R_{N-1}\times I_{N}}$ are the factor matrices. The TT rank is defined as the vector $[R_1,R_2,\cdots,R_{N-1}]$. 
	\label{TT}
\end{definition}

\begin{definition}[TR Decomposition \cite{Zhao2016TensorRD, ZHENG2025107808}] For an Nth-order tensor $\mathcal{X}\in$
	$\mathbb{R}^{I_1\times I_2\times \cdots \times I_N}$, the TR decomposition is defined as
		\begin{equation}
		\mathcal{X}=\mathcal{G}_{1}\times_{3}^{1}\mathcal{G}_{2}\cdots\times_{n+1}^{1}\mathcal{G}_{n}\cdots\times_{N+1,1}^{1,3}\mathcal{G}_{N},
	\end{equation}
%	with elements as
%	\begin{equation}
%		\begin{aligned}
%			\mathcal{X}{(i_{1},i_{2},\cdots,i_{N})}=&\sum_{r_1=1}^{R_1}\sum_{r_2=1}^{R_2}\cdots\sum_{r_{N}=1}^{R_{N}} \\&\mathcal{G}_{1}(r_{1},i_{1},r_{2}) \mathcal{G}_{2}(r_{2},i_{2},r_{3})\cdots\\&\mathcal{G}_N(r_{N},i_{N},r_{1}),
%		\end{aligned}
%	\end{equation}
	where $\mathcal{G}_{n}\in\mathbb{R}^{R_{n}\times I_{n}\times R_{n+1}}( n=1,2,\cdots,N;R_1=R_{N+1})$ is the third-order factor tensor. The TR rank is defined as the vector $[R_1,R_2,\cdots,R_N]$. 
	\label{TR}
\end{definition}

\begin{definition}[FCTN Decomposition \cite{fctn,ZHENG2025107808}]For an Nth-order tensor $\mathcal{X}\in$
	$\mathbb{R}^{I_1\times I_2\times \cdots \times I_N}$, the FCTN decomposition is defined as
	\begin{equation}
		\mathcal{X}=\mathcal{G}_{1}\times_{b^1_1}^{a_1^1}\mathcal{G}_{2}\cdots\times_{b_{1:N-1}^{N-1}}^{a_{1:N-1}^{N-1}}\mathcal{G}_{N},
	\end{equation}
%	with elements as
%\begin{equation}
%\begin{aligned}
%	& \mathcal{X}(i_{1},i_{2},\cdots,i_{N})= \\
%	& 
%	\begin{aligned}
%		\sum_{r_{1,2}=1}^{R_{1,2}}\sum_{r_{1,3}=1}^{R_{1,3}}\cdots\sum_{r_{1,N}=1}^{R_{1,N}}\sum_{r_{2,3}=1}^{R_{2,3}}\cdots\sum_{r_{2,N}=1}^{R_{2,N}}\cdots\sum_{r_{N-1,N}=1}^{R_{N-1,N}}
%	\end{aligned} \\
%	  &\mathcal{G}_{1}(i_{1},r_{1,2},r_{1,3},\cdots,r_{1,N}) \\
%	& \mathcal{G}_{2}(r_{1,2},i_{2},r_{2,3},\cdots,r_{2,N})\cdots \\
%	& \mathcal{G}_{N}(r_{1,N},r_{2,N},\cdots,r_{N-1,N},i_{N}),
%\end{aligned}
%\end{equation}
where $\mathcal{G}_{n}\in\mathbb{R}^{R_{1,n}\times R_{2,n}\times \cdots \times R_{n-1,n} \times I_{n}\times R_{n,n+1} \times \cdots \times R_{n,N}}( n=1,2,\cdots,N)$ is the Nth-order factor tensor. The FCTN rank is defined as the vector collected by $R_{n_1,n_2}(1\le n_1 <n_2\le N\ and \ n_1, n_2 \in \mathbb{N}^+)$ and
\[
\begin{cases}
	a_i^j = i, \\
	b_i^j = 2+(i - 1)(N - j + 1),
\end{cases}
\]
for \(i = 1,2,\cdots,j\) and \(j = 1,2,\cdots,N - 1\).
\label{FCTN}
\end{definition}

%\begin{definition}[T-SVD \cite{34234014}] For a $3$rd-order tensor $\mathcal{X}\in$
%	$\mathbb{R}^{I_1\times I_2\times  I_3}$, the invertible transform-based T-SVD is defined as
%	
%	\begin{equation}
%		\mathcal{X}=\mathcal{U}*_M\mathcal{S}*_M\mathcal{V^{\mathtt{H}}},
%	\end{equation}
%	where $\mathcal{U}\in \mathbb{R}^{I_{1}\times I_1 \times I_3}$ and $\mathcal{V}\in \mathbb{R}^{I_{2}\times I_2 \times I_3}$ are orthogonal tensors, and  $\mathcal{S}\in \mathbb{R}^{I_{1}\times I_2 \times I_3}$is an f-diagonal tensor.
%	\label{t-svd}
%\end{definition}

%\begin{definition}[Tensor Tubal-Rank \cite{34234014}] The invertible transform-based tensor tubal-rank of $\mathcal{X} \in \mathbb{R}^{I_{1} \times I_{2} \times I_{3}}$ is defined as
%	\begin{equation}
%		\mathrm{rank}_t(\mathcal{X})\triangleq \max \limits_{k=1,2,\cdots,I_{3}}{\mathrm{rank}\left((\mathcal{X}\times_3 \mathbf{M})^{(k)}\right)},
%	\end{equation}
%	where $\mathbf{M} \in \mathbb{R}^{I_{3} \times I_{3}}$ is the invertible matrix.
%\end{definition}

\begin{definition}[T-Product-Induced  Tensor Factorization \cite{8066348}] 
	For a third-order tensor $\mathcal{X}\in$
	$\mathbb{R}^{I_1\times I_2\times I_3}$ with tubal-rank\footnote{The tubal-rank of $\mathcal{X} \in \mathbb{R}^{I_{1} \times I_{2} \times I_{3}}$ is defined as $\mathrm{rank}_t(\mathcal{X})\triangleq \max \limits_{n=1,2,\cdots,I_{3}}{\mathrm{rank}\left((\mathcal{X}\times_3 \mathbf{M})^{(n)}\right)}$, where $\mathbf{M} \in \mathbb{R}^{I_{3} \times I_{3}}$ is the invertible matrix.} up to $R$, the t-product-induced tensor factorization (TF) is defined as
	\begin{equation}
		\mathcal{X} =	\mathcal{A}*_M\mathcal{B},
	\end{equation}
where $\mathcal{A}\in \mathbb{R}^{I_{1} \times R \times I_{3}}$ and $\mathcal{B}\in \mathbb{R}^{R \times I_{2} \times I_{3}}$ are the third-order factor tensors.
%	Let $\mathcal{X}\in \mathbb{R}^{I_{1} \times I_{2} \times I_{3}}, \mathcal{Y}\in \mathbb{R}^{I_{1} \times I_{2} \times I_{3}}$, and $\mathcal{Z}\in\mathbb{R}^{I_{2} \times I_{4} \times I_{3}}$ be arbitrary tensors, then
%	
%	(i) If $\mathrm{rank}_t(\mathcal{X})=R$\footnote{The tubal-rank of $\mathcal{X} \in \mathbb{R}^{I_{1} \times I_{2} \times I_{3}}$ is defined as $
%		\mathrm{rank}_t(\mathcal{X})\triangleq \max \limits_{k=1,2,\cdots,I_{3}}{\mathrm{rank}\left((\mathcal{X}\times_3 \mathbf{M})^{(k)}\right)},$
%		where $\mathbf{M} \in \mathbb{R}^{I_{3} \times I_{3}}$ is the invertible matrix.}, then there exist two tensors $\mathcal{A}\in \mathbb{R}^{I_{1} \times R \times I_{3}}$ and $\mathcal{B}\in \mathbb{R}^{R \times I_{2} \times I_{3}}$ such that $\mathcal{X}= \mathcal{A}*_M\mathcal{B}$ holds and they meet $\mathrm{rank}_t(\mathcal{A})=\mathrm{rank}_t(\mathcal{B})=R$.
%	
%	(ii) $\mathrm{rank}_t(\mathcal{Y}*_M\mathcal{Z})\le min \{\mathrm{rank}_t(\mathcal{Y}), \mathrm{rank}_t(\mathcal{Z})\}$.
%
%	Based on Definition~\ref{tp}, we can factorize any tensor of a tensor tubal-rank up to $R$ into the tensor product form $\mathcal{X} =	\mathcal{A}*_M\mathcal{B}$. 
	For efficiency, instead of adopting invertible matrix $\mathbf{M} \in \mathbb{R}^{I_3\times I_3}$ and its inverse matrix $\mathbf{M}^{\mathtt{-1}}$ to obtain $\mathcal{X}$, we can cleverly start from the latent factor tensors $\hat{\mathcal{A}}_M \triangleq \mathcal{A}\times_3 \mathbf{M}$ and $\hat{\mathcal{B}}_M \triangleq \mathcal{B}\times_3 \mathbf{M}$ to reformulate the tensor factorization as 
	\begin{equation}
		\mathcal{X} =(\hat{\mathcal{A}}_M\triangle\hat{\mathcal{B}}_M)\times_3\mathbf{M}^{\mathtt{-1}}.
	\end{equation}
	Thus, we only need to learn factor tensors $\hat{\mathcal{A}}_M$ and $\hat{\mathcal{B}}_M$ to obtain $\mathcal{X}$, which avoids the complicated calculation of the $\mathcal{A}\times_3 \mathbf{M}$ and $\mathcal{B}\times_3 \mathbf{M}$ \textup{\cite{9878497}}. Furthermore, to enable flexible adaptation to different data, we can replace the invertible transform with a learnable linear transform, as formulated below:
	\begin{equation}
		\mathcal{X} =(\hat{\mathcal{A}}\triangle\hat{\mathcal{B}})\times_3\mathbf{L},
	\end{equation}
	where $\mathbf{L} \in \mathbb{R}^{\ell \times I_{3}}$ is a learnable linear transform, $\hat{\mathcal{A}}\in \mathbb{R}^{I_{1} \times R \times \ell}$ and $\hat{\mathcal{B}}\in \mathbb{R}^{R \times I_{2} \times \ell}$ are latent third-order factor tensors. 
	\label{t0}
\end{definition}

	%-----------------------------------------------------------------------

	%-------------------------------------------------------------------------
	\section{Tensor Decomposition Structure Search Framework}
	\label{section:4}
	In this section, we elaborate on the design details of the proposed I-TSS. Theoretically, we derive an approximation error bound for I-TSS, thereby establishing its approximation capability. To examine the potential of the proposed I-TSS, we evaluate it on the representative multi-dimensional data recovery task and suggest an unsupervised I-TSS-based multi-dimensional data recovery model.

	 \label{ere}
	\subsection{The Proposed I-TSS}
	Identifying an appropriate interaction-induced tensor decomposition structure for given data is a fundamental yet challenging problem in tensor modeling, and remains largely under-explored. Existing tensor decomposition structure search methods are typically restricted to a predefined interaction family (e.g., tensor contraction). To address this problem, we suggest a tensor decomposition structure search framework from an interaction perspective that can identify either a single structure beyond a predefined interaction family or a mixture of structures involving heterogeneous interaction families. Specifically, we first systematically review existing tensor decomposition structures from an interaction perspective and construct a heterogeneous interaction-induced candidate structure set. Based on it, we propose a unified energy-based rank estimation scheme for heterogeneous interaction-induced candidate structure set. We then introduce the top-$k$ gating mechanism with learnable gating scores to dynamically select a suitable single-candidate structure or combine suitable multi-candidate structures, thereby identifying the tensor decomposition structure in a data-adaptive manner.
	 
	 In summary, the proposed I-TSS is composed of three key components, i.e., the heterogeneous interaction-induced candidate structure set, the unified energy-based rank estimation, and the gating mechanism; as shown in Fig.~\ref{fig2}. The mathematical representation of I-TSS can be expressed as follows:
	 \begin{equation}
	 		\mathcal{\hat{X}} = \sum_{i=1}^L\lambda_{\theta_i}\mathcal{E}_{\phi_i},
	 \end{equation}
	 where $\mathcal{\hat{X}}$ is the data estimated by the proposed I-TSS. $\{\mathcal{E}_{\phi_i}\}_{i=1}^L$ represent heterogeneous interaction-induced candidate structures and $\{{\phi_i}\}_{i=1}^L$ denote the learnable parameters of heterogeneous interaction-induced candidate structures. $\{\lambda_{\theta_i}\}_{i=1}^L$ are the normalized gating scores and $\{{\theta_i}\}_{i=1}^L$ denote the learnable parameters of gating scores. $L$ denotes the number of heterogeneous interaction-induced candidate structures.
	 Now, we introduce the three components of the proposed I-TSS.
	
	\subsubsection{Heterogeneous Interaction-Induced Candidate Structure Set}
	\label{cis}
	To offer a heterogeneous interaction-induced candidate structure set, we first conduct a systematic review of existing tensor decomposition structures.	
	For tensor decomposition structure, the interaction between factors plays a crucial role. Hence, we reorganize the relationships between different tensor decomposition structures from a interaction perspective and divide them into three categories: the mode-$n$ product family (including CP and Tucker decompositions), the tensor contraction family (including TT, TR, and FCTN decompositions), and the t-product family (including TF).
	\begin{theorem} Outer product-induced CP decomposition can be viewed as a special case of mode-$n$ product-induced Tucker decomposition where the core tensor is superdiagonal and $R_1=R_2=\cdots=R_N$ \textup{\cite{sr}}.\\
		%		\footnote{An $N$th-order tensor $\mathcal{X}\in \mathbb{R}^{I_1\times I_2\times \cdots \times I_N}$ is superdiagonal if 
			%			$
			%					\mathcal{X}({i_{1},i_{2},\cdots, i_{N}})=\left\{{\begin{array}{l l}{1,}&{{\mathrm{~if~}}i_{1}=i_{2}=\cdots=i_{N}},\\ {0,}&{{\mathrm{~otherwise.}}}\end{array}}\right.$} 
		
		\label{r1}	
	\end{theorem}
%	\begin{theorem} Tensor contraction induced TR decomposition is equivalent to tensor contraction induced FCTN decomposition.\\
%		\label{r3}
%	\end{theorem}

	\begin{theorem} Tensor contraction-induced TT decomposition can be viewed as a special case of tensor contraction-induced TR decomposition where the TR ranks $R_1=1$ \cite{ZHENG2025107808}.\\
		\label{r2}
	\end{theorem}

	\begin{theorem} Tensor contraction-induced TT decomposition can be viewed as a special case of tensor contraction-induced FCTN decomposition where the FCTN ranks $R_{n_1,n_2}=1(|n_1-n_2|>1)$ \cite{ZHENG2025107808}.\\
		\label{r6}
	\end{theorem}
	
	%	\begin{remark} For a 3rd-order tensor, TR decomposition is equivalent to FCTN decomposition.\\
		%	\label{r4}
		%	\end{remark}
	
	\begin{theorem} For the third-order case, tensor contraction-induced TR decomposition is equivalent to tensor contraction-induced FCTN decomposition. For higher-order cases, tensor contraction-induced TR decomposition can be viewed as a special case of tensor contraction-induced FCTN decomposition where the FCTN ranks $R_{n_1,n_2}=1(|n_1-n_2|>1)$ except for $R_{1,N}$ \cite{ZHENG2025107808}.\\
		\label{r7}
	\end{theorem}
	
	\begin{lemma}
	 Let
		$\mathcal{F}_{\mathrm{1}},\mathcal{F}_{\mathrm{2}},\mathcal{F}_{\mathrm{3}}$ are mode-$n$ product, tensor contraction, t-product interaction-induced representative tensor decompositions, respectively. Then, $\mathcal{F}_{\mathrm{1}},\mathcal{F}_{\mathrm{2}},\mathcal{F}_{\mathrm{3}}$ are mutually incommensurable, i.e., for 
		$\mathcal{F}_i,\mathcal{F}_j$ ($1\le i\neq j \le 3$), $\mathcal{F}_{i} \not\subset \mathcal{F}_{j}$ and $\mathcal{F}_{j} \not\subset \mathcal{F}_{i}$.

	 The proof is provided in supplementary material.
	\label{le}
\end{lemma}
	Based on the above Theorems~\ref{r1}--\ref{r7} and Lemma \ref{le}, for the third-order case, we consider Tucker decomposition, FCTN decomposition, and TF as the heterogeneous interaction-induced candidate structure set. Notably, due to the standard t-product being only applicable to the third-order case. Therefore, for the higher-order case, we consider Tucker decomposition and FCTN decomposition as the heterogeneous interaction-induced candidate structure set.
	
	Based on this rationally structured heterogeneous interaction-induced candidate structure set, we detail how to estimate ranks for heterogeneous interaction-induced candidate structures in the following section.
	\subsubsection{Unified Energy-Based Rank Estimation}
	Rank is a critical component of tensor decomposition structure, as it accurately characterizes the intrinsic structure of data. However, the definition of tensor rank varies substantially across distinct interactions, so it is important to find a unified rank estimation scheme for different tensor decomposition structures. To achieve this goal, we proposed a unified energy-based scheme for	rank estimation across all heterogeneous interaction-induced candidate structures. Specifically, we define the accumulation energy ratio of top-$k$ singular values as $\sum_{i=1}^{k} \sigma_i^2/\sum_{j}\sigma_j^2$, where $\sigma_i$ is $i$-th singular value. Then, we decompose the input matrix into ``dominant components'' (main information) and ``secondary components'' (redundancy information) and use the energy ratio threshold $\tau$ to determine the boundary between them. To exemplify this, the pseudocode for the energy-based rank estimation algorithm is provided in the algorithm \ref{u}.
	\subsubsection{Gating Mechanism}
	To identify the tensor decomposition structure in a data-adaptive manner, we employ the top-$k$ gating mechanism with learnable gating scores to dynamically select a suitable single-candidate structure or combine suitable multi-candidate structures. The $\mathrm{softmax}(\cdot)$ gating function models the probability distribution across heterogeneous interaction-induced candidate structures by computing the weighted sum exclusively over the outputs of the top-$k$ heterogeneous interaction-induced candidate structures. The mathematical formulation of the top-$k$ gating mechanism is presented as follows:
	\begin{equation}
	\lambda_{\theta_i}=\mathrm{softmax}(\mathrm{top}(\{g_{\theta_i}\}_{i=1}^L,k))_{i},
	\end{equation}
	where $\{\lambda_{\theta_i}\}_{i=1}^k$ are the normalized gating scores, $\{g_{\theta_i}\}_{i=1}^L$ represent the gating scores, $\{{\theta_i}\}_{i=1}^L$ denote the learnable parameters of gating scores, and $L$ denotes the number of heterogeneous interaction-induced candidate structures. The $\mathrm{top}(\cdot,k)$ function retains only the top-$k$ entries of a vector at their original values, while setting all other entries to $-\infty$. Following the $\mathrm{softmax}(\cdot)$ operation, those entries assigned $-\infty$ become approximately zero.
	
	Based on this gating mechanism, the proposed I-TSS can identify either a single structure beyond a predefined interaction family or a mixture of structures involving heterogeneous interaction families, which is detailed below.
	\begin{remark} 
		The search results of the proposed I-TSS with different top-$k$ values:
		\begin{itemize}
			\item When $k=1$, the proposed I-TSS provides a suitable single structure beyond a predefined interaction family.
			\item When $k>1$, the proposed I-TSS configures a suitable mixture of structures involving heterogeneous interaction families.
		\end{itemize}
		\label{sm}
		
	\end{remark}
%	\begin{remark}The complexity analysis of the proposed I-TSS with different top-$k$ values: 
%		\begin{itemize}
%			\item When $k<L$, the proposed I-TSS activates top-$k$ heterogeneous interaction-induced candidate structures, making effective parameters lower than total.
%			\item When $k=L$, even if all heterogeneous interaction-induced candidate structures are activated, the low-rank structure design can alleviate parameter explosion.
%		\end{itemize}
%	\end{remark}
%		
\begin{algorithm}[t]
	\caption{Unified Energy-Based Rank Estimation}
	\label{u}
	\begin{algorithmic}[1] % 每行显示行号
		\REQUIRE An $N$th-order ($N \geq 3$) tensor $\mathcal{X}\in \mathbb{R}^{I_1\times I_2\times \cdots \times I_N}$ and the energy ratio thresholds $\{\tau_i\}_{i=1}^3$. % 输入
		\ENSURE Estimated rank and parameter corresponding to the specified decomposition type.  % 输出
		%			\STATE {\textbf{Rank Estimation for Tucker:}}
		\STATE {\textbf{Energy-Based Rank Estimation $\mathrm{ERM}(\mathbf{X},\tau)$:}} 
		\STATE Compute SVD of input matrix $\mathbf{X}$: $\sigma_1 \geq \sigma_2 \geq \dots \geq \sigma_n \geq 0$ (singular values) and left singular matrix $\mathbf{U}$;
		\STATE {\textbf{Initialization:}} $E_{\text{total}} = \sum_{j=1}^n \sigma_j^2$, $E_{\text{accum}} = 0$, estimated rank $\hat{r} = 0$;
		\FOR{$k = 1$ to  $n$}
		\STATE Compute $E_{\text{accum}} += \sigma_k^2$; $E_{\text{ratio}} = E_{\text{accum}}/E_{\text{total}}$;
		\IF{$E_{\text{ratio}} \geq \tau$}
		\STATE $\hat{r}=k$ and break.
		\ENDIF
		\ENDFOR
		\RETURN Estimated rank $\hat{r}$ and truncated left singular matrix $\mathbf{U}(:,1:\hat{r})$.
		\IF {Rank Estimation for Tucker}
		\FOR{$n = 1$ to  $N$}
		\STATE Estimated $\mathrm{rank}(\mathbf{X}_{<n>})$ by $\mathrm{ERM}(\mathbf{X}_{<n>},\tau_1)$;
		\ENDFOR
		\STATE Compute Tucker rank $R_n=\mathrm{rank}(\mathbf{X}_{<n>})$.
		%			\RETURN Tucker rank $[R_1,R_2,\cdots,R_N]$.
		\ELSIF{Rank Estimation for FCTN}
		%			\STATE {\textbf{Rank Estimation for FCTN:}}
		\FOR{$n = 1$ to  $N$}
		\STATE Estimate $\mathrm{rank}(\mathbf{X}_{<n>})$ by $\mathrm{ERM}(\mathbf{X}_{<n>},\tau_2)$;
		%		\FOR{$n_2 = n_1+1$ to  $N$}
		%		\STATE Estimate $\mathrm{rank}(\mathbf{X}_{<n_1n_2>})$ by $\mathrm{ERM}(\mathbf{X}_{<n_1n_2>},\tau_2)$;
		\ENDFOR 
		%		\ENDFOR
		%		\STATE Compute FCTN rank $R_{n_{1},n_{2}}=\text{round}\left(\sqrt{\frac{\mathrm{rank}({\mathbf{X}_{< n_{1}>})\mathrm{rank}(\mathbf{X}_{< n_{2}>})}}{\mathrm{rank}(\mathbf{X}_{< n_{1}n_{2}>})}}\right)$.
		\STATE Compute FCTN rank $R_{n_{1},n_{2}}=\min(\mathrm{rank}(\mathbf{X}_{<n_1>}),\mathrm{rank}(\mathbf{X}_{<n_2>}))$.
		%			\RETURN FCTN rank $R_{n_1,n_2}(1\le n_1 < n_2\le N)$.
		\ELSIF{Rank Estimation for TF}
		%			\STATE {\textbf{Rank Estimation for TF:}}
		\STATE Estimate $\mathrm{rank}(\mathbf{X}_{<3>})$ and truncated left singular matrix $\hat{\mathbf{L}}$ by $\mathrm{ERM}(\mathbf{X}_{<3>},\tau_3)$;
		\FOR{$n = 1$ to  $\mathrm{rank}(\mathbf{X}_{<3>})$}
		\STATE Estimate $\mathrm{rank}\left((\mathcal{X}\times_3\hat{\mathbf{L}}^\top)^{(n)}\right)$ by $\mathrm{ERM}((\mathcal{X}\times_3\hat{\mathbf{L}}^\top)^{(n)},\tau_3)$;
		\ENDFOR
		\STATE Compute TF tubal-rank $R=\max \limits_{n=1,2,\cdots,\mathrm{rank}(\mathbf{X}_{<3>})}{\mathrm{rank}\left((\mathcal{X}\times_3\hat{\mathbf{L}}^\top)^{(n)}\right)}$.
		\STATE Compute latent dimension $\ell=\mathrm{rank}(\mathbf{X}_{<3>})$.
		%			\RETURN TF tubal-rank $R$ and parameter $\ell$.
		\ENDIF
	\end{algorithmic}
\end{algorithm}
\begin{algorithm}[t]
	\caption{Unsupervised Multi-Dimensional Data Recovery Model}
	\label{I-TSS}
	\begin{algorithmic}[1] % 每行显示行号
		\REQUIRE Observed data $\mathcal{O}$, observed binary mask $\mathcal{M}$, energy ratio thresholds $\{\tau_i\}_{i=1}^L$, number of max iterations $t_{max}$, and top-$k$ value.  % 输入
		\ENSURE Recovered data $\mathcal{X}$. % 输出
		\STATE  {\textbf{Initialization:}} Heterogeneous interaction-induced candidate structures $\{\mathcal{E}_{\phi_{i}}\}_{i=1}^L$ and gating scores $\{g_{\theta_i}\}_{i=1}^L$;
		\STATE  Estimate ranks for heterogeneous interaction-induced candidate structures via algorithm\ref{u};
		\FOR{$t = 1$ to  $t_{max}$}
		\STATE  Compute estimated data $\mathcal{\hat{X}}\!=\!\sum_{i=1}^L\lambda_{\theta_i}\mathcal{E}_{\phi_i}$ with top-$L$;
		\STATE  Alternating update $\{\phi_i\}_{i=1}^L$ and $\{{\theta_i}\}_{i=1}^L$ via Adam;
		\ENDFOR
		\IF {$k<L$}
		\STATE {\textbf{Reinitialization:}} The selected top-$k$ heterogeneous interaction-induced candidate structures  $\{\mathcal{E}_{\theta_{i}}\}_{i=1}^k$;
		\FOR{$t = 1$ to  $t_{max}$}
		\STATE  Compute estimated data $\mathcal{\hat{X}}=\sum_{i=1}^k \lambda_{\theta_i}\mathcal{E}_{\phi_i}$ with top-$k$;
		\STATE  Update $\{\phi_i\}_{i=1}^k$ via Adam;
		\ENDFOR
		\ENDIF
		\IF {$\mathcal{M}==1$}
		\STATE Compute recovered data $\mathcal{X}=\mathcal{\hat{X}}$.
		\ELSIF{$\mathcal{M}\neq1$}
		\STATE Compute recovered data $\mathcal{X}=\mathcal{M}\odot\mathcal{O}+(1-\mathcal{M})\odot\mathcal{\hat{X}}$.
		\ENDIF
	\end{algorithmic}
\end{algorithm}
		\subsection{The Approximation Error Bound of I-TSS}
	  Theoretically, we provide the approximation error bound of I-TSS in theorem~\ref{r5}--\ref{error bound}, which reveals the approximation capability of I-TSS, clarifying its capability to approximate arbitrary tensors within a guaranteed error range.
	  \begin{theorem}
	  	Let $\mathcal{X} \in \mathbb{R}^{I_1 \times I_2 \times \dots \times I_N}$ be an $N$th-order tensor. $\mathcal{X}$ can be linearly decomposed as
	  	\[
	  	\mathcal{X} = \sum_{i=1}^K \lambda_i \mathcal{T}_i,
	  	\]
	  	where $\lambda_i$ denotes the weighting coefficient, $\{\mathcal{T}_i\}_{i=1}^L$ denotes a set of mutually independent tensors, and $K$ denotes the number of independent tensors.
	  	
	  	Define $\mathcal{S} = \{\mathcal{E}_1,\mathcal{E}_2,\dots,\mathcal{E}_L\}$ as the heterogeneous interaction-induced candidate structure set. Let $\hat{\mathcal{X}}_{\mathcal{S}}$ and $\hat{\mathcal{X}}_{\mathcal{E}}$ denote the optimal approximations of $\mathcal{X}$ fitted by $\mathcal{S}$ and single structure $\mathcal{E}\in\mathcal{S}$, respectively. Based on the Lemma \ref{le}, it exits that
	  	\[
	  	\|\mathcal{X} - \hat{\mathcal{X}}_{\mathcal{S}}\|_F \leq \|\mathcal{X} - \hat{\mathcal{X}}_{\mathcal{E}}\|_F, \quad \forall \mathcal{E}\in\mathcal{S}.
	  	\]
	  	\label{r5}
	  \end{theorem}
%	  	\begin{figure}[t]
%	  	\centering
%	  	\subfloat{
%	  		\begin{minipage}[b]{0.48\linewidth}
%	  			\captionsetup{
%	  				font={small}, 
%	  				labelfont=bf,        % 标签字体为粗体
%	  				textfont={rm}, % 正文字体为 Times new roman
%	  				singlelinecheck=true % 不允许换行 
%	  			}
%	  			\includegraphics[width=\hsize]{er1.png}	
%	  	\end{minipage}}
%	  	\subfloat{
%	  		\begin{minipage}[b]{0.48\linewidth}
%	  			\captionsetup{
%	  				font={small}, 
%	  				labelfont=bf,        % 标签字体为粗体
%	  				textfont={rm}, % 正文字体为 Times new roman
%	  				singlelinecheck=true % 不允许换行 
%	  			}
%	  			\includegraphics[width=\hsize]{er2.png}
%	  	\end{minipage}}
%	  	\caption{Comparison between candidate structures and our proposed I-TSS in terms of relative error for fitting synthetic data. ``Mixture of Decompositions" denotes a uniform mixture strategy for Tucker decomposition, FCTN decomposition, and TF. For simplicity, we uniformly set all ranks of Tucker decomposition, FCTN decomposition, and TF as the same values.}
%	  	\label{fig}
%	  \end{figure}
	  
	 	\begin{theorem}[Approximation Error Bound of I-TSS]
	 	\label{error bound}
	 	Let $\mathcal{X} \in \mathbb{R}^{I_1 \times I_2 \times I_3}$ be a third-order\footnote{This approximation error bound of I-TSS can be easily extended to higher-order.} tensor with Tucker rank $[R_1, R_2, R_3]$, FCTN rank $ [R_{1,2}, R_{1,3}, R_{2,3}]$, TF tubal-rank $R$ and latent dimension $\ell$. Consider the I-TSS representation:
	 	\[
	 	\mathcal{\hat{X}} =\sum_{i=1}^3\lambda_{\theta_i}\mathcal{E}_{\phi_i},
	 	\]
	 	where $\{\mathcal{E}_{\phi_{i}}\}_{i=1}^3\in \mathbb{R}^{I_{1}\times I_2 \times I_3}$ represent the candidate Tucker decomposition, FCTN decomposition, and TF, respectively. $\{\lambda_{\theta_i}\}_{i=1}^3$ are the normalized gating scores. Under the following assumptions:
	 	\begin{itemize}
	 		\item $||\mathcal{X}||_F\leq K$, $K$ is a constant.
	 		\item The factor matrices $\mathbf{H}_{n}$ of Tucker decomposition are orthogonal.
	 		\item The linear transform matrix $\mathbf{L}$ of TF is semi-orthogonal.
	 		\item $R_n \leq \mathrm{rank}(\mathbf{X}_{<n>})$ $\forall n \in \{1,2,3\}$.
%	 		\item $\prod_{n_2=1}^{3}R_{n_1,n_2} \leq \mathrm{rank}(\mathbf{X}_{<n_1>})$ $\forall n_1 \in \{1,2,3\}$.
	 		\item $R_{n_1,n_2} \leq \min(\mathrm{rank}(\mathbf{X}_{<n_1>},\mathbf{X}_{<n_2>}))$ $\forall n_1,n_2 \in \{1,2,3\}\ and\ n_1 <n_2$.
	 		\item $R \leq \mathrm{rank}(\mathbf{X}^{(n)})$ $\forall n \in \{1,2,\cdots, \ell\}$.
	 		\item $\ell \leq \mathrm{rank}(\mathbf{X}_{<3>})$.
	 	\end{itemize}
	 	the approximation error bound of I-TSS satisfies:
	 	\begin{equation}
	 		\begin{aligned}
	 			&\|\mathcal{X} - \mathcal{\hat{X}}\|_F^2\\ \leq 
	 			&\lambda_{\theta_1}^2\sum_{n=1}^3\sum_{i=R_n+1}^{\mathrm{rank}(\mathbf{X}_{<n>})} \sigma_i^2(\mathbf{X}_{<n>})\\
	 			+&\lambda_{\theta_2}^2\sum_{n_1=1}^2\sum_{n_2=n_1+1}^3\sum_{i=R_{n_1,n_2}+1}^{\mathrm{rank}(\mathbf{X}_{<n_1n_2>})} \sigma_i^2(\mathbf{X}_{<n_1n_2>})\\
	 			+&\lambda_{\theta_3}^2\!\biggl(\sum_{n=1}^{\ell}\!\sum_{i=R+1}^{\mathrm{rank}(\mathbf{X}^{(n)})} \!\sigma_i^2(\mathbf{X}^{(n)})+\!\sum_{i=\ell+1}^{\mathrm{rank}(\mathbf{X}_{<3>})}\! \sigma_i^2(\mathbf{X}_{<3>})\!\biggr)\!,
	 		\end{aligned}
	 		\label{iq}
	 	\end{equation}
	 	where $\sigma_i(\cdot)$ denotes the $i$-th singular value and $\mathbf{X}_{<n_1n_2>}$ represents the matrix with the smaller rank between $\mathbf{X}_{<n_1>}$ and $\mathbf{X}_{<n_2>}$.
	 	
	 	The proof is provided in supplementary material.
%	 	\begin{equation}
%	\begin{aligned}
%		&\|\mathcal{X} - \mathcal{\hat{X}}\|_F^2\\ \leq 
%		&\lambda_{\theta_1}^2\sum_{n=1}^3\sum_{i=R_n+1}^{\mathrm{rank}(\mathbf{X}_{<n>})} \sigma_i^2(\mathbf{X}_{<n>})\\
%		+&\lambda_{\theta_2}^2\sum_{n_1=1}^3\sum_{i=\alpha_{n_2}}^{\mathrm{rank}(\mathbf{X}_{<n_1>})} \sigma_i^2(\mathbf{X}_{<n_1>})\\
%		+&\lambda_{\theta_3}^2\!\biggl(\sum_{n=1}^{\ell}\!\sum_{i=R+1}^{\mathrm{rank}(\mathbf{X}^{(n)})} \!\sigma_i^2(\mathbf{X}^{(n)})+\!\sum_{i=\ell+1}^{\mathrm{rank}(\mathbf{X}_{<3>})}\! \sigma_i^2(\mathbf{X}_{<3>})\!\biggr)\!,
%	\end{aligned}
%\end{equation}

	 \end{theorem}

	%-------------------------------------------------------------------------

	\subsection{Unsupervised Multi-Dimensional Data Recovery Model}
	\label{section:5}
	To examine the potential of the proposed I-TSS, we evaluate it on the representative multi-dimensional data recovery task and suggest an unsupervised I-TSS-based multi-dimensional data recovery model as follows:
	\begin{equation}
		\begin{aligned}
			\min\limits_{\substack{\{\phi_i\}_{i=1}^L, \{\theta_i\}_{i=1}^L}} &\|\mathcal{M}\odot(\mathcal{\hat{X}}-\mathcal{O})\|_{F}^2,\\ 
			\text{s.t.}\ \mathcal{\hat{X}} = &\sum_{i=1}^k\lambda_{\theta_i}\mathcal{E}_{\phi_i},\\
			\lambda_{\theta_i}=&\ \mathrm{softmax}(\mathrm{top}(\{g_{\theta_i}\}_{i=1}^L,k))_{i},
		\end{aligned}
		\label{e9}
	\end{equation}
	where $\mathcal{O}$ is the observed multi-dimensional data and $\mathcal{\hat{X}}$ is the data estimated by the proposed I-TSS. $\mathcal{M}$ is a binary mask tensor that sets the observed location to ones and others to zeros.  Specifically, when \(\mathcal{M}\) is an all-ones tensor, this recovery model degenerates to a data fitting model. When \(\mathcal{M}\) is not an all-ones tensor, this recovery model is a data completion model. $\{\mathcal{E}_{\phi_i}\}_{i=1}^L$ are the heterogeneous interaction-induced candidate structures and $\{{\phi_i}\}_{i=1}^L$ denote the learnable parameters of heterogeneous interaction-induced candidate structures. $\{\lambda_{\theta_i}\}_{i=1}^k$ are the normalized gating scores, $\{g_{\theta_i}\}_{i=1}^L$ are the gating scores, and $\{{\theta_i}\}_{i=1}^L$ denote the learnable parameters of gating scores. $L$ denotes the number of heterogeneous interaction-induced candidate structures. The $\mathrm{top}(\cdot,k)$ function retains only the top-$k$ entries of a vector at their original values, while setting all other entries to $-\infty$.

	To solve model (\ref{e9}), we leverage the adaptive moment estimation (Adam) algorithm \cite{Kingma2014AdamAM}, which is widely used in deep learning, to efficiently optimize the parameter set $\psi=\{\{\phi_i\}_{i=1}^L, \{\theta_i\}_{i=1}^L\}$. We summarize the complete procedure of the proposed unsupervised I-TSS-based multi-dimensional data recovery model in algorithm \ref{I-TSS}.

 \section{Experiments}
 
 \label{section:6}
 To verify the effectiveness and superiority of the proposed I-TSS, we conduct extensive comparative experiments and analyses on various multi-dimensional datasets, encompassing both synthetic and real-world datasets (including MSIs, color videos, and light field data).
 First, we provide a detailed introduction to the experimental settings, followed by the corresponding experimental results and in-depth analysis. 
  All experiments are performed on a PC equipped with one Intel(R) Core(TM) i7-8700K CPU, one NVIDIA GeForce RTX 4070 Ti GPU, and 32 GB RAM. The proposed I-TSS are implemented by using Python and the PyTorch library with GPU calculation.
 
 \subsection{Fitting Experiments for Synthetic and Real-World Datasets}
 \subsubsection{Evaluation Metrics}
 To quantitatively evaluate the performance of different methods, we adopted two widely used metrics: the compression rate (CR) and the relative error (RE). CR is defined as \(N_\mathcal{S}/N_\mathcal{X} \times 100\%\), with \(N_\mathcal{S}\) denoting the number of elements in the factor matrices/tensors, and \(N_\mathcal{X}\) representing the total number of elements in the original tensor. RE is defined as \(||\mathcal{X}-\hat{\mathcal{X}}||_F/||\mathcal{X}||_F\), where \(\mathcal{X}\) is the original tensor and \(\hat{\mathcal{X}}\) is the estimated tensor.
 
  \subsubsection{Compared Methods}
 To comprehensively evaluate the effectiveness of I-TSS for synthetic data, we compare it against four representative tensor decomposition structure search methods: Tucker \cite{tucker}, TF \cite{8066348}, TNGreedy \cite{Hashemizadeh}, and SVDinsTN \cite{10655696}. For the sake of fairness, the rank settings of Tucker and TF are also estimated by algorithm \ref{u}.
  
  \subsubsection{Experimental Data}
 For synthetic data of size $50\times50\times50$, we first randomly generate the factor matrices/tensors of Tucker decomposition, FCTN decomposition, and TF, where all elements of these factors follow a Gaussian distribution (with values constrained between 0 and 1). Then, the synthetic tensors are constructed via the specific factor interactions detailed in Section \ref{section:2}.
 For real-world data, we collect three MSIs\footnote[2]{Available at https://cave.cs.columbia.edu/repository/Multispectral} (\emph{Cloth}, \emph{Beads}, and \emph{Jelly}) as testing data.

\begin{table}[t]
	\footnotesize
	\centering

	\caption{ Quantitative comparison of CR $(\downarrow)$ and RE $(\downarrow)$ of different	methods on third-order synthetic data.}
	\label{sy3}
	\setlength{\tabcolsep}{1.4mm}{
		\begin{tabular}{ccccccc}
			\toprule
			\multirow{2}{*}{Method} & \multicolumn{2}{c}{Rank: 3} & \multicolumn{2}{c}{Rank: 5} & \multicolumn{2}{c}{Rank: 7}\\
			\cmidrule(r){2-3} \cmidrule(r){4-5} \cmidrule(r){6-7}
			&CR&RE&CR&RE&CR&RE\\
			\midrule
			\multicolumn{7}{c}{Synthetic Data (Tucker)}\\
			\midrule
			
			Tucker&0.3\%&$\bf 3*10^{-5}$&0.7\%&$\bf2*10^{-5}$&1.1\%&$\bf2*10^{-5}$\\
			TF&0.4\%&$4*10^{-1}$&1.0\%&$2*10^{-1}$&1.9\%&$2*10^{-1}$\\
			TNGreedy&0.4\%&$6*10^{-2}$&1.5\%&$9*10^{-4}$&3.0\%&$8*10^{-4}$\\
			SVDinsTN&0.6\%&$8*10^{-4}$&2.2\%&$5*10^{-4}$&4.3\%&$2*10^{-4}$\\
			I-TSS&0.3\%&$\bf 3*10^{-5}$&0.7\%&$\bf2*10^{-5}$&1.1\%&$\bf2*10^{-5}$\\
			%				I-TSS-M&14.8\%&$3*10^{-5}$&19.7\%&$9*10^{-5}$&24.1\%&$5*10^{-5}$\\
			\midrule
			
			\multicolumn{7}{c}{Synthetic Data (FCTN)}\\
			\midrule
			Tucker&1.0\%&$4*10^{-1}$&3.4\%&$6*10^{-1}$&8.0\%&$6*10^{-1}$\\
			TF&1.4\%&$4*10^{-1}$&3.6\%&$6*10^{-1}$&6.8\%&$7*10^{-1}$\\
			TNGreedy&1.0\%&$5*10^{-4}$&3.4\%&$5*10^{-5}$&5.8\%&$8*10^{-4}$\\
			SVDinsTN&1.6\%&$2*10^{-4}$&5.2\%&$9*10^{-4}$&8.9\%&$1*10^{-3}$\\
			I-TSS&1.0\%&$\bf1*10^{-5}$&3.0\%&$\bf1*10^{-5}$&5.8\%&$\bf9*10^{-5}$\\
			%				I-TSS-M&28.2\%&$4*10^{-5}$&55.2\%&$1*10^{-4}$&79.9\%&$4*10^{-4}$\\
			\midrule
			\multicolumn{7}{c}{Synthetic Data (TF)}\\
			\midrule
			%				TNGA&17.6\%&$8*10^{-1}$&27.0\%&$8*10^{-1}$&38.4\%&$8*10^{-1}$\\
			%				TNLS&16.2\%&$5*10^{-1}$&25.8\%&$5*10^{-1}$&41.3\%&$4*10^{-1}$\\
			%				TNALE&15.1\%&$8*10^{-1}$&24.6\%&$6*10^{-1}$&38.7\%&$4*10^{-1}$\\
			%				SVDinsTN&18.1\%&$6*10^{-1}$&23.9\%&$5*10^{-1}$&31.6\%&$5*10^{-1}$\\
			%				I-TSS&14.0\%&$\bf1*10^{-5}$&22.0\%&$\bf1*10^{-5}$&30.0\%&$\bf1*10^{-5}$\\
			%				I-TSS-M&48.2\%&$7*10^{-5}$&63.2\%&$6*10^{-5}$&74.3\%&$6*10^{-5}$\\
			Tucker&6.9\%&$2*10^{-1}$&8.2\%&$3*10^{-1}$&9.2\%&$4*10^{-1}$\\
			TF&6.0\%&$\bf2*10^{-5}$&7.6\%&$\bf2*10^{-5}$&8.4\%&$\bf2*10^{-5}$\\
			TNGreedy&6.4\%&$7*10^{-1}$&7.6\%&$8*10^{-1}$&9.0\%&$8*10^{-1}$\\
			SVDinsTN&6.8\%&$8*10^{-1}$&8.3\%&$8*10^{-1}$&10.7\%&$7*10^{-1}$\\
			I-TSS&6.0\%&$\bf2*10^{-5}$&7.6\%&$\bf2*10^{-5}$&8.4\%&$\bf2*10^{-5}$\\
			\midrule
			\multicolumn{7}{c}{Synthetic Data (Mixture of Structures)}\\
			\midrule
			%				TNGA&38.4\%&$7*10^{-1}$&67.2\%&$6*10^{-1}$&104.0\%&$6*10^{-1}$\\
			%				TNLS&57.9\%&$1*10^{-1}$&71.9\%&$1*10^{-1}$&102.3\%&$8*10^{-2}$\\
			%				TNALE&40.0\%&$1*10^{-1}$&70.5\%&$7*10^{-2}$&114.0\%&$4*10^{-2}$\\
			%				SVDinsTN&43.2\%&$9*10^{-2}$&61.4\%&$6*10^{-2}$&96.0\%&$3*10^{-2}$\\
			%%				I-TSS&30.0\%&$2*10^{-1}$&11.1\%&$3*10^{-1}$&5.8\%&$5*10^{-1}$\\
			%				I-TSS&38.1\%&$\bf1*10^{-4}$&61.4\%&$\bf3*10^{-4}$&94.6\%&$\bf4*10^{-4}$\\
			Tucker&30.7\%&$3*10^{-2}$&59.1\%&$1*10^{-1}$&86.8\%&$1*10^{-1}$\\
			TF&29.1\%&$4*10^{-1}$&56.2\%&$4*10^{-1}$&89.6\%&$4*10^{-1}$\\
			TNGreedy&29.3\%&$6*10^{-2}$&54.3\%&$2*10^{-2}$&87.3\%&$8*10^{-3}$\\
			SVDinsTN&43.2\%&$9*10^{-2}$&67.0\%&$5*10^{-2}$&89.5\%&$4*10^{-2}$\\
			I-TSS&28.7\%&$\bf1*10^{-4}$&54.0\%&$\bf1*10^{-4}$&85.5\%&$\bf1*10^{-4}$\\
			\bottomrule
	\end{tabular}}
\end{table}

 \subsubsection{Experimental Results of Synthetic and Real-World Data}
 To systematically evaluate the performance of the proposed I-TSS, we conduct fitting experiments on synthetic third-order tensor datasets and real-world MSIs. Tables~\ref{sy3} and \ref{sy4} summarize the full quantitative results of the compared methods and the proposed I-TSS. 
 
 In Table~\ref{sy3}, several experimental settings and notation definitions need to be clarified to avoid confusion: (i) The column labeled ``Rank'' refers to the ranks used for Tucker decomposition, FCTN decomposition, and TF. For simplicity, we uniformly set all ranks of Tucker decomposition, FCTN decomposition, and TF as the same values. Meanwhile, we set the latent dimension $\ell$ of TF as $10$.
(ii) The terms ``Tucker", ``FCTN", and ``TF"  in the table denote synthetic single structures. These data are designed to verify the effectiveness of top-$1$ I-TSS. (iii) The term ``Mixture of Structures" in the table denotes a uniform mixture strategy for heterogeneous interaction-induced candidate structures. This strategy involves combining all heterogeneous interaction-induced candidate structures at equal weights to verify the effectiveness of top-$L$ I-TSS. In Table~\ref{sy4}, to evaluate the effectiveness of I-TSS for complex real-world data, we compare top-$L$ I-TSS against other tensor decomposition structure search methods.

The quantitative results reported in Tables~\ref{sy3}-\ref{sy4} yield two insightful observations that illustrate the advantages brought by our I-TSS framework.
First, under the single structure experimental setting, the proposed I-TSS can dynamically provide a suitable single structure beyond a predefined interaction family. This advantage is reflected in smaller or the same relative errors with almost the same CR compared to competitors, which are restricted to single structure within a predefined interaction family. 
Second, when handling tensors generated from mixture of structures and practical real‑world MSIs, I-TSS achieves significantly better results than classic tensor decomposition structure search methods, which can be attributed to configuring a suitable mixture of structures involving heterogeneous interaction families.

\begin{table}[t]
	\footnotesize
	\centering

	\caption{ Quantitative comparison of CR $(\downarrow)$ and RE $(\downarrow)$ of different methods on real-world MSIs.}
	\label{sy4}
	\setlength{\tabcolsep}{1.4mm}{
		\begin{tabular}{ccccccc}
			\toprule
			\multirow{2}{*}{Method} & \multicolumn{2}{c}{\emph{Cloth}} & \multicolumn{2}{c}{\emph{Beads}}& \multicolumn{2}{c}{\emph{Jelly}}\\
			\cmidrule(r){2-3} \cmidrule(r){4-5} \cmidrule(r){6-7} 
			&CR&RE&CR&RE&CR&RE\\
			\midrule
			Tucker&37.8\%&$1*10^{-2}$&39.3\%&$1*10^{-2}$&34.5\%&$1*10^{-2}$\\
			TF&34.4\%&$1*10^{-2}$&39.0\%&$1*10^{-2}$&35.4\%&$1*10^{-2}$\\
			TNGreedy&31.7\%&$7*10^{-3}$&37.8\%&$6*10^{-3}$&32.9\%&$7*10^{-3}$\\
			SVDinsTN&32.8\%&$7*10^{-3}$&39.9\%&$9*10^{-3}$&34.8\%&$8*10^{-3}$\\
			I-TSS&31.7\%&$\bf6*10^{-3}$&37.8\%&$\bf5*10^{-3}$&32.6\%&$\bf6*10^{-3}$\\
			% 			I-TSS-M&0.07\%&$1*10^{-4}$&0.10\%&$5*10^{-5}$\\
			% 			\midrule
			% 			\multicolumn{5}{c}{Synthetic Data (FCTN)}\\
			% 			\midrule
			% 			Tucker&0.1\%&$9*10^{-1}$&2.5\%&$9*10^{-1}$\\
			% 			TF&0.2\%&$8*10^{-1}$&0.5\%&$8*10^{-1}$\\
			% 			TNGreedy&0.3\%&$3*10^{-1}$&0.5\%&$7*10^{-1}$\\
			% 			SVDinsTN&0.2\%&$1*10^{-1}$&0.4\%&$5*10^{-1}$\\
			% 			I-TSS&0.1\%&$\bf2*10^{-5}$&0.4\%&$\bf1*10^{-5}$\\
			% 			% 			I-TSS-M&0.5\%&$4*10^{-5}$&14.3\%&$2*10^{-5}$\\
			% 			\midrule
			% 			
			% 			\multicolumn{5}{c}{Synthetic Data (Mixture of Decompositions)}\\
			% 			\midrule
			% 			Tucker&3.1\%&$9*10^{-1}$&3.5\%&$9*10^{-1}$\\
			% 			TF&0.4\%&$3*10^{-1}$&3.0\%&$3*10^{-1}$\\
			% 			TNGreedy&0.7\%&$1*10^{-2}$&3.0\%&$2*10^{-2}$\\
			% 			SVDinsTN&0.5\%&$1*10^{-2}$&3.4\%&$2*10^{-3}$\\
			% 			% 			I-TSS-T&0.2\%&$1*10^{-1}$&0.4\%&$3*10^{-1}$\\
			% 			I-TSS&0.4\%&$\bf8*10^{-5}$&2.9\%&$\bf5*10^{-5}$\\
			\bottomrule
	\end{tabular}}
\end{table}
 \subsection{Completion Experiments for Real-World  Datasets}
 \subsubsection{Evaluation Metrics}
To quantitatively evaluate the performance of different methods, we adopted three widely used evaluation metrics: peak signal-to-noise ratio (PSNR), structural similarity index measure (SSIM), and RE. In general, higher PSNR and SSIM values, and lower RE values, indicate better performance of the evaluated methods.
 \subsubsection{Compared Methods}
 To comprehensively evaluate the effectiveness of I-TSS for complex real-world data, we compare top-$L$ I-TSS against six classical tensor decomposition-based methods: HaLRTC \cite{101109}, SiLRTCTT \cite{7859390}, TRLRF \cite{101609}, HTNN \cite{9730793}, LTNN \cite{10507804}, and SVDinsTN \cite{10655696}. To ensure all methods operate at their peak performance, their respective hyperparameters are carefully tuned.
 
 \subsubsection{Experimental Data}
 We collect four MSIs (\emph{Flowers}, \emph{Beers}, \emph{Balloons}, and \emph{Feathers}), four color videos\footnote[3]{Available at https://traces.cnets.io/trace.eas.asu.edu/yuv/index.html} (\emph{News}, \emph{Claire}, \emph{Grandma}, and \emph{Akiyo}), and four light field data\footnote[4]{Available at  http://hci-lightfield.iwr.uni-heidelberg.de} (\emph{Greek}, \emph{Museum}, \emph{Medieval2}, and \emph{Vinyl}) as testing data. The sampling rates (SRs) are set as $[0.1, 0.15, 0.2, 0.25, 0.3]$. Before the experiments, the gray values of all datasets are normalized into the interval $[0, 1]$. 
%\begin{table}[t]
%	\footnotesize
%	\centering
%	
%	
%	\caption{ Quantitative comparison of CR $(\downarrow)$ and RE $(\downarrow)$ of different methods on fourth-order synthetic data.}
%	\label{sy4}
%	\setlength{\tabcolsep}{3.8mm}{
%		\begin{tabular}{ccccc}
%			\toprule
%			\multirow{2}{*}{Method} & \multicolumn{2}{c}{Rank: 2} & \multicolumn{2}{c}{Rank: 3}\\
%			\cmidrule(r){2-3} \cmidrule(r){4-5} 
%			&CR&RE&CR&RE\\
%			\midrule
%			\multicolumn{5}{c}{Synthetic Data (Tucker)}\\
%			\midrule
%			TNGA&0.08\%&$5*10^{-1}$&0.15\%&$8*10^{-1}$\\
%			TNLS&0.11\%&$2*10^{-2}$&0.17\%&$3*10^{-1}$\\
%			TNALE&0.03\%&$6*10^{-1}$&0.10\%&$5*10^{-1}$\\
%			SVDinsTN&0.08\%&$7*10^{-1}$&0.08\%&$3*10^{-1}$\\
%			I-TSS&0.03\%&$\bf3*10^{-5}$&0.05\%&$\bf2*10^{-5}$\\
%			% 			I-TSS-M&0.07\%&$1*10^{-4}$&0.10\%&$5*10^{-5}$\\
%			\midrule
%			\multicolumn{5}{c}{Synthetic Data (FCTN)}\\
%			\midrule
%			TNGA&0.1\%&$9*10^{-1}$&2.5\%&$9*10^{-1}$\\
%			TNLS&0.2\%&$8*10^{-1}$&0.5\%&$8*10^{-1}$\\
%			TNALE&0.3\%&$3*10^{-1}$&0.5\%&$7*10^{-1}$\\
%			SVDinsTN&0.2\%&$1*10^{-1}$&0.4\%&$5*10^{-1}$\\
%			I-TSS&0.1\%&$\bf2*10^{-5}$&0.4\%&$\bf1*10^{-5}$\\
%			% 			I-TSS-M&0.5\%&$4*10^{-5}$&14.3\%&$2*10^{-5}$\\
%			\midrule
%			
%			\multicolumn{5}{c}{Synthetic Data (Mixture of Decompositions)}\\
%			\midrule
%			TNGA&3.1\%&$9*10^{-1}$&3.5\%&$9*10^{-1}$\\
%			TNLS&0.4\%&$3*10^{-1}$&3.0\%&$3*10^{-1}$\\
%			TNALE&0.7\%&$1*10^{-2}$&3.0\%&$2*10^{-2}$\\
%			SVDinsTN&0.5\%&$1*10^{-2}$&3.4\%&$2*10^{-3}$\\
%			% 			I-TSS-T&0.2\%&$1*10^{-1}$&0.4\%&$3*10^{-1}$\\
%			I-TSS&0.4\%&$\bf8*10^{-5}$&2.9\%&$\bf5*10^{-5}$\\
%			\bottomrule
%	\end{tabular}}
%\end{table}

\begin{table*}[!h]\scriptsize
	\centering
	
	\caption{The quantitative results for real-world MSI completion by different methods. The \textbf{best} and \underline{second-best} values are highlighted.}
	\label{result3}
	\setlength{\tabcolsep}{2mm}{
		\begin{tabular}{ccccccccccccccccc}
			\toprule
			\multicolumn{2}{ c }{SR}&\multicolumn{3}{ c }{0.1 }&\multicolumn{3}{ c }{0.15 }&\multicolumn{3}{ c }{0.2}&\multicolumn{3}{ c }{0.25}&\multicolumn{3}{ c }{0.3}\\
			\midrule
			Data&Method&PSNR&SSIM&RE&PSNR&SSIM&RE&PSNR&SSIM&RE&PSNR&SSIM&RE&PSNR&SSIM&RE\\
			\midrule
			&HaLRTC&20.54&0.672&0.592&29.89&0.830&0.202&31.73&0.864&0.163&33.50&0.888&0.133&33.86&0.895&0.127\\
			&SiLRTCTT&28.06&0.749&0.249&31.22&0.809&0.173&33.53&0.843&0.132&35.44&0.864&0.106&36.93&0.880&0.089\\
			\multirow{2}{*}{\emph{Flowers}}&TRLRF&20.32&0.552&0.607&25.34&0.568&0.340&26.52&0.656&0.297&26.81&0.670&0.287&27.09&0.685&0.278\\
			&HTNN&29.94&0.790&0.200&32.92&0.844&0.142&34.70&0.873&0.116&36.28&0.890&0.096&37.41&0.894&0.085\\
			(256$\times$256$\times$31)&LTNN&29.74&0.784&0.205&32.95&0.840&0.142&34.81&0.869&0.114&36.32&0.886&0.096&37.52&0.898&0.083\\
			&SVDinsTN&\underline{35.03}&\underline{0.887}&\underline{0.111}&\underline{37.28}&\underline{0.890}&\underline{0.086}&\underline{38.18}&\underline{0.901}&\underline{0.077}&\underline{40.22}&\underline{0.930}&\underline{0.061}&\underline{41.63}&\underline{0.940}&\underline{0.052}\\
			
			&I-TSS&\textbf{37.86}&\textbf{0.894}&\textbf{0.080}&\textbf{40.78}&\textbf{0.936}&\textbf{0.057}&\textbf{42.70}&\textbf{0.939}&\textbf{0.046}&\textbf{44.35}&\textbf{0.942}&\textbf{0.038}&\textbf{45.64}&\textbf{0.955}&\textbf{0.032}\\
			\midrule
			&HaLRTC&33.03&0.952&0.065&36.10&0.972&0.045&38.69&0.982&0.034&40.77&0.988&0.026&42.42&0.991&0.022\\
			&SiLRTCTT&31.22&0.925&0.080&34.00&0.952&0.058&36.29&0.967&0.045&38.07&0.975&0.036&39.61&0.981&0.030\\
			\multirow{2}{*}{\emph{Beers}}&TRLRF&34.83&0.945&0.053&36.69&0.960&0.042&38.19&0.969&0.036&39.56&0.976&0.030&41.10&0.982&0.025\\
			&HTNN&32.36&0.928&0.070&39.73&\underline{0.983}&0.030&42.32&\underline{0.990}&0.022&\underline{44.45}&\underline{0.993}&\underline{0.017}&\underline{46.08}&\underline{0.995}&\underline{0.014}\\
			(256$\times$256$\times$31)&LTNN&33.53&0.934&0.061&38.29&0.977&0.035&41.34&0.987&0.025&43.84&\underline{0.993}&0.018&45.89&\underline{0.995}&\underline{0.014}\\
			&SVDinsTN&\underline{37.99}&\underline{0.973}&\underline{0.036}&\underline{40.60}&\underline{0.983}&\underline{0.027}&\underline{42.72}&0.988&\underline{0.021}&43.88&0.990&0.018&45.28&0.992&0.015\\
			&I-TSS&\textbf{43.17}&\textbf{0.991}&\textbf{0.020}&\textbf{46.28}&\textbf{0.995}&\textbf{0.014}&\textbf{48.79}&\textbf{0.997}&\textbf{0.010}&\textbf{49.91}&\textbf{0.998}&\textbf{0.009}&\textbf{50.54}&\textbf{0.998}&\textbf{0.008}\\
			\midrule
			&HaLRTC&31.64&0.932&0.120&34.91&0.959&0.082&37.55&0.975&0.061&39.60&0.983&0.048&41.61&0.988&0.038\\
			&SiLRTCTT&30.75&0.910&0.133&33.81&0.941&0.093&36.01&0.958&0.072&37.89&0.968&0.058&39.53&0.976&0.048\\
			\multirow{2}{*}{\emph{Balloons}}&TRLRF&34.03&0.919&0.091&35.89&0.936&0.073&37.93&0.959&0.058&39.16&0.967&0.050&40.37&0.974&0.044\\
			&HTNN&34.91&0.943&0.082&38.75&0.973&0.053&41.19&0.984&0.040&43.21&\underline{0.990}&0.031&\underline{45.19}&\underline{0.993}&\underline{0.025}\\
			(256$\times$256$\times$31)&LTNN&34.22&0.931&0.089&38.15&0.970&0.057&40.71&0.983&0.042&43.02&\underline{0.990}&0.032&\underline{45.19}&\underline{0.993}&\underline{0.025}\\
			&SVDinsTN&\underline{37.55}&\underline{0.964}&\underline{0.061}&\underline{40.18}&\underline{0.978}&\underline{0.045}&\underline{42.08}&\underline{0.985}&\underline{0.036}&\underline{43.61}&0.988&\underline{0.030}&45.10&0.991&\underline{0.025}\\
			
			&I-TSS&\textbf{44.46}&\textbf{0.991}&\textbf{0.027}&\textbf{48.23}&\textbf{0.996}&\textbf{0.017}&\textbf{49.49}&\textbf{0.996}&\textbf{0.015}&\textbf{51.48}&\textbf{0.997}&\textbf{0.012}&\textbf{52.76}&\textbf{0.998}&\textbf{0.010}\\
			\midrule
			&HaLRTC&20.49&0.695&0.450&27.74&0.862&0.195&29.61&0.896&0.157&31.31&0.920&0.129&32.91&0.938&0.107\\
			&SiLRTCTT&28.78&0.858&0.173&32.14&0.910&0.117&34.49&0.938&0.089&36.40&0.954&0.072&38.01&0.965&0.059\\
			\multirow{2}{*}{\emph{Feathers}}&TRLRF&31.28&0.837&0.130&33.43&0.889&0.101&35.19&0.922&0.082&36.35&0.938&0.072&36.99&0.943&0.067\\
			&HTNN&28.28&0.830&0.183&32.58&0.910&0.111&34.70&0.940&0.087&36.36&0.957&0.072&37.83&0.969&0.061\\
			(256$\times$256$\times$31)&LTNN&29.08&0.837&0.167&32.74&0.912&0.109&34.79&0.943&0.086&36.41&0.959&0.072&37.88&0.970&0.060\\
			&SVDinsTN&\underline{34.46}&\underline{0.936}&\underline{0.090}&\underline{37.23}&\underline{0.961}&\underline{0.065}&\underline{38.83}&\underline{0.971}&\underline{0.054}&\underline{40.17}&\underline{0.976}&\underline{0.046}&\underline{41.18}&\underline{0.980}&\underline{0.041}\\
			
			&I-TSS&\textbf{37.19}&\textbf{0.956}&\textbf{0.065}&\textbf{39.42}&\textbf{0.974}&\textbf{0.050}&\textbf{42.54}&\textbf{0.984}&\textbf{0.035}&\textbf{43.86}&\textbf{0.988}&\textbf{0.030}&\textbf{46.63}&\textbf{0.992}&\textbf{0.022}\\
			\bottomrule
	\end{tabular}}
\end{table*}
\begin{figure*}[!h]
	\centering
	\subfloat{
		\begin{minipage}[b]{0.1\linewidth}
			\captionsetup{
				font={small}, 
				labelfont=bf,        % 标签字体为粗体
				textfont={rm}, % 正文字体为 Times new roman
				singlelinecheck=true % 不允许换行 
			}
			\includegraphics[width=1\linewidth]{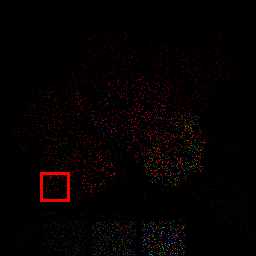}\vspace{1pt}
			\put(-18.8,0){\includegraphics[scale=0.75]{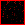}}\\
			\includegraphics[width=1\linewidth]{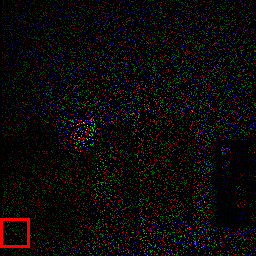}\vspace{1pt}
			\put(-18.8,0){\includegraphics[scale=0.75]{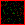}}\\
			\includegraphics[width=1\linewidth]{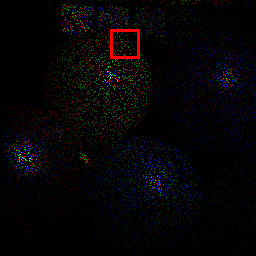}\vspace{1pt}
			\put(-18.8,0){\includegraphics[scale=0.75]{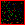}}\\
			\includegraphics[width=1\linewidth]{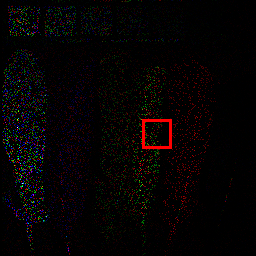}\vspace{1pt}
			\put(-18.8,0){\includegraphics[scale=0.75]{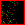}}
			\caption*{\fontfamily{Times New Roman}{Observed}}
	\end{minipage}}
	\hspace{-1.7mm}
	\subfloat{
		\begin{minipage}[b]{0.1\linewidth}
			\captionsetup{
				font={small},
				labelfont=bf,        % 标签字体为粗体
				textfont={rm}, % 正文字体为 Times new roman
				singlelinecheck=true % 不允许换行 
			}
			\includegraphics[width=1\linewidth]{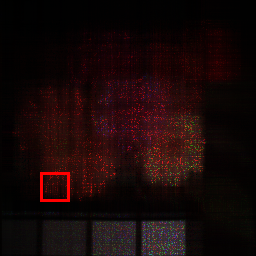}\vspace{1pt}
			\put(-18.8,0){\includegraphics[scale=0.75]{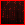}}\\
			\includegraphics[width=1\linewidth]{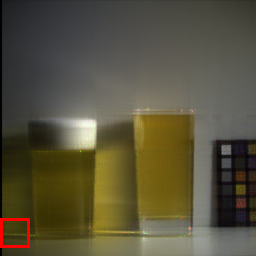}\vspace{1pt}
			\put(-18.8,0){\includegraphics[scale=0.75]{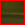}}\\
			\includegraphics[width=1\textwidth]{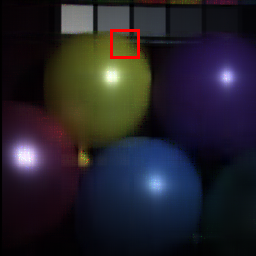}\vspace{1pt}
			\put(-18.8,0){\includegraphics[scale=0.75]{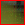}}\\
			\includegraphics[width=1\linewidth]{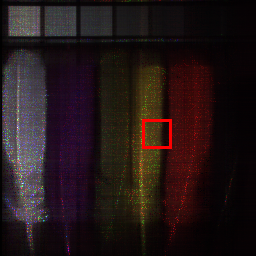}\vspace{1pt}
			\put(-18.8,0){\includegraphics[scale=0.75]{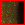}}
			\caption*{\fontfamily{Times New Roman}{HaLRTC}}
	\end{minipage}}
	\hspace{-1.7mm}
	\subfloat{
		\begin{minipage}[b]{0.1\linewidth}
			\captionsetup{
				font={small},
				labelfont=bf,        % 标签字体为粗体
				textfont={rm}, % 正文字体为 Times new roman
				singlelinecheck=true % 不允许换行 
			}
			\includegraphics[width=1\linewidth]{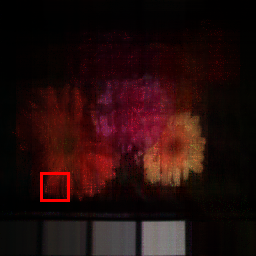}\vspace{1pt}
			\put(-18.8,0){\includegraphics[scale=0.75]{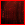}}\\
			\includegraphics[width=1\linewidth]{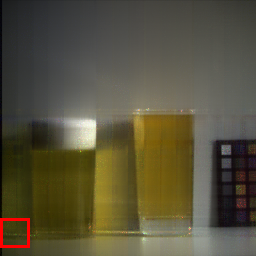}\vspace{1pt}
			\put(-18.8,0){\includegraphics[scale=0.75]{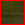}}\\
			\includegraphics[width=1\linewidth]{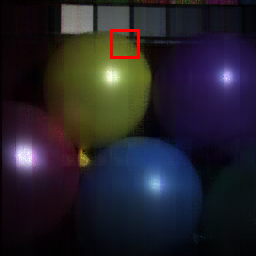}\vspace{1pt}
			\put(-18.8,0){\includegraphics[scale=0.75]{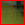}}\\
			\includegraphics[width=1\linewidth]{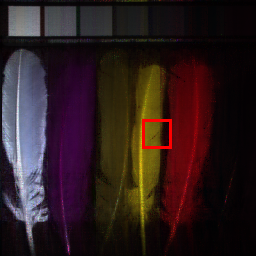}\vspace{1pt}
			\put(-18.8,0){\includegraphics[scale=0.75]{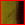}}
			\caption*{\fontfamily{Times New Roman}{SiLRTCTT}}
	\end{minipage}}
	\hspace{-1.7mm}
	\subfloat{
		\begin{minipage}[b]{0.1\linewidth}
			\captionsetup{
				font={small},
				labelfont=bf,        % 标签字体为粗体
				textfont={rm}, % 正文字体为 Times new roman
				singlelinecheck=true % 不允许换行 
			}
			\includegraphics[width=1\linewidth]{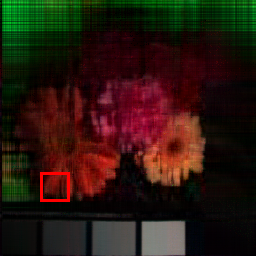}\vspace{1pt}
			\put(-18.8,0){\includegraphics[scale=0.75]{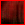}}\\
			\includegraphics[width=1\linewidth]{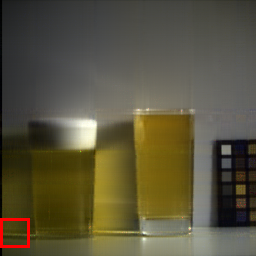}\vspace{1pt}
			\put(-18.8,0){\includegraphics[scale=0.75]{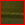}}\\
			\includegraphics[width=1\linewidth]{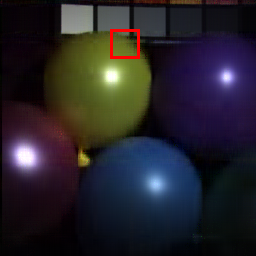}\vspace{1pt}
			\put(-18.8,0){\includegraphics[scale=0.75]{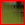}}\\
			\includegraphics[width=1\linewidth]{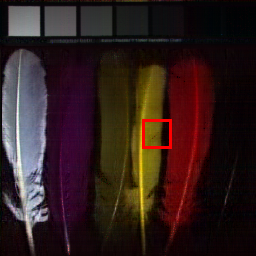}\vspace{1pt}
			\put(-18.8,0){\includegraphics[scale=0.75]{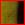}}
			\caption*{\fontfamily{Times New Roman}{TRLRF}}
	\end{minipage}}
	\hspace{-1.7mm}
	\subfloat{
		\begin{minipage}[b]{0.1\linewidth}
			\captionsetup{
				font={small},
				labelfont=bf,        % 标签字体为粗体
				textfont={rm}, % 正文字体为 Times new roman
				singlelinecheck=true % 不允许换行 
			}
			\includegraphics[width=1\linewidth]{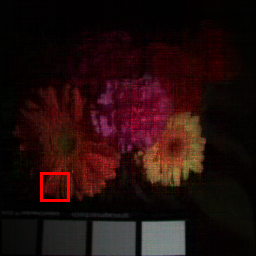}\vspace{1pt}
			\put(-18.8,0){\includegraphics[scale=0.75]{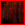}}\\
			\includegraphics[width=1\linewidth]{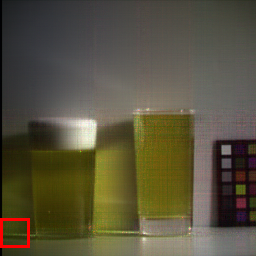}\vspace{1pt}
			\put(-18.8,0){\includegraphics[scale=0.75]{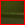}}\\
			\includegraphics[width=1\linewidth]{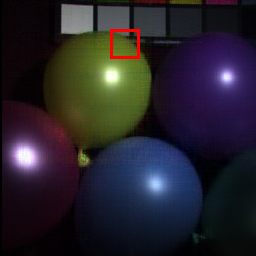}\vspace{1pt}
			\put(-18.8,0){\includegraphics[scale=0.75]{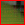}}\\
			\includegraphics[width=1\linewidth]{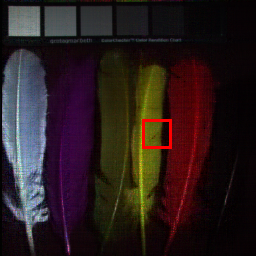}\vspace{1pt}
			\put(-18.8,0){\includegraphics[scale=0.75]{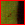}}
			\caption*{\fontfamily{Times New Roman}{HTNN}}
	\end{minipage}}
	\hspace{-1.7mm}
	\subfloat{
		\begin{minipage}[b]{0.1\linewidth}
			\captionsetup{
				font={small},
				labelfont=bf,        % 标签字体为粗体
				textfont={rm}, % 正文字体为 Times new roman
				singlelinecheck=true % 不允许换行 
			}
			\includegraphics[width=1\linewidth]{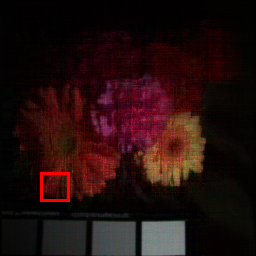}\vspace{1pt}
			\put(-18.8,0){\includegraphics[scale=0.75]{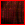}}\\
			\includegraphics[width=1\linewidth]{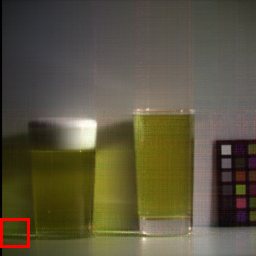}\vspace{1pt}
			\put(-18.8,0){\includegraphics[scale=0.75]{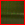}}\\
			\includegraphics[width=1\linewidth]{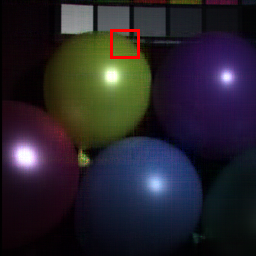}\vspace{1pt}
			\put(-18.8,0){\includegraphics[scale=0.75]{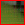}}\\
			\includegraphics[width=1\linewidth]{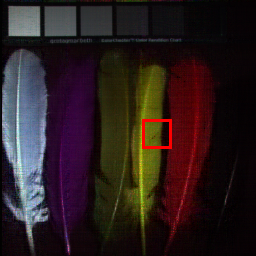}\vspace{1pt}
			\put(-18.8,0){\includegraphics[scale=0.75]{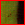}}
			\caption*{\fontfamily{Times New Roman}{LTNN}}
	\end{minipage}}
	\hspace{-1.7mm}
	\subfloat{
		\begin{minipage}[b]{0.1\linewidth}
			\captionsetup{
				font={small},
				labelfont=bf,        % 标签字体为粗体
				textfont={rm}, % 正文字体为 Times new roman
				singlelinecheck=true % 不允许换行 
			}
			\includegraphics[width=1\linewidth]{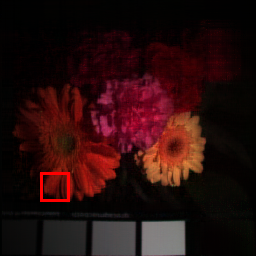}\vspace{1pt}
			\put(-18.8,0){\includegraphics[scale=0.75]{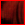}}\\
			\includegraphics[width=1\linewidth]{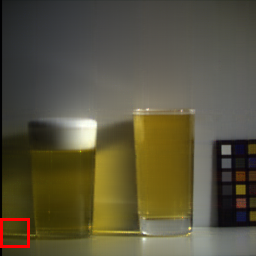}\vspace{1pt}
			\put(-18.8,0){\includegraphics[scale=0.75]{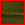}}\\
			\includegraphics[width=1\linewidth]{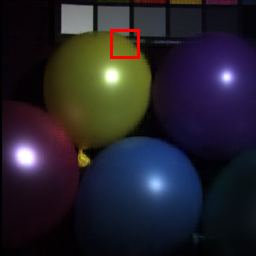}\vspace{1pt}
			\put(-18.8,0){\includegraphics[scale=0.75]{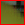}}\\
			\includegraphics[width=1\linewidth]{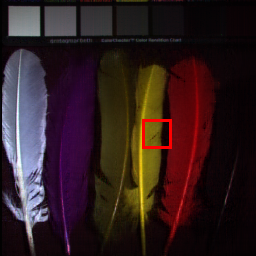}\vspace{1pt}
			\put(-18.8,0){\includegraphics[scale=0.75]{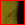}}
			\caption*{\fontfamily{Times New Roman}{SVDinsTN}}
	\end{minipage}}
	\hspace{-1.7mm}
	\subfloat{
		\begin{minipage}[b]{0.1\linewidth}
			\captionsetup{
				font={small},
				labelfont=bf,        % 标签字体为粗体
				textfont={rm}, % 正文字体为 Times new roman
				singlelinecheck=true % 不允许换行 
			}
			\includegraphics[width=1\linewidth]{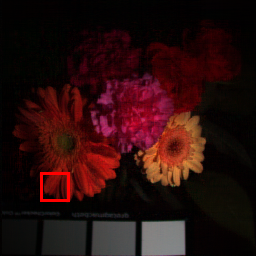}\vspace{1pt}
			\put(-18.8,0){\includegraphics[scale=0.75]{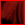}}\\
			\includegraphics[width=1\linewidth]{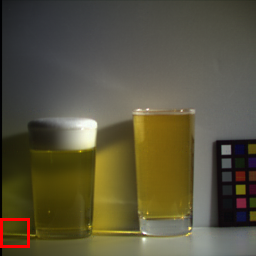}\vspace{1pt}
			\put(-18.8,0){\includegraphics[scale=0.75]{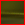}}\\
			\includegraphics[width=1\linewidth]{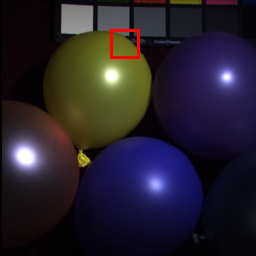}\vspace{1pt}
			\put(-18.8,0){\includegraphics[scale=0.75]{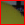}}\\
			\includegraphics[width=1\linewidth]{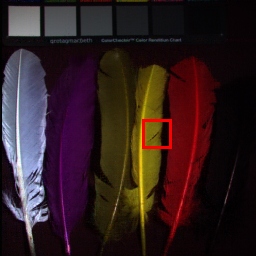}\vspace{1pt}
			\put(-18.8,0){\includegraphics[scale=0.75]{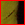}}
			\caption*{\fontfamily{Times New Roman}{I-TSS}}
	\end{minipage}}
	\hspace{-1.7mm}
	\subfloat{
		\begin{minipage}[b]{0.1\linewidth}
			\captionsetup{
				font={small},
				labelfont=bf,        % 标签字体为粗体
				textfont={rm}, % 正文字体为 Times new roman
				singlelinecheck=true % 不允许换行 
			}
			\includegraphics[width=1\linewidth]{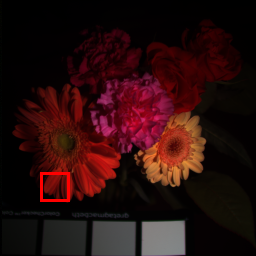}\vspace{1pt}
			\put(-18.8,0){\includegraphics[scale=0.75]{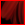}}\\
			\includegraphics[width=1\linewidth]{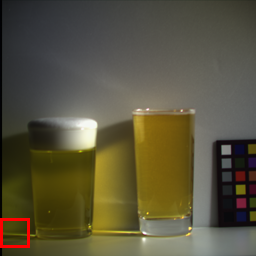}\vspace{1pt}
			\put(-18.8,0){\includegraphics[scale=0.75]{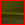}}\\
			\includegraphics[width=1\linewidth]{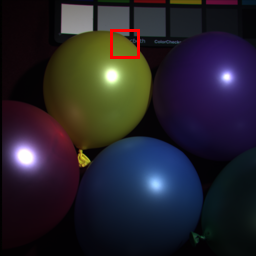}\vspace{1pt}
			\put(-18.8,0){\includegraphics[scale=0.75]{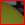}}\\
			\includegraphics[width=1\linewidth]{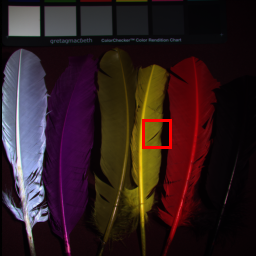}\vspace{1pt}
			\put(-18.8,0){\includegraphics[scale=0.75]{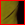}}
			\caption*{\fontfamily{Times New Roman}{Original}}
	\end{minipage}}
	\caption{The completion results of MSIs by different methods on \emph{Flowers}, \emph{Beers}, \emph{Balloons}, and \emph{Feathers} with SR = 0.1.}	
	\label{visual1}
\end{figure*}

\begin{table*}[!h]\scriptsize
	\centering
	
	\caption{The quantitative results for real-world color video completion by different methods. The \textbf{best} and \underline{second-best} values are highlighted.}
	\label{result4}
	\setlength{\tabcolsep}{1.9mm}{
		\begin{tabular}{ccccccccccccccccc}
			\toprule
			\multicolumn{2}{ c }{SR}&\multicolumn{3}{ c }{0.1 }&\multicolumn{3}{ c }{0.15 }&\multicolumn{3}{ c }{0.2}&\multicolumn{3}{ c }{0.25}&\multicolumn{3}{ c }{0.3}\\
			\midrule
			Data&Method&PSNR&SSIM&RE&PSNR&SSIM&RE&PSNR&SSIM&RE&PSNR&SSIM&RE&PSNR&SSIM&RE\\
			\midrule
			&HaLRTC&18.24&0.589&0.328&22.09&0.705&0.211&24.28&0.772&0.164&25.72&0.814&0.138&27.08&0.848&0.118\\
			&SiLRTCTT&27.18&0.851&0.117&29.84&0.903&0.086&31.81&0.931&0.068&33.36&0.947&0.057&34.67&0.959&0.049\\
			\multirow{2}{*}{\emph{News}}&TRLRF&29.75&0.861&0.087&30.11&0.863&0.083&31.02&0.882&0.075&33.10&0.919&0.059&34.82&0.942&0.048\\
			&HTNN&\underline{31.70}&\underline{0.916}&\underline{0.069}&\underline{33.53}&\underline{0.939}&\underline{0.056}&34.94&\underline{0.953}&\underline{0.048}&\underline{36.13}&0.962&\underline{0.041}&\underline{37.35}&\underline{0.970}&\underline{0.036}\\
			(144$\times$176$\times$3$\times$30)&LTNN&27.24&0.860&0.116&31.29&0.929&0.073&33.54&0.952&0.056&35.13&\underline{0.964}&0.047&36.58&\textbf{0.972}&0.039\\
			&SVDinsTN&31.58&0.897&0.070&33.16&0.922&0.058&\underline{34.95}&0.945&\underline{0.048}&36.07&0.958&0.042&37.16&0.967&0.037\\
			&I-TSS&\textbf{33.13}&\textbf{0.924}&\textbf{0.059}&\textbf{34.98}&\textbf{0.946}&\textbf{0.047}&\textbf{36.60}&\textbf{0.960}&\textbf{0.039}&\textbf{37.47}&\textbf{0.965}&\textbf{0.035}&\textbf{38.60}&\textbf{0.972}&\textbf{0.031}\\
			\midrule
			&HaLRTC&19.86&0.632&0.214&27.00&0.859&0.094&29.58&0.897&0.070&31.11&0.918&0.058&32.60&0.935&0.049\\
			&SiLRTCTT&29.74&0.884&0.068&32.42&0.921&0.050&34.34&0.941&0.040&35.87&0.954&0.034&37.21&0.962&0.029\\
			\multirow{2}{*}{\emph{Claire}}&TRLRF&33.67&0.922&0.043&35.08&0.935&0.037&36.06&0.945&0.033&38.05&0.963&0.026&38.42&0.964&0.025\\
			&HTNN&34.42&\underline{0.940}&0.040&36.33&0.955&0.032&37.81&\underline{0.965}&0.027&39.23&0.973&0.023&40.54&0.978&0.019\\
			(144$\times$176$\times$3$\times$30)&LTNN&32.13&0.922&0.052&35.31&0.952&0.036&37.27&0.963&0.028&39.21&0.974&0.023&40.91&\textbf{0.980}&0.019\\
			&SVDinsTN&\underline{34.73}&0.933&\underline{0.038}&\underline{37.12}&\underline{0.956}&\underline{0.029}&\underline{38.29}&0.964&\underline{0.025}&\underline{40.64}&\underline{0.975}&\underline{0.019}&\underline{41.65}&\underline{0.979}&\underline{0.017}\\
			&I-TSS&\textbf{36.19}&\textbf{0.947}&\textbf{0.032}&\textbf{38.00}&\textbf{0.961}&\textbf{0.026}&\textbf{39.20}&\textbf{0.967}&\textbf{0.023}&\textbf{41.15}&\textbf{0.976}&\textbf{0.018}&\textbf{42.00}&0.978&\textbf{0.016}\\
			\midrule
			&HaLRTC&23.82&0.705&0.236&27.22&0.793&0.159&29.04&0.838&0.129&30.62&0.872&0.107&32.13&0.899&0.090\\
			&SiLRTCTT&30.57&0.885&0.108&33.52&0.926&0.077&35.52&0.946&0.061&37.08&0.958&0.051&38.39&0.965&0.044\\
			\multirow{2}{*}{\emph{Grandma}}&TRLRF&32.88&0.879&0.083&34.46&0.908&0.069&36.44&0.937&0.055&37.83&0.953&0.047&38.26&0.956&0.044\\
			&HTNN&\underline{36.65}&\underline{0.946}&\underline{0.053}&38.16&0.959&0.045&39.35&0.967&0.039&40.31&0.973&0.035&41.34&\underline{0.977}&\underline{0.031}\\
			(144$\times$176$\times$3$\times$30)&LTNN&33.40&0.928&0.078&37.96&\underline{0.963}&0.046&\underline{39.70}&\underline{0.971}&\underline{0.037}&\underline{40.60}&\underline{0.974}&\underline{0.034}&\underline{41.42}&\underline{0.977}&\underline{0.031}\\
			&SVDinsTN&36.48&0.939&0.054&\underline{38.26}&0.957&\underline{0.044}&39.03&0.963&0.041&40.33&0.970&0.035&41.40&0.976&\underline{0.031}\\
			&I-TSS&\textbf{38.97}&\textbf{0.961}&\textbf{0.044}&\textbf{39.83}&\textbf{0.967}&\textbf{0.037}&\textbf{40.94}&\textbf{0.973}&\textbf{0.032}&\textbf{41.57}&\textbf{0.976}&\textbf{0.030}&\textbf{42.05}&\textbf{0.978}&\textbf{0.028}\\
			\midrule
			&HaLRTC&18.39&0.576&0.263&24.97&0.762&0.123&27.62&0.829&0.091&29.12&0.864&0.076&30.54&0.892&0.065\\
			&SiLRTCTT&29.49&0.876&0.073&32.41&0.922&0.052&34.85&0.946&0.040&36.32&0.960&0.033&37.70&0.969&0.028\\
			\multirow{2}{*}{\emph{Akiyo}}&TRLRF&32.44&0.895&0.052&33.40&0.909&0.046&34.44&0.925&0.041&36.83&0.950&0.031&37.12&0.952&0.030\\
			&HTNN&35.18&\underline{0.947}&0.038&37.34&\underline{0.964}&0.029&\underline{38.95}&\underline{0.973}&\underline{0.024}&40.46&\underline{0.979}&\underline{0.020}&41.77&\textbf{0.984}&\underline{0.017}\\
			(144$\times$176$\times$3$\times$30)&LTNN&32.27&0.902&0.059&36.22&0.960&0.033&38.61&\underline{0.973}&0.025&\underline{40.69}&\textbf{0.981}&\underline{0.020}&41.89&\textbf{0.984}&\underline{0.017}\\
			&SVDinsTN&\underline{35.90}&0.945&\underline{0.035}&\underline{38.02}&0.962&\underline{0.027}&38.73&0.967&0.025&40.33&0.977&0.021&\underline{42.02}&\underline{0.983}&\underline{0.017}\\
			&I-TSS&\textbf{37.32}&\textbf{0.957}&\textbf{0.029}&\textbf{39.21}&\textbf{0.969}&\textbf{0.024}&\textbf{40.99}&\textbf{0.977}&\textbf{0.019}&\textbf{41.95}&\textbf{0.981}&\textbf{0.017}&\textbf{43.12}&\textbf{0.984}&\textbf{0.015}\\
			\bottomrule
	\end{tabular}}
\end{table*}

\begin{figure*}[!h]
	\centering
	\subfloat{
		\begin{minipage}[b]{0.1\linewidth}
			\captionsetup{
				font={small}, 
				labelfont=bf,        % 标签字体为粗体
				textfont={rm}, % 正文字体为 Times new roman
				singlelinecheck=true % 不允许换行 
			}
			\includegraphics[width=1\linewidth]{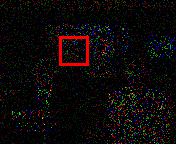}\vspace{1pt}
			\put(-18.8,0){\includegraphics[scale=0.75]{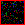}}\\
			\includegraphics[width=1\linewidth]{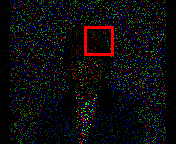}\vspace{1pt}
			\put(-18.8,0){\includegraphics[scale=0.75]{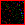}}\\
			\includegraphics[width=1\linewidth]{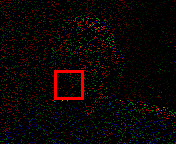}\vspace{1pt}
			\put(-18.8,0){\includegraphics[scale=0.75]{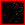}}\\
			\includegraphics[width=1\linewidth]{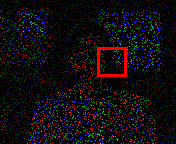}\vspace{1pt}
			\put(-18.8,0){\includegraphics[scale=0.75]{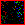}}
			\caption*{\fontfamily{Times New Roman}{Observed}}
	\end{minipage}}
	\hspace{-1.7mm}
	\subfloat{
		\begin{minipage}[b]{0.1\linewidth}
			\captionsetup{
				font={small},
				labelfont=bf,        % 标签字体为粗体
				textfont={rm}, % 正文字体为 Times new roman
				singlelinecheck=true % 不允许换行 
			}
			\includegraphics[width=1\linewidth]{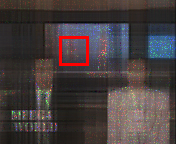}\vspace{1pt}
			\put(-18.8,0){\includegraphics[scale=0.75]{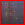}}\\
			\includegraphics[width=1\linewidth]{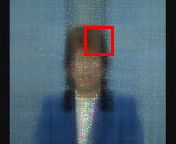}\vspace{1pt}
			\put(-18.8,0){\includegraphics[scale=0.75]{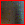}}\\
			\includegraphics[width=1\textwidth]{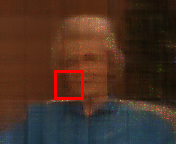}\vspace{1pt}
			\put(-18.8,0){\includegraphics[scale=0.75]{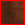}}\\
			\includegraphics[width=1\linewidth]{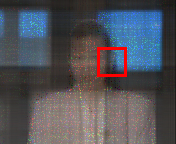}\vspace{1pt}
			\put(-18.8,0){\includegraphics[scale=0.75]{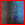}}
			\caption*{\fontfamily{Times New Roman}{HaLRTC}}
	\end{minipage}}
	\hspace{-1.7mm}
	\subfloat{
		\begin{minipage}[b]{0.1\linewidth}
			\captionsetup{
				font={small},
				labelfont=bf,        % 标签字体为粗体
				textfont={rm}, % 正文字体为 Times new roman
				singlelinecheck=true % 不允许换行 
			}
			\includegraphics[width=1\linewidth]{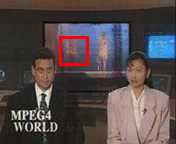}\vspace{1pt}
			\put(-18.8,0){\includegraphics[scale=0.75]{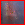}}\\
			\includegraphics[width=1\linewidth]{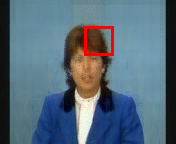}\vspace{1pt}
			\put(-18.8,0){\includegraphics[scale=0.75]{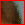}}\\
			\includegraphics[width=1\linewidth]{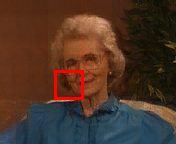}\vspace{1pt}
			\put(-18.8,0){\includegraphics[scale=0.75]{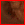}}\\
			\includegraphics[width=1\linewidth]{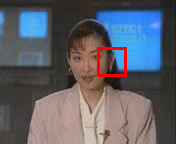}\vspace{1pt}
			\put(-18.8,0){\includegraphics[scale=0.75]{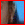}}
			\caption*{\fontfamily{Times New Roman}{SiLRTCTT}}
	\end{minipage}}
	\hspace{-1.7mm}
	\subfloat{
		\begin{minipage}[b]{0.1\linewidth}
			\captionsetup{
				font={small},
				labelfont=bf,        % 标签字体为粗体
				textfont={rm}, % 正文字体为 Times new roman
				singlelinecheck=true % 不允许换行 
			}
			\includegraphics[width=1\linewidth]{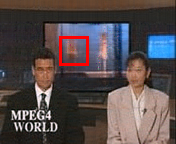}\vspace{1pt}
			\put(-18.8,0){\includegraphics[scale=0.75]{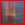}}\\
			\includegraphics[width=1\linewidth]{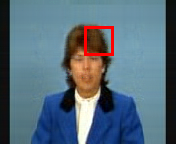}\vspace{1pt}
			\put(-18.8,0){\includegraphics[scale=0.75]{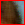}}\\
			\includegraphics[width=1\linewidth]{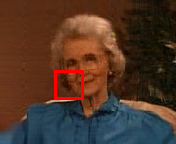}\vspace{1pt}
			\put(-18.8,0){\includegraphics[scale=0.75]{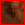}}\\
			\includegraphics[width=1\linewidth]{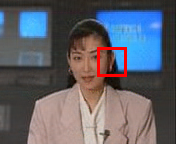}\vspace{1pt}
			\put(-18.8,0){\includegraphics[scale=0.75]{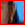}}
			\caption*{\fontfamily{Times New Roman}{TRLRF}}
	\end{minipage}}
	\hspace{-1.7mm}
	\subfloat{
		\begin{minipage}[b]{0.1\linewidth}
			\captionsetup{
				font={small},
				labelfont=bf,        % 标签字体为粗体
				textfont={rm}, % 正文字体为 Times new roman
				singlelinecheck=true % 不允许换行 
			}
			\includegraphics[width=1\linewidth]{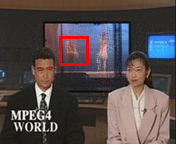}\vspace{1pt}
			\put(-18.8,0){\includegraphics[scale=0.75]{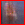}}\\
			\includegraphics[width=1\linewidth]{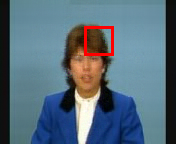}\vspace{1pt}
			\put(-18.8,0){\includegraphics[scale=0.75]{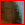}}\\
			\includegraphics[width=1\linewidth]{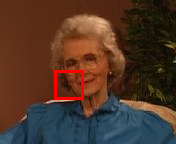}\vspace{1pt}
			\put(-18.8,0){\includegraphics[scale=0.75]{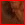}}\\
			\includegraphics[width=1\linewidth]{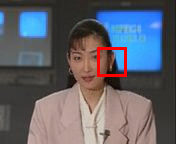}\vspace{1pt}
			\put(-18.8,0){\includegraphics[scale=0.75]{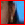}}
			\caption*{\fontfamily{Times New Roman}{HTNN}}
	\end{minipage}}
	\hspace{-1.7mm}
	\subfloat{
		\begin{minipage}[b]{0.1\linewidth}
			\captionsetup{
				font={small},
				labelfont=bf,        % 标签字体为粗体
				textfont={rm}, % 正文字体为 Times new roman
				singlelinecheck=true % 不允许换行 
			}
			\includegraphics[width=1\linewidth]{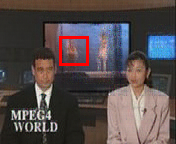}\vspace{1pt}
			\put(-18.8,0){\includegraphics[scale=0.75]{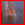}}\\
			\includegraphics[width=1\linewidth]{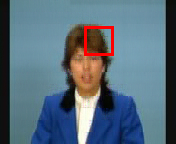}\vspace{1pt}
			\put(-18.8,0){\includegraphics[scale=0.75]{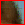}}\\
			\includegraphics[width=1\linewidth]{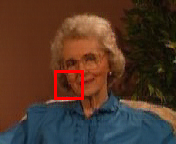}\vspace{1pt}
			\put(-18.8,0){\includegraphics[scale=0.75]{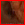}}\\
			\includegraphics[width=1\linewidth]{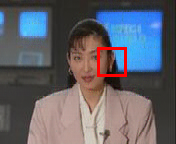}\vspace{1pt}
			\put(-18.8,0){\includegraphics[scale=0.75]{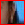}}
			\caption*{\fontfamily{Times New Roman}{LTNN}}
	\end{minipage}}
	\hspace{-1.7mm}
	\subfloat{
		\begin{minipage}[b]{0.1\linewidth}
			\captionsetup{
				font={small},
				labelfont=bf,        % 标签字体为粗体
				textfont={rm}, % 正文字体为 Times new roman
				singlelinecheck=true % 不允许换行 
			}
			\includegraphics[width=1\linewidth]{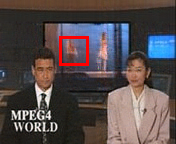}\vspace{1pt}
			\put(-18.8,0){\includegraphics[scale=0.75]{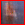}}\\
			\includegraphics[width=1\linewidth]{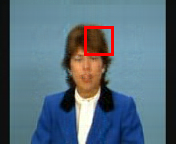}\vspace{1pt}
			\put(-18.8,0){\includegraphics[scale=0.75]{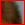}}\\
			\includegraphics[width=1\linewidth]{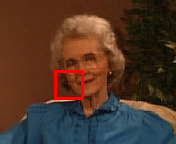}\vspace{1pt}
			\put(-18.8,0){\includegraphics[scale=0.75]{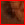}}\\
			\includegraphics[width=1\linewidth]{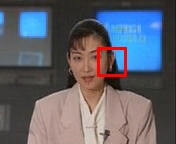}\vspace{1pt}
			\put(-18.8,0){\includegraphics[scale=0.75]{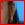}}
			\caption*{\fontfamily{Times New Roman}{SVDinsTN}}
	\end{minipage}}
	\hspace{-1.7mm}
	\subfloat{
		\begin{minipage}[b]{0.1\linewidth}
			\captionsetup{
				font={small},
				labelfont=bf,        % 标签字体为粗体
				textfont={rm}, % 正文字体为 Times new roman
				singlelinecheck=true % 不允许换行 
			}
			\includegraphics[width=1\linewidth]{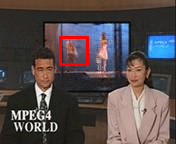}\vspace{1pt}
			\put(-18.8,0){\includegraphics[scale=0.75]{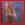}}\\
			\includegraphics[width=1\linewidth]{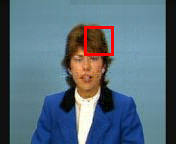}\vspace{1pt}
			\put(-18.8,0){\includegraphics[scale=0.75]{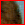}}\\
			\includegraphics[width=1\linewidth]{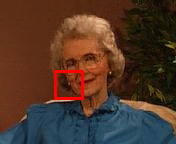}\vspace{1pt}
			\put(-18.8,0){\includegraphics[scale=0.75]{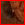}}\\
			\includegraphics[width=1\linewidth]{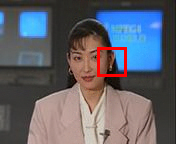}\vspace{1pt}
			\put(-18.8,0){\includegraphics[scale=0.75]{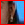}}
			\caption*{\fontfamily{Times New Roman}{I-TSS}}
	\end{minipage}}
	\hspace{-1.7mm}
	\subfloat{
		\begin{minipage}[b]{0.1\linewidth}
			\captionsetup{
				font={small},
				labelfont=bf,        % 标签字体为粗体
				textfont={rm}, % 正文字体为 Times new roman
				singlelinecheck=true % 不允许换行 
			}
			\includegraphics[width=1\linewidth]{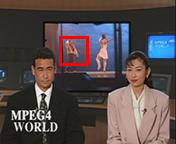}\vspace{1pt}
			\put(-18.8,0){\includegraphics[scale=0.75]{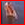}}\\
			\includegraphics[width=1\linewidth]{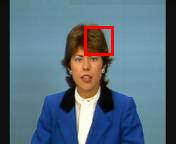}\vspace{1pt}
			\put(-18.8,0){\includegraphics[scale=0.75]{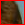}}\\
			\includegraphics[width=1\linewidth]{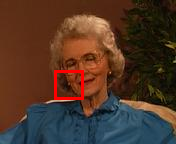}\vspace{1pt}
			\put(-18.8,0){\includegraphics[scale=0.75]{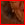}}\\
			\includegraphics[width=1\linewidth]{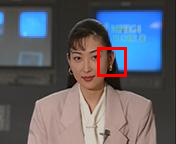}\vspace{1pt}
			\put(-18.8,0){\includegraphics[scale=0.75]{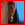}}
			\caption*{\fontfamily{Times New Roman}{Original}}
	\end{minipage}}
	\caption{The completion results of real-world color videos by different methods on \emph{News}, \emph{Claire}, \emph{Grandma}, and \emph{Akiyo} with SR = 0.1.}	
	\label{visual2}
\end{figure*}

\begin{table*}[!h]\scriptsize
	\centering
	
	\caption{The quantitative results for real-world light field data completion by different methods. The \textbf{best} and \underline{second-best} values are highlighted.}
	\label{result5}
	\setlength{\tabcolsep}{1.9mm}{
		\begin{tabular}{ccccccccccccccccc}
			\toprule
			\multicolumn{2}{ c }{SR}&\multicolumn{3}{ c }{0.1 }&\multicolumn{3}{ c }{0.15 }&\multicolumn{3}{ c }{0.2}&\multicolumn{3}{ c }{0.25}&\multicolumn{3}{ c }{0.3}\\
			\midrule
			Data&Method&PSNR&SSIM&RE&PSNR&SSIM&RE&PSNR&SSIM&RE&PSNR&SSIM&RE&PSNR&SSIM&RE\\
			\midrule
			&HaLRTC&17.16&0.524&0.247&22.78&0.737&0.129&25.10&0.815&0.099&26.57&0.856&0.083&27.94&0.888&0.071\\
			&SiLRTCTT&27.53&0.880&0.074&28.18&0.887&0.069&30.07&0.921&0.055&31.67&0.943&0.046&33.15&0.957&0.039\\
			\multirow{2}{*}{\emph{Greek}}&TRLRF&24.37&0.715&0.107&28.70&0.869&0.065&29.05&0.876&0.062&29.41&0.887&0.060&29.76&0.890&0.057\\
			&HTNN&29.06&0.890&0.062&31.07&0.927&0.049&32.95&0.950&0.040&34.67&0.965&0.032&36.39&0.975&0.027\\
			(128$\times$128$\times$3$\times$30)&LTNN&29.37&\underline{0.909}&0.060&32.28&\underline{0.952}&0.043&\underline{34.86}&\underline{0.972}&\underline{0.032}&36.85&\underline{0.982}&0.025&38.62&\underline{0.987}&0.020\\
			&SVDinsTN&\underline{30.94}&0.907&\underline{0.050}&\underline{32.96}&0.936&\underline{0.040}&34.37&0.955&0.034&\underline{37.26}&0.977&\underline{0.024}&\underline{39.42}&0.984&\underline{0.019}\\
			&I-TSS&\textbf{32.52}&\textbf{0.943}&\textbf{0.042}&\textbf{35.04}&\textbf{0.962}&\textbf{0.031}&\textbf{37.72}&\textbf{0.979}&\textbf{0.023}&\textbf{39.77}&\textbf{0.986}&\textbf{0.018}&\textbf{40.99}&\textbf{0.989}&\textbf{0.015}\\
			\midrule
			&HaLRTC&21.71&0.694&0.158&27.59&0.866&0.080&29.44&0.899&0.065&30.93&0.921&0.054&32.31&0.938&0.046\\
			&SiLRTCTT&29.95&0.901&0.061&30.93&0.921&0.054&33.08&0.944&0.042&34.70&0.958&0.035&36.12&0.967&0.030\\
			\multirow{2}{*}{\emph{Museum}}&TRLRF&29.35&0.847&0.065&33.32&0.926&0.041&34.88&0.942&0.034&35.64&0.950&0.031&36.92&0.961&0.027\\
			&HTNN&35.08&\underline{0.955}&0.034&36.90&0.968&0.027&38.47&0.977&0.022&39.89&0.983&0.019&41.22&\underline{0.987}&0.016\\
			(128$\times$128$\times$3$\times$30)&LTNN&32.60&0.939&0.045&36.77&\underline{0.972}&0.027&39.53&\underline{0.983}&0.020&\underline{41.62}&\underline{0.989}&\underline{0.016}&\underline{44.46}&\textbf{0.994}&\underline{0.011}\\
			&SVDinsTN&\underline{35.27}&0.947&\underline{0.033}&\underline{37.22}&0.964&\underline{0.026}&\underline{39.90}&0.978&\underline{0.019}&41.30&0.983&\underline{0.016}&42.19&0.985&0.014\\
			&I-TSS&\textbf{37.85}&\textbf{0.967}&\textbf{0.024}&\textbf{41.22}&\textbf{0.984}&\textbf{0.016}&\textbf{42.47}&\textbf{0.987}&\textbf{0.014}&\textbf{45.83}&\textbf{0.993}&\textbf{0.009}&\textbf{46.14}&\textbf{0.994}&\textbf{0.009}\\
			\midrule
			&HaLRTC&23.67&0.775&0.202&26.53&0.859&0.145&28.35&0.896&0.118&29.89&0.922&0.098&31.43&0.942&0.082\\
			&SiLRTCTT&29.23&0.914&0.106&30.62&0.933&0.090&32.56&0.954&0.072&34.05&0.965&0.061&35.48&0.974&0.051\\
			\multirow{2}{*}{\emph{Medieval2}}&TRLRF&29.09&0.892&0.108&31.79&0.937&0.079&33.72&0.960&0.063&35.35&0.971&0.052&36.78&0.979&0.044\\
			&HTNN&33.95&0.966&0.061&35.57&0.976&0.051&37.00&0.982&0.043&38.26&0.986&0.037&39.47&0.989&0.032\\
			(128$\times$128$\times$3$\times$30)&LTNN&32.76&0.958&0.071&37.63&\underline{0.987}&0.040&\underline{40.48}&\underline{0.992}&\underline{0.029}&\underline{42.25}&\underline{0.995}&\underline{0.023}&\underline{43.74}&\underline{0.996}&\underline{0.020}\\
			&SVDinsTN&\underline{35.84}&\underline{0.975}&\underline{0.049}&\underline{38.52}&0.986&\underline{0.036}&39.96&0.990&0.031&41.58&0.993&0.025&42.40&0.993&0.023\\
			&I-TSS&\textbf{36.11}&\textbf{0.977}&\textbf{0.048}&\textbf{39.96}&\textbf{0.991}&\textbf{0.031}&\textbf{42.02}&\textbf{0.994}&\textbf{0.024}&\textbf{44.88}&\textbf{0.996}&\textbf{0.017}&\textbf{45.61}&\textbf{0.997}&\textbf{0.016}\\
			\midrule
			&HaLRTC&23.30&0.722&0.228&26.26&0.826&0.162&28.16&0.876&0.130&30.07&0.912&0.104&31.73&0.936&0.086\\
			&SiLRTCTT&29.87&0.903&0.107&31.07&0.923&0.093&33.11&0.948&0.073&34.83&0.963&0.060&36.29&0.972&0.051\\
			\multirow{2}{*}{\emph{Vinyl}}&TRLRF&26.86&0.787&0.151&31.63&0.903&0.087&33.27&0.934&0.072&34.59&0.947&0.062&38.45&0.975&0.039\\
			&HTNN&34.68&\underline{0.958}&0.061&36.80&0.973&0.048&38.52&0.982&0.039&40.16&0.987&0.032&41.59&0.990&0.027\\
			(128$\times$128$\times$3$\times$30)&LTNN&32.99&0.946&0.074&39.60&\underline{0.986}&0.034&\underline{42.62}&\underline{0.993}&\underline{0.024}&\underline{44.70}&\underline{0.995}&\underline{0.019}&\underline{46.24}&\underline{0.996}&\underline{0.016}\\
			&SVDinsTN&\underline{34.87}&0.952&\underline{0.060}&\underline{40.12}&0.984&\underline{0.032}&42.14&0.990&0.026&43.81&0.993&0.021&45.37&0.995&0.017\\
			&I-TSS&\textbf{38.23}&\textbf{0.975}&\textbf{0.040}&\textbf{44.31}&\textbf{0.993}&\textbf{0.020}&\textbf{46.51}&\textbf{0.995}&\textbf{0.015}&\textbf{50.39}&\textbf{0.998}&\textbf{0.010}&\textbf{50.75}&\textbf{0.998}&\textbf{0.009}\\
			\bottomrule
	\end{tabular}}
\end{table*}

\begin{figure*}[!h]
	\centering
	\subfloat{
		\begin{minipage}[b]{0.1\linewidth}
			\captionsetup{
				font={small}, 
				labelfont=bf,        % 标签字体为粗体
				textfont={rm}, % 正文字体为 Times new roman
				singlelinecheck=true % 不允许换行 
			}
			\includegraphics[width=1\linewidth]{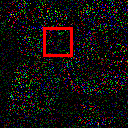}\vspace{1pt}
			\put(-18.8,0){\includegraphics[scale=0.75]{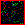}}\\
			\includegraphics[width=1\linewidth]{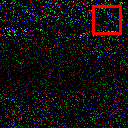}\vspace{1pt}
			\put(-18.8,0){\includegraphics[scale=0.75]{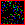}}\\
			\includegraphics[width=1\linewidth]{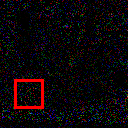}\vspace{1pt}
			\put(-18.8,0){\includegraphics[scale=0.75]{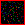}}\\
			\includegraphics[width=1\linewidth]{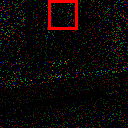}\vspace{1pt}
			\put(-18.8,0){\includegraphics[scale=0.75]{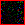}}
			\caption*{\fontfamily{Times New Roman}{Observed}}
	\end{minipage}}
	\hspace{-1.7mm}
	\subfloat{
		\begin{minipage}[b]{0.1\linewidth}
			\captionsetup{
				font={small},
				labelfont=bf,        % 标签字体为粗体
				textfont={rm}, % 正文字体为 Times new roman
				singlelinecheck=true % 不允许换行 
			}
			\includegraphics[width=1\linewidth]{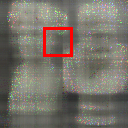}\vspace{1pt}
			\put(-18.8,0){\includegraphics[scale=0.75]{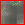}}\\
			\includegraphics[width=1\linewidth]{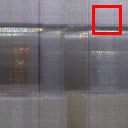}\vspace{1pt}
			\put(-18.8,0){\includegraphics[scale=0.75]{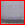}}\\
			\includegraphics[width=1\textwidth]{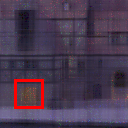}\vspace{1pt}
			\put(-18.8,0){\includegraphics[scale=0.75]{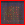}}\\
			\includegraphics[width=1\linewidth]{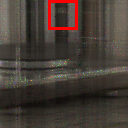}\vspace{1pt}
			\put(-18.8,0){\includegraphics[scale=0.75]{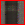}}
			\caption*{\fontfamily{Times New Roman}{HaLRTC}}
	\end{minipage}}
	\hspace{-1.7mm}
	\subfloat{
		\begin{minipage}[b]{0.1\linewidth}
			\captionsetup{
				font={small},
				labelfont=bf,        % 标签字体为粗体
				textfont={rm}, % 正文字体为 Times new roman
				singlelinecheck=true % 不允许换行 
			}
			\includegraphics[width=1\linewidth]{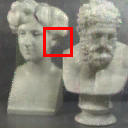}\vspace{1pt}
			\put(-18.8,0){\includegraphics[scale=0.75]{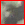}}\\
			\includegraphics[width=1\linewidth]{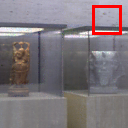}\vspace{1pt}
			\put(-18.8,0){\includegraphics[scale=0.75]{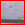}}\\
			\includegraphics[width=1\linewidth]{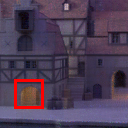}\vspace{1pt}
			\put(-18.8,0){\includegraphics[scale=0.75]{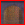}}\\
			\includegraphics[width=1\linewidth]{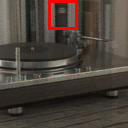}\vspace{1pt}
			\put(-18.8,0){\includegraphics[scale=0.75]{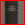}}
			\caption*{\fontfamily{Times New Roman}{SiLRTCTT}}
	\end{minipage}}
	\hspace{-1.7mm}
	\subfloat{
		\begin{minipage}[b]{0.1\linewidth}
			\captionsetup{
				font={small},
				labelfont=bf,        % 标签字体为粗体
				textfont={rm}, % 正文字体为 Times new roman
				singlelinecheck=true % 不允许换行 
			}
			\includegraphics[width=1\linewidth]{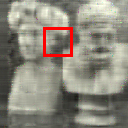}\vspace{1pt}
			\put(-18.8,0){\includegraphics[scale=0.75]{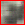}}\\
			\includegraphics[width=1\linewidth]{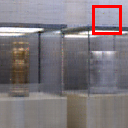}\vspace{1pt}
			\put(-18.8,0){\includegraphics[scale=0.75]{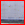}}\\
			\includegraphics[width=1\linewidth]{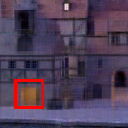}\vspace{1pt}
			\put(-18.8,0){\includegraphics[scale=0.75]{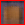}}\\
			\includegraphics[width=1\linewidth]{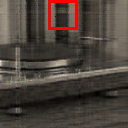}\vspace{1pt}
			\put(-18.8,0){\includegraphics[scale=0.75]{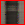}}
			\caption*{\fontfamily{Times New Roman}{TRLRF}}
	\end{minipage}}
	\hspace{-1.7mm}
	\subfloat{
		\begin{minipage}[b]{0.1\linewidth}
			\captionsetup{
				font={small},
				labelfont=bf,        % 标签字体为粗体
				textfont={rm}, % 正文字体为 Times new roman
				singlelinecheck=true % 不允许换行 
			}
			\includegraphics[width=1\linewidth]{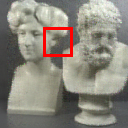}\vspace{1pt}
			\put(-18.8,0){\includegraphics[scale=0.75]{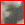}}\\
			\includegraphics[width=1\linewidth]{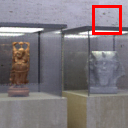}\vspace{1pt}
			\put(-18.8,0){\includegraphics[scale=0.75]{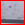}}\\
			\includegraphics[width=1\linewidth]{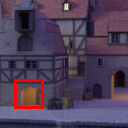}\vspace{1pt}
			\put(-18.8,0){\includegraphics[scale=0.75]{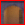}}\\
			\includegraphics[width=1\linewidth]{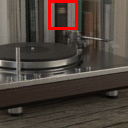}\vspace{1pt}
			\put(-18.8,0){\includegraphics[scale=0.75]{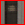}}
			\caption*{\fontfamily{Times New Roman}{HTNN}}
	\end{minipage}}
	\hspace{-1.7mm}
	\subfloat{
		\begin{minipage}[b]{0.1\linewidth}
			\captionsetup{
				font={small},
				labelfont=bf,        % 标签字体为粗体
				textfont={rm}, % 正文字体为 Times new roman
				singlelinecheck=true % 不允许换行 
			}
			\includegraphics[width=1\linewidth]{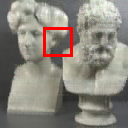}\vspace{1pt}
			\put(-18.8,0){\includegraphics[scale=0.75]{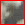}}\\
			\includegraphics[width=1\linewidth]{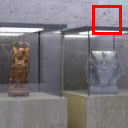}\vspace{1pt}
			\put(-18.8,0){\includegraphics[scale=0.75]{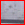}}\\
			\includegraphics[width=1\linewidth]{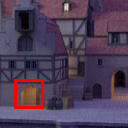}\vspace{1pt}
			\put(-18.8,0){\includegraphics[scale=0.75]{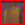}}\\
			\includegraphics[width=1\linewidth]{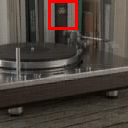}\vspace{1pt}
			\put(-18.8,0){\includegraphics[scale=0.75]{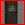}}
			\caption*{\fontfamily{Times New Roman}{LTNN}}
	\end{minipage}}
	\hspace{-1.7mm}
	\subfloat{
		\begin{minipage}[b]{0.1\linewidth}
			\captionsetup{
				font={small},
				labelfont=bf,        % 标签字体为粗体
				textfont={rm}, % 正文字体为 Times new roman
				singlelinecheck=true % 不允许换行 
			}
			\includegraphics[width=1\linewidth]{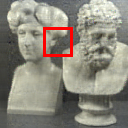}\vspace{1pt}
			\put(-18.8,0){\includegraphics[scale=0.75]{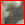}}\\
			\includegraphics[width=1\linewidth]{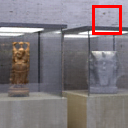}\vspace{1pt}
			\put(-18.8,0){\includegraphics[scale=0.75]{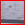}}\\
			\includegraphics[width=1\linewidth]{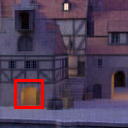}\vspace{1pt}
			\put(-18.8,0){\includegraphics[scale=0.75]{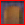}}\\
			\includegraphics[width=1\linewidth]{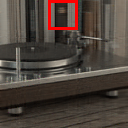}\vspace{1pt}
			\put(-18.8,0){\includegraphics[scale=0.75]{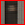}}
			\caption*{\fontfamily{Times New Roman}{SVDinsTN}}
	\end{minipage}}
	\hspace{-1.7mm}
	\subfloat{
		\begin{minipage}[b]{0.1\linewidth}
			\captionsetup{
				font={small},
				labelfont=bf,        % 标签字体为粗体
				textfont={rm}, % 正文字体为 Times new roman
				singlelinecheck=true % 不允许换行 
			}
			\includegraphics[width=1\linewidth]{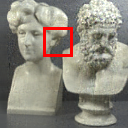}\vspace{1pt}
			\put(-18.8,0){\includegraphics[scale=0.75]{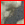}}\\
			\includegraphics[width=1\linewidth]{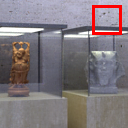}\vspace{1pt}
			\put(-18.8,0){\includegraphics[scale=0.75]{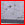}}\\
			\includegraphics[width=1\linewidth]{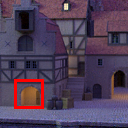}\vspace{1pt}
			\put(-18.8,0){\includegraphics[scale=0.75]{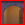}}\\
			\includegraphics[width=1\linewidth]{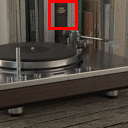}\vspace{1pt}
			\put(-18.8,0){\includegraphics[scale=0.75]{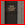}}
			\caption*{\fontfamily{Times New Roman}{I-TSS}}
	\end{minipage}}
	\hspace{-1.7mm}
	\subfloat{
		\begin{minipage}[b]{0.1\linewidth}
			\captionsetup{
				font={small},
				labelfont=bf,        % 标签字体为粗体
				textfont={rm}, % 正文字体为 Times new roman
				singlelinecheck=true % 不允许换行 
			}
			\includegraphics[width=1\linewidth]{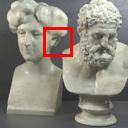}\vspace{1pt}
			\put(-18.8,0){\includegraphics[scale=0.75]{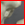}}\\
			\includegraphics[width=1\linewidth]{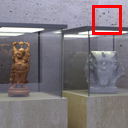}\vspace{1pt}
			\put(-18.8,0){\includegraphics[scale=0.75]{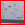}}\\
			\includegraphics[width=1\linewidth]{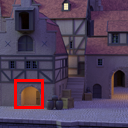}\vspace{1pt}
			\put(-18.8,0){\includegraphics[scale=0.75]{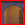}}\\
			\includegraphics[width=1\linewidth]{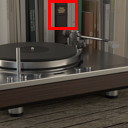}\vspace{1pt}
			\put(-18.8,0){\includegraphics[scale=0.75]{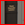}}
			\caption*{\fontfamily{Times New Roman}{Original}}
	\end{minipage}}
	\caption{The completion results of light field data by different methods on \emph{Greek}, \emph{Museum}, \emph{Medieval2}, and \emph{Vinyl} with SR = 0.1.}	
	\label{visual3}
\end{figure*}
\subsubsection{Experimental Results of Multispectral Images}

Table \ref{result3} reports the PSNR, SSIM, and RE values for the completion results of different tensor decomposition-based methods on MSIs. Among these methods, HTNN and SVDinsTN demonstrate comparable performance in most cases. Notably, SVDinsTN achieves the second-best or third-best performance across most MSIs, which can be attributed to its effective low-rank characterization of data. Importantly, the proposed I-TSS significantly outperforms all other methods across all evaluation metrics. This superiority stems from its ability to configure a suitable mixture of structures involving heterogeneous interaction families.

Fig. \ref{visual1} provides a visual comparison of the MSI completion results from all compared tensor decomposition-based methods. For enhanced clarity, the region of interest (marked by a red rectangle) at the bottom-right corner of each image is enlarged. From the visual results, we can observe that
HaLRTC effectively restores the primary outlines of missing regions but struggles with capturing fine-grained details and maintaining color fidelity.
In contrast, SiLRTCTT, TRLRF, HTNN, and LTNN produce completion results with richer spatial details. However, their performance remains unsatisfactory in terms of local fine details.
SVDinsTN yields visually pleasing results in both spatial geometry and color consistency, though subtle blurring artifacts are still observable.
By comparison, our I-TSS outperforms all other approaches in preserving both global structure and local details, which further validates its powerful recovery capability.

\subsubsection{Experimental Results of Color Videos}

Table \ref{result4} presents the quantitative metrics (PSNR, SSIM, and RE) of all competing methods for color video completion under various sampling rates. A comparison of these results reveals that our I-TSS distinctly outperforms other baseline methods, achieving the optimal PSNR, SSIM, and RE values across most experimental settings. Specifically, I-TSS delivers a notable PSNR improvement of approximately 1–3 dB over the second-best performing methods (i.e., HTNN or LTNN).

Further visual validation of the performance of our method is provided in Fig. \ref{visual2}, which displays a representative frame from the reconstructed videos under the challenging SR=0.1 condition (i.e., 90\% of data missing). The visual results strongly confirm the effectiveness of I-TSS. Compared to other baseline methods,  the proposed I-TSS significantly excels at accurately reconstructing both the global structural coherence of video frames and their intricate local details (e.g., texture and edge sharpness). This superiority directly stems from the powerful representation capability of the proposed I-TSS, which leverages a suitable mixture of structures involving heterogeneous interaction families to model the spatiotemporal correlations of video data.

\subsubsection{Experimental Results of Light Field Data}

Table \ref{result5} provides a quantitative comparison of all compared methods for light field data completion under various sampling rates. For clarity, the best-performing results are highlighted in bold, and the second-best in underline. Notably, the proposed I-TSS method consistently outperforms all competitors, achieving the optimal PSNR, SSIM, and RE values across all sampling rates. This consistent superiority underscores the effectiveness of I-TSS in handling multi-dimensional data with complex structural dependencies and intricate textural details.

Additionally, Fig. \ref{visual3} offers visual validation of the reconstruction quality for light field data under the stringent SR=0.1 condition. In this extreme scenario, where the high dimensionality and structural complexity of light field data pose significant challenges, the compared methods often exhibit pronounced limitations: reconstructed results suffer from distorted local details, loss of textural fidelity, or excessive smoothing artifacts. In sharp contrast, I-TSS distinguishes itself by preserving fine textures and structural details. These visual results further validate its superior capability in processing complex light field data, aligning with the quantitative metrics in Table \ref{result5}.

Overall, these results validate the innovative value of the I-TSS, which provides a framework that can identify either a single structure beyond a predefined interaction family or a mixture of structures involving different interaction families. These two unique advantages surpass the state-of-the-art tensor decomposition structure search methods. 
\begin{table}[t]
	\centering
	\caption{The PSNR values and corresponding interaction families of the representative search method (i.e., SVDinsTN) and the proposed I-TSS across various data with a sampling rate of 0.2. The symbol `+' represents the mixture of structures.}
	\setlength{\tabcolsep}{1.2mm}{
		\begin{tabular}{ccccc}
			\toprule
			\multicolumn{1}{c}{\multirow{6}{*}{Data}}&\multicolumn{1}{ c }{\emph{Flowers}}&\multicolumn{1}{ c }{\emph{Beers}}&\multicolumn{1}{ c }{\emph{Balloons}}&\multicolumn{1}{ c }{\emph{Feathers}}\\
			\cmidrule{2-5}
			&$\raisebox{-.5\height}{\includegraphics[width=0.14\linewidth]{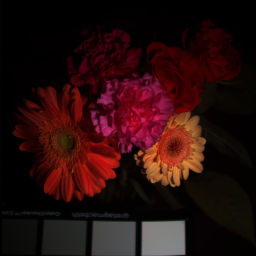}}$&$\raisebox{-.5\height}{\includegraphics[width=0.14\linewidth]{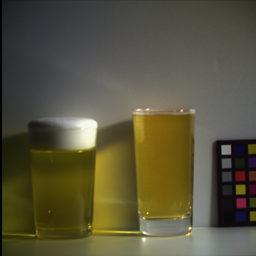}}$&$\raisebox{-.5\height}{\includegraphics[width=0.14\linewidth]{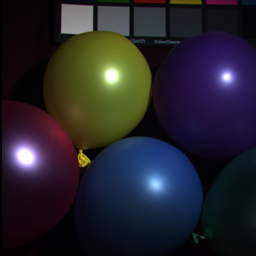}}$&$\raisebox{-.5\height}{\includegraphics[width=0.14\linewidth]{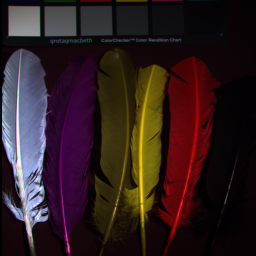}}$\\
			%			\midrule
			%			\multicolumn{1}{c}{Method}&PSNR/TG&PSNR/TG&PSNR/TG&PSNR/TG\\
			
			\midrule
			\multicolumn{1}{c}{\multirow{3.5}{*}{SVDinsTN}}&Tensor&Tensor&Tensor&Tensor\\
			&Contraction&Contraction&Contraction&Contraction\\
			\cmidrule{2-5}
			&38.18&42.72&42.08&38.83\\
			\midrule
			\multicolumn{1}{c}{\multirow{3.5}{*}{Top-1 I-TSS}}&\multicolumn{1}{c}{\multirow{2}{*}{T-Product}}&Tensor&\multicolumn{1}{c}{\multirow{2}{*}{T-Product}}&\multicolumn{1}{c}{\multirow{2}{*}{T-Product}}\\
			&&Contraction&&\\
			\cmidrule{2-5}
			&42.51&46.42&42.87&41.54\\
			\midrule
			\multicolumn{1}{c}{\multirow{4.5}{*}{Top-2 I-TSS}}&Tensor&Tensor&Tensor&Tensor\\
			&Contraction+&Contraction+&Contraction+&Contraction+\\
			&T-Product&T-Product&T-Product&T-Product\\
			\cmidrule{2-5}
			&42.57&48.30&48.08&41.86\\
			\midrule
			\multicolumn{1}{c}{\multirow{2.5}{*}{Top-3 I-TSS}}&ALL&ALL&ALL&ALL\\
			\cmidrule{2-5}
			&\textbf{42.70}&\textbf{48.79}&\textbf{49.49}&\textbf{42.54}\\
			\bottomrule
	\end{tabular}}\label{result6}
\end{table}
\section{Discussion}
\label{section:7}
In this section, we discuss the influence of four aspects of the I-TSS, i.e., the search capability, the candidate structure selection strategy, the parameters updating order, and the energy ratio threshold.

\subsection{The Search Capability of the Proposed I-TSS}
	To comprehensively evaluate the search capability of the I-TSS, we present the PSNR values and corresponding interaction families for the representative search method (i.e., SVDinsTN) and the proposed I-TSS across various data with a sampling rate of 0.2.
  As presented in Table~\ref{result6}, I-TSS demonstrates a superior capability to identify a suitable tensor decomposition structure. Specifically, SVDinsTN can only search the tensor network decomposition within the predefined tensor contraction family. In contrast, the proposed I-TSS can identify either a single structure beyond a predefined interaction family or a mixture of structures involving different interaction families, which yields higher PSNR values than SVDinsTN. Additionally, we can observe that increasing the top-$k$ value enhances the performance, as it combines the unique advantages of different interactions.

 \begin{table}[!h]
	\centering
	\caption{The proposed I-TSS variants with different candidate structure selection strategies for MSIs \emph{Flowers}, \emph{Balloons}, and \emph{Feathers} with a sampling rate of 0.2.}
	\setlength{\tabcolsep}{1.9mm}{
		\begin{tabular}{ccccccc}
			\toprule
			\multicolumn{1}{c}{Data}&\multicolumn{2}{ c }{\emph{Flowers}}&\multicolumn{2}{ c }{\emph{Balloons}}&\multicolumn{2}{ c }{\emph{Feathers}}\\
			\midrule
			Selection Strategy &PSNR&SSIM&PSNR&SSIM&PSNR&SSIM\\
			\midrule
			Max Sampling&\textbf{42.51}&\textbf{0.985}&\textbf{49.49}&\textbf{0.996}&\textbf{41.54}&\textbf{0.980}\\
			Random Sampling&37.64&0.880&40.70&0.973&37.97&0.949\\
			\bottomrule
	\end{tabular}}\label{result7}
\end{table}
\subsection{The Influence of the Candidate Structure Selection Strategy}
 To analyze the influence of candidate structures selection strategy on the proposed I-TSS, we illustrate the performance (in terms of PSNR and SSIM) of I-TSS variants with different candidate structure selection strategies (i.e., max sampling and random sampling) for MSIs \emph{Flowers}, \emph{Balloons}, and \emph{Feathers} with a sampling rate of 0.2 in Table~\ref{result7}. The max sampling strategy (selecting one candidate structure with the highest gating score) consistently outperforms the random sampling strategy (randomly selecting one candidate structure based on their predicted gating scores) across all data. Thus, in our experiments, we adopt the max sampling strategy as the candidate structure selection strategy.

%\begin{figure}[h]
%	\centering
%	\includegraphics[width=0.9\hsize]{ss2.png}		
%	\caption{The different selection strategies of the proposed I-TSS variants for MSIs \emph{Flowers}, \emph{Balloons}, and \emph{Feathers} with a sampling rate 0.2.}
%	
%	\label{fig4}
%\end{figure}

 \begin{table}[!h]
 	\centering
 	\caption{The proposed I-TSS variants with different parameters updating orders for MSIs \emph{Flowers}, \emph{Balloons}, and \emph{Feathers} with a sampling rate of 0.2.}
 	\setlength{\tabcolsep}{2.1mm}{
 		\begin{tabular}{ccccccc}
 			\toprule
 			\multicolumn{1}{c}{Data}&\multicolumn{2}{ c }{\emph{Flowers}}&\multicolumn{2}{ c }{\emph{Balloons}}&\multicolumn{2}{ c }{\emph{Feathers}}\\
 			\midrule
 			Updating Order &PSNR&SSIM&PSNR&SSIM&PSNR&SSIM\\
 			\midrule
 			I&\textbf{42.99}&0.938&46.76&0.994&41.65&0.981\\
 			II&42.70&\textbf{0.939}&\textbf{49.49}&\textbf{0.996}&\textbf{42.54}&\textbf{0.984}\\
 			III&34.34&0.895&41.63&0.983&35.94&0.941\\
 			\bottomrule
 	\end{tabular}}\label{result8}
 \end{table}
\subsection{The Influence of the Parameters Updating Order}
To discuss the influence of the parameters updating order, we present the performance of I-TSS variants with different parameters updating orders for MSIs \emph{Flowers}, \emph{Balloons}, and \emph{Feathers} with a sampling rate of 0.2 in Table~\ref{result8}. The notations in the table are defined as 
\begin{itemize}
	\item Order I: Simultaneous update of all candidate structures and all gating scores. 
	\item Order II: Alternating updates between all candidate structures and all gating scores.
	\item Order III: Alternating updates between each individual candidate structure and its corresponding gating score.
\end{itemize}

From Table~\ref{result8}, we can observe that updating order II consistently outperforms the other two updating orders across all data. The success of updating order II lies in its capability to balance the need for joint optimization between candidate structures and gating scores while allowing for iterative refinement through alternation, thus leveraging the synergies between these components to achieve enhanced reconstruction performance.
\begin{figure}[!h]
	\centering
	\subfloat{
		\begin{minipage}[b]{0.32\linewidth}
			\captionsetup{
				font={small}, 
				labelfont=bf,        % 标签字体为粗体
				textfont={rm}, % 正文字体为 Times new roman
				singlelinecheck=true % 不允许换行 
			}
			\includegraphics[width=\hsize]{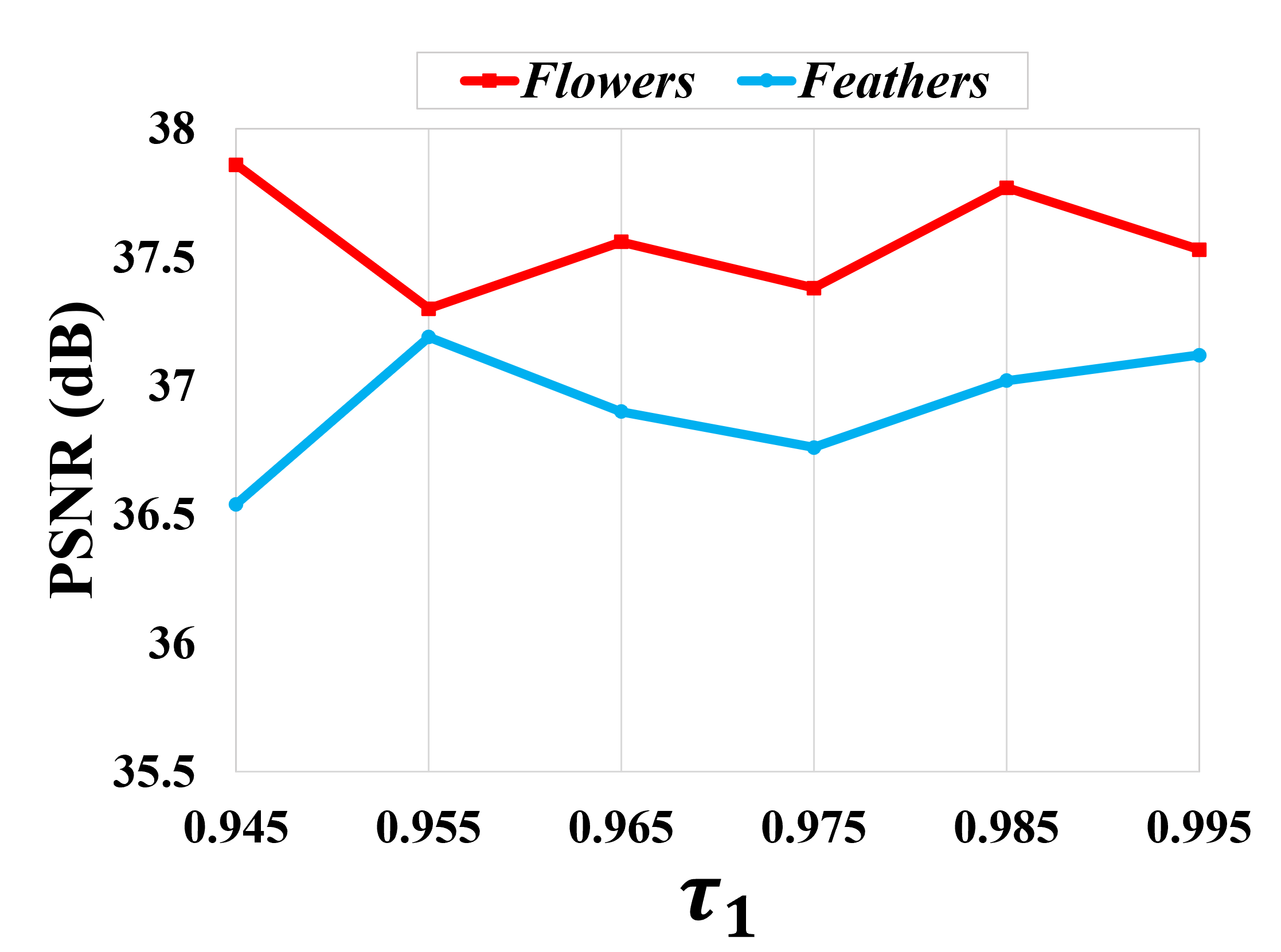}	
	\end{minipage}}
	\subfloat{
		\begin{minipage}[b]{0.32\linewidth}
			\captionsetup{
				font={small}, 
				labelfont=bf,        % 标签字体为粗体
				textfont={rm}, % 正文字体为 Times new roman
				singlelinecheck=true % 不允许换行 
			}
			\includegraphics[width=\hsize]{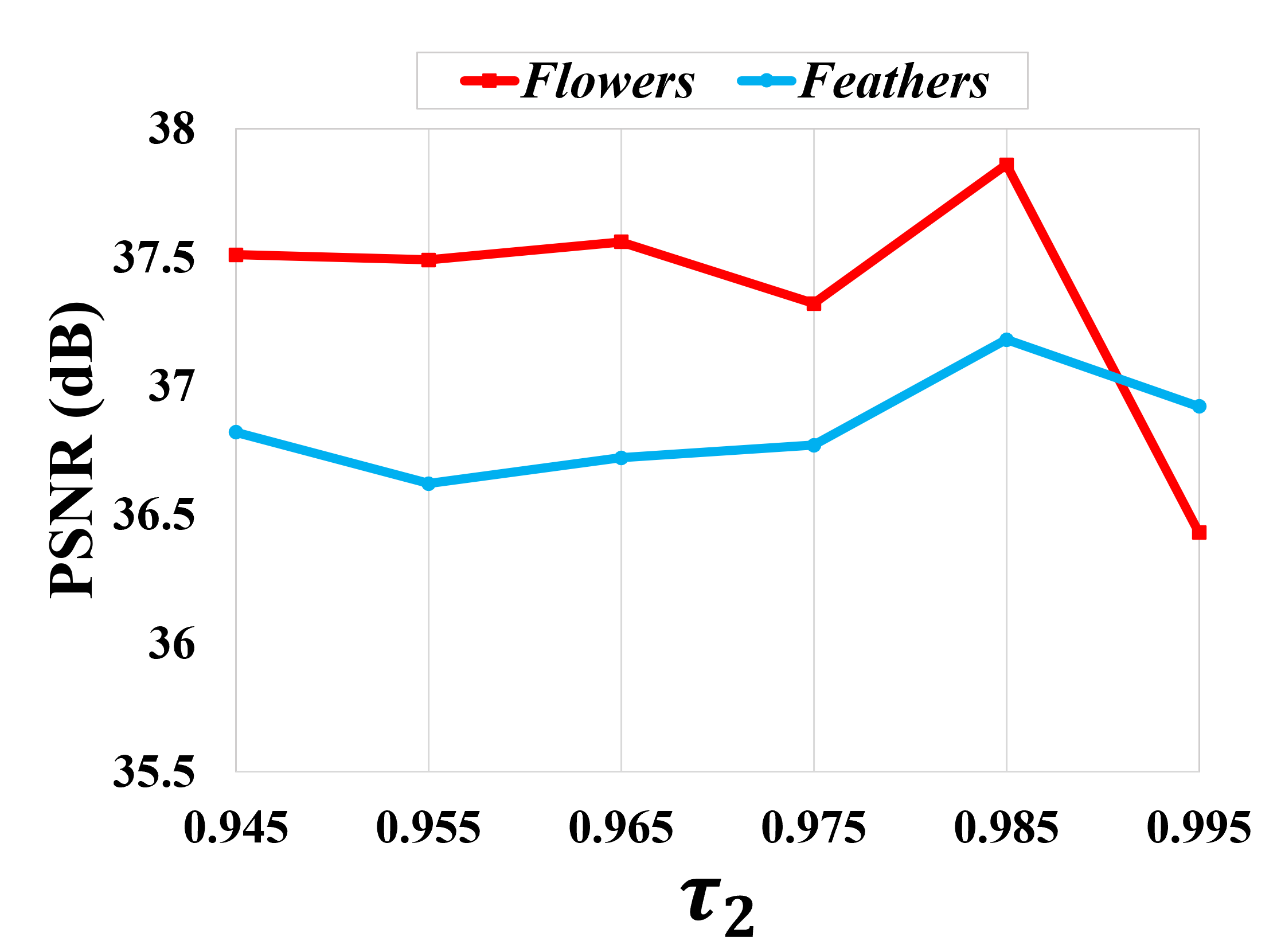}
	\end{minipage}}
	\subfloat{
		\begin{minipage}[b]{0.32\linewidth}
			\captionsetup{
				font={small}, 
				labelfont=bf,        % 标签字体为粗体
				textfont={rm}, % 正文字体为 Times new roman
				singlelinecheck=true % 不允许换行 
			}
			\includegraphics[width=\hsize]{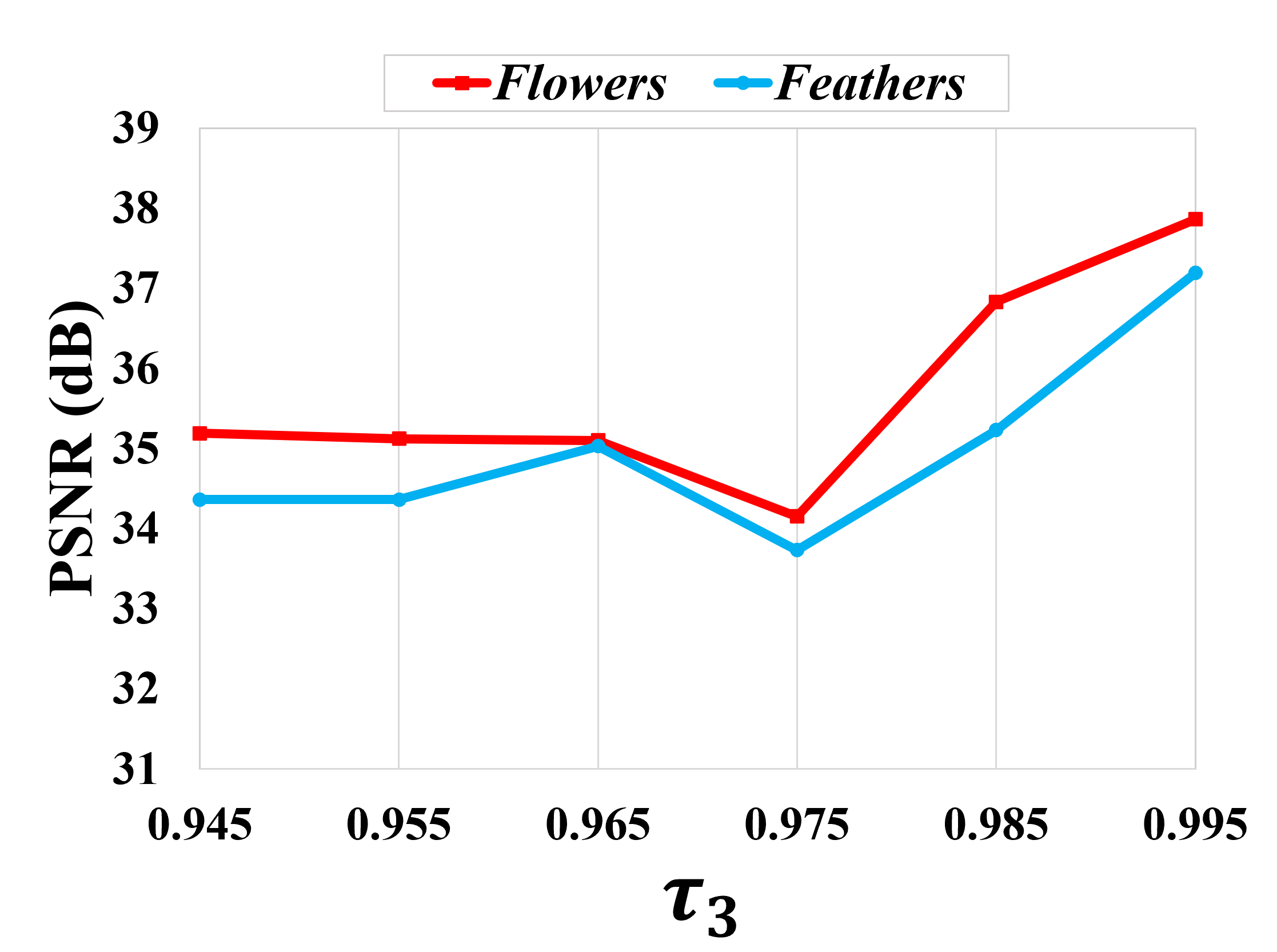}
	\end{minipage}}
	\caption{The PSNR values of the proposed I-TSS with different values of energy ratio thresholds for MSIs \emph{Flowers} and \emph{Feathers} with a sampling rate of 0.1.
	}
	\label{fig5}
\end{figure}

\subsection{The Influence of the Energy Ratio Threshold}
Choosing appropriate energy ratio thresholds is a crucial step in the proposed I-TSS. To comprehensively examine the influence of energy ratio thresholds on the performance of the proposed I-TSS, we modify the value of one energy ratio threshold at a time while fixing the others. The results are presented in
Fig.~\ref{fig5}. From these results, we find that our method shows a certain degree of robustness, given its ability to achieve satisfactory performance across a wide range of values.

\section{Conclusion}
\label{section:8}
In this work, to address the challenging and relatively under-explored problem of identifying a suitable tensor decomposition structure to represent given data, we suggested a tensor decomposition structure search framework from an interaction perspective that can identify either a single structure beyond a predefined interaction family or a mixture of structures involving heterogeneous interaction families. Specifically, we constructed a heterogeneous interaction-induced candidate structure set via a systematic interaction-perspective review, estimated its ranks uniformly via an energy-based scheme, and adopted top-$k$ gating to adaptively select suitable single structure or mixture of structures. Theoretically, we derived an approximation error bound for I-TSS, thereby establishing its approximation capability. Extensive experiments on synthetic and real-world datasets demonstrated that I-TSS consistently outperforms state-of-the-art tensor decomposition methods.
\bibliographystyle{IEEEtran}
\bibliography{egbib}

@inproceedings{Han023,
  author={Kang Han and Wei Xiang},
  title={Multiscale Tensor Decomposition and Rendering Equation Encoding for View Synthesis},
  year={2023},
  pages={4232-4241},
  booktitle={Proceedings of the IEEE/CVF Conference on Computer Vision and Pattern Recognition},
}

@inproceedings{bay,
author = {Rai, Piyush and Wang, Yingjian and Guo, Shengbo and Chen, Gary and Dunson, David and Carin, Lawrence},
title = {Scalable Bayesian low-rank decomposition of incomplete multiway tensors},
year = {2014},
booktitle = {Proceedings of the International Conference on Machine Learning},
volume = {32},
}

@article{Zhao2014BayesianCF,
  title={Bayesian CP Factorization of Incomplete Tensors with Automatic Rank Determination},
  author={Qibin Zhao and Liqing Zhang and Andrzej Cichocki},
  journal={IEEE Transactions on Pattern Analysis and Machine Intelligence},
  year={2014},
  volume={37},
  pages={1751-1763},
}

@INPROCEEDINGS{9878497,
  author={Luo, Yisi and Zhao, Xile and Meng, Deyu and Jiang, Taixiang},
  booktitle={Proceedings of the IEEE/CVF Conference on Computer Vision and Pattern Recognition}, 
  title={{HLRTF}: Hierarchical Low-Rank Tensor Factorization for Inverse Problems in Multi-Dimensional Imaging}, 
  year={2022},
  volume={},
  number={},
  pages={19281-19290},
}

@ARTICLE{7859390,
  author={Bengua, Johann A. and Phien, Ho N. and Tuan, Hoang Duong and Do, Minh N.},
  journal={IEEE Transactions on Image Processing}, 
  title={Efficient Tensor Completion for Color Image and Video Recovery: Low-Rank Tensor Train}, 
  year={2017},
  volume={26},
  number={5},
  pages={2466-2479},
}

@ARTICLE{10530915,
  author={Han, Zhi-Long and Huang, Ting-Zhu and Zhao, Xi-Le and Zhang, Hao and Wu, Wei-Hao},
  journal={IEEE Transactions on Circuits and Systems for Video Technology}, 
  title={Nested Fully-Connected Tensor Network Decomposition for Multi-Dimensional Visual Data Recovery}, 
  year={2024},
  volume={34},
  number={10},
  pages={10092-10106},
}

@article{sr,
author = {Kolda, Tamara G. and Bader, Brett W.},
title = {Tensor Decompositions and Applications},
journal = {SIAM Review},
volume = {51},
number = {3},
pages = {455-500},
year = {2009},

}

@ARTICLE{7891546,
  author={Sidiropoulos, Nicholas D. and De Lathauwer, Lieven and Fu, Xiao and Huang, Kejun and Papalexakis, Evangelos E. and Faloutsos, Christos},
  journal={IEEE Transactions on Signal Processing}, 
  title={Tensor Decomposition for Signal Processing and Machine Learning}, 
  year={2017},
  volume={65},
  number={13},
  pages={3551-3582},
  }

@INPROCEEDINGS{dtt,
  author={Zhengyu Chen and Ziqing Xu and Donglin Wang},
  booktitle={Proceedings of the AAAI Conference on Artificial Intelligence}, 
  title={Deep Transfer Tensor Decomposition with Orthogonal Constraint for Recommender Systems}, 
  year={2021},
  volume={35},
  number={5},
  pages={4010-4018},
}

@ARTICLE{7902201,
  author={Chien, Jen-Tzung and Bao, Yi-Ting},
  journal={IEEE Transactions on Neural Networks and Learning Systems}, 
  title={Tensor-factorized neural networks}, 
  year={2018},
  volume={29},
  number={5},
  pages={1998-2011},
}

@inproceedings{3524938,
author = {Li, Chao and Sun, Zhun},
title = {Evolutionary topology search for tensor network decomposition},
year = {2020},
 volume={119},
pages={5947–5957},
booktitle = {Proceedings of the International Conference on Machine Learning},
}

@article{doi:10.1137/090752286,
author = {Oseledets, I. V.},
title = {Tensor-Train Decomposition},
journal = {SIAM Journal on Scientific Computing},
volume = {33},
number = {5},
pages = {2295-2317},
year = {2011},

}

@article{Zhao2016TensorRD,
  title={Tensor Ring Decomposition},
  author={Qibin Zhao and Guoxu Zhou and Shengli Xie and Liqing Zhang and Andrzej Cichocki},
  journal={ArXiv},
  year={2016},
  volume={abs/1606.05535},
}

@INPROCEEDINGS{fctn,
  author={Yu-Bang Zheng and Ting-Zhu Huang and Xi-Le Zhao and Qibin Zhao and Tai-Xiang Jiang},
  booktitle={Proceedings of the AAAI Conference on Artificial Intelligence}, 
  title={Fully-Connected Tensor Network Decomposition and Its Application to Higher-Order Tensor Completion}, 
  year={2021},
  volume={35},
  number={12},
  pages={11071-11078},
}

@INPROCEEDINGS{8237869,
  author={Wang, Wenqi and Aggarwal, Vaneet and Aeron, Shuchin},
  booktitle={Proceedings of the IEEE International Conference on Computer Vision}, 
  title={Efficient Low Rank Tensor Ring Completion}, 
  year={2017},
  volume={},
  number={},
  pages={5698-5706},
}

@article{doi:10.1137/110837711,
author = {Kilmer, Misha E. and Braman, Karen and Hao, Ning and Hoover, Randy C.},
title = {Third-Order Tensors as Operators on Matrices: A Theoretical and Computational Framework with Applications in Imaging},
journal = {SIAM Journal on Matrix Analysis and Applications},
volume = {34},
number = {1},
pages = {148-172},
year = {2013},
}

@ARTICLE{8740980,
  author={Zhou, Yang and Cheung, Yiu-Ming},
  journal={IEEE Transactions on Pattern Analysis and Machine Intelligence}, 
  title={Bayesian Low-Tubal-Rank Robust Tensor Factorization with Multi-Rank Determination}, 
  year={2021},
  volume={43},
  number={1},
  pages={62-76},
}

@ARTICLE{9064895,
  author={Zhang, Feng and Wang, Jianjun and Wang, Wendong and Xu, Chen},
  journal={IEEE Transactions on Pattern Analysis and Machine Intelligence}, 
  title={Low-Tubal-Rank Plus Sparse Tensor Recovery With Prior Subspace Information}, 
  year={2021},
  volume={43},
  number={10},
  pages={3492-3507}
}

@ARTICLE{9369083,
  author={Hou, Jingyao and Zhang, Feng and Qiu, Haiquan and Wang, Jianjun and Wang, Yao and Meng, Deyu},
  journal={IEEE Transactions on Pattern Analysis and Machine Intelligence}, 
  title={Robust Low-Tubal-Rank Tensor Recovery From Binary Measurements}, 
  year={2022},
  volume={44},
  number={8},
  pages={4355-4373}}

@inproceedings{3618408,
author = {Ghadiri, Mehrdad and Fahrbach, Matthew and Fu, Gang and Mirrokni, Vahab},
title = {Approximately optimal core shapes for tensor decompositions},
year = {2023},
pages={11237 - 11254},
number={451},
booktitle = {Proceedings of the International Conference on Machine Learning},
}

@article{Hashemizadeh,
  title={Adaptive Learning of Tensor Network Structures},
  author={Meraj Hashemizadeh and Michelle Liu and Jacob Miller and Guillaume Rabusseau},
  journal    = {ArXiv},
  volume = {abs/2008.05437},
  year={2020},
}

@InProceedings{pmlr-v162-li22y,
  title = 	 {Permutation Search of Tensor Network Structures via Local Sampling},
  author =       {Li, Chao and Zeng, Junhua and Tao, Zerui and Zhao, Qibin},
  booktitle = 	 {Proceedings of the International Conference on Machine Learning},
  pages = 	 {13106--13124},
  year = 	 {2022},
  volume = 	 {162},
}

@InProceedings{pmlr-v202-li23ar,
  title = 	 {Alternating Local Enumeration ({T}n{ALE}): Solving Tensor Network Structure Search with Fewer Evaluations},
  author =       {Li, Chao and Zeng, Junhua and Li, Chunmei and Caiafa, Cesar F and Zhao, Qibin},
  booktitle = 	 {Proceedings of the International Conference on Machine Learning},
  pages = 	 {20384--20411},
  year = 	 {2023},
  volume = 	 {202},
}

@ARTICLE{cis,
  author={Nannan Li and Yu Pan and Yaran Chen and Zixiang Ding and Dongbin Zhao and Zenglin Xu},
  journal={ Complex \& Intelligent Systems}, 
  title={Heuristic rank selection with progressively searching tensor ring network}, 
  year={2022},
  volume={8},
  pages={771–785},}

@article{10.1109,
author = {Liu, Yipeng and Chen, Jie and Lu, Yingcong and Ou, Weiting and Long, Zhen and Zhu, Ce},
title = {Adaptively Topological Tensor Network for Multi-View Subspace Clustering},
year = {2024},
volume = {36},
number = {11},
journal = {IEEE Transactions on Knowledge and Data Engineering},
pages = {5562–5575},
}

@inproceedings{NieWT21,
  title = {Adaptive Tensor Networks Decomposition},
  author = {Chang Nie and Huan Wang and Le Tian},
  year = {2021},
  pages = {148},
  booktitle = {Proceedings of the British Machine Vision Conference},
}

@ARTICLE{9321501,
  author={Sedighin, Farnaz and Cichocki, Andrzej and Phan, Anh-Huy},
  journal={IEEE Journal of Selected Topics in Signal Processing}, 
  title={Adaptive Rank Selection for Tensor Ring Decomposition}, 
  year={2021},
  volume={15},
  number={3},
  pages={454-463},
}

@INPROCEEDINGS{10655696,
  author={Zheng, Yu-Bang and Zhao, Xi-Le and Zeng, Junhua and Li, Chao and Zhao, Qibin and Li, Heng -Chao and Huang, Ting-Zhu},
  booktitle={Proceedings of the IEEE/CVF Conference on Computer Vision and Pattern Recognition}, 
  title={{SVDinsTN}: A Tensor Network Paradigm for Efficient Structure Search from Regularized Modeling Perspective}, 
  year={2024},
  volume={},
  number={},
  pages={22413-22422},}

@ARTICLE{34234014,
  author={Kilmer, Misha E and Horesh, Lior and Avron, Haim and Newman, Elizabeth},
  journal={Proceedings of the National Academy of Sciences of the United States of America}, 
  title={Tensor-tensor algebra for optimal representation and compression of multiway data}, 
  year={2021},
  volume={118},
  number={28},
  pages={e2015851118},
}

@ARTICLE{8066348,
  author={Zhou, Pan and Lu, Canyi and Lin, Zhouchen and Zhang, Chao},
  journal={IEEE Transactions on Image Processing}, 
  title={Tensor Factorization for Low-Rank Tensor Completion}, 
  year={2018},
  volume={27},
  number={3},
  pages={1152-1163},
}

@article{Kingma2014AdamAM,
  title={Adam: A Method for Stochastic Optimization},
  author={Diederik P. Kingma and Jimmy Ba},
  journal    = {ArXiv},
  year={2014},
  volume={abs/1412.6980},
}

@article{101109,
author = {Liu, Ji and Musialski, Przemyslaw and Wonka, Peter and Ye, Jieping},
title = {Tensor Completion for Estimating Missing Values in Visual Data},
year = {2013},
volume = {35},
number = {1},
journal = {IEEE Transactions on Pattern Analysis and Machine Intelligence},
pages = {208–220},
numpages = {13},
}

@ARTICLE{9730793,
  author={Qin, Wenjin and Wang, Hailin and Zhang, Feng and Wang, Jianjun and Luo, Xin and Huang, Tingwen},
  journal={IEEE Transactions on Image Processing}, 
  title={Low-Rank High-Order Tensor Completion With Applications in Visual Data}, 
  year={2022},
  volume={31},
  number={},
  pages={2433-2448},
}

@inproceedings{101609,
author = {Yuan, Longhao and Li, Chao and Mandic, Danilo and Cao, Jianting and Zhao, Qibin},
title = {Tensor ring decomposition with rank minimization on latent space: an efficient approach for tensor completion},
year = {2019},
 booktitle={Proceedings of the AAAI Conference on Artificial Intelligence}, 
numpages = {8},
}

@ARTICLE{10507804,
  author={Qiu, Yuning and Zhou, Guoxu and Wang, Andong and Zhao, Qibin and Xie, Shengli},
  journal={IEEE Transactions on Neural Networks and Learning Systems}, 
  title={Balanced Unfolding Induced Tensor Nuclear Norms for High-Order Tensor Completion}, 
  year={2025},
  volume={36},
  number={3},
  pages={4724-4737},
}

@ARTICLE{cp,
  author={Hitchcock, Frank L.},
  journal={Journal of Mathematics and Physics}, 
  title={The Expression of a Tensor or a Polyadic as a Sum of Products}, 
  year={1927},
  volume={6},
  number={1-4},
  pages={164-189},
}

@ARTICLE{tucker,
  author={Tucker, Ledyard R},
  journal={Psychometrika}, 
  title={Some mathematical notes on three-mode factor analysis}, 
  year={1966},
  volume={31},
  number={3},
  pages={279-311},
}

@ARTICLE{11184870,
  author={Xu, Le and Cheng, Lei and Wong, Ngai and Wu, Yik-Chung},
  journal={IEEE Transactions on Pattern Analysis and Machine Intelligence}, 
  title={{To Fold or Not to Fold}: Graph Regularized Tensor Train for Visual Data Completion}, 
  year={2025},
  volume={},
  number={},
  pages={1-18},}

@ARTICLE{10665981,
  author={Liu, Sheng and Zhao, Xi-Le and Leng, Jinsong and Li, Ben-Zheng and Yang, Jing-Hua and Chen, Xinyu},
  journal={IEEE Transactions on Signal Processing}, 
  title={Revisiting High-Order Tensor Singular Value Decomposition From Basic Element Perspective}, 
  year={2024},
  volume={72},
  number={},
  pages={4589-4603},}

@article{ZHENG2025107808,
title = {Tensor network decomposition for data recovery: Recent advancements and future prospects},
journal = {Neural Networks},
volume = {191},
pages = {107808},
year = {2025},
author = {Yu-Bang Zheng and Xi-Le Zhao and Heng-Chao Li and Chao Li and Ting-Zhu Huang and Qibin Zhao},
}

@ARTICLE{9791317,
  author={Zhang, Yang and Wang, Yao and Han, Zhi and Chen, Xi’ai and Tang, Yandong},
  journal={IEEE Transactions on Circuits and Systems for Video Technology}, 
  title={Effective Tensor Completion via Element-Wise Weighted Low-Rank Tensor Train With Overlapping Ket Augmentation}, 
  year={2022},
  volume={32},
  number={11},
  pages={7286-7300},}

@ARTICLE{10556743,
  author={Qiu, Duo and Yang, Bei and Zhang, Xiongjun},
  journal={IEEE Transactions on Circuits and Systems for Video Technology}, 
  title={Robust Tensor Completion via Dictionary Learning and Generalized Nonconvex Regularization for Visual Data Recovery}, 
  year={2024},
  volume={34},
  number={11},
  pages={11026-11039},}

@ARTICLE{10026252,
  author={Han, Zhi and Zhang, Shaojie and Liu, Zhiyu and Wang, Yanmei and Yao, Junping and Wang, Yao},
  journal={IEEE Transactions on Circuits and Systems for Video Technology}, 
  title={Tensor Robust Principal Component Analysis With Side Information: Models and Applications}, 
  year={2023},
  volume={33},
  number={8},
  pages={3713-3725}}

@ARTICLE{11386834,
  author={Shi, Zhi-Wei and Zheng, Yu-Bang and Li, Heng-Chao and Plaza, Antonio},
  journal={IEEE Transactions on Circuits and Systems for Video Technology}, 
  title={Multidimensional Image Reconstruction via Deep Nonlinear Low-Rank Tensor Decomposition}, 
  year={2026},
  volume={36},
  number={6},
  pages={8260-8273}}

@ARTICLE{11020636,
  author={Chen, Yong and Yuan, Feiwang and Lai, Wenzhen and Zeng, Jinshan and He, Wei and Huang, Qing},
  journal={IEEE Transactions on Circuits and Systems for Video Technology}, 
  title={Low-Rank Tensor Meets Deep Prior: Coupling Model-Driven and Data-Driven Methods for Hyperspectral Image Reconstruction}, 
  year={2025},
  volume={35},
  number={11},
  pages={11685-11697}}

@ARTICLE{11203237,
  author={Chen, Yong and Zhou, Min and Gui, Xinfeng and Yuan, Feiwang and He, Wei and Zeng, Jinshan},
  journal={IEEE Transactions on Circuits and Systems for Video Technology}, 
  title={Learning a Self-Supervised Low-Rank Decomposition Network for Hyperspectral Image Super-Resolution}, 
  year={2026},
  volume={36},
  number={3},
  pages={3964-3974}}

@ARTICLE{9007523,
  author={Xu, Yang and Wu, Zebin and Chanussot, Jocelyn and Wei, Zhihui},
  journal={IEEE Transactions on Circuits and Systems for Video Technology}, 
  title={Hyperspectral Computational Imaging via Collaborative Tucker3 Tensor Decomposition}, 
  year={2021},
  volume={31},
  number={1},
  pages={98-111}}

@ARTICLE{11083646,
  author={Shen, Qiangqiang and Wang, Hanzhang and Zhao, Yin-Ping and Chen, Yongyong and Liang, Yongsheng and Li, Xuelong},
  journal={IEEE Transactions on Circuits and Systems for Video Technology}, 
  title={Dual Tensor Low-Rank Representation for Subspace Clustering}, 
  year={2026},
  volume={36},
  number={1},
  pages={37-50}}

@ARTICLE{11231345,
  author={Wang, Hanzhang and Liu, Zonglin and Zhong, Zhiwei and Shen, Qiangqiang and Liang, Yongsheng and Chen, Yongyong},
  journal={IEEE Transactions on Circuits and Systems for Video Technology}, 
  title={Anchor-Induced Serial Tensor Representation for Multi-View Clustering}, 
  year={2026},
  volume={36},
  number={4},
  pages={4275-4286}}

\begin{IEEEbiography}[{\includegraphics[width=1in,height=1.25in,clip]{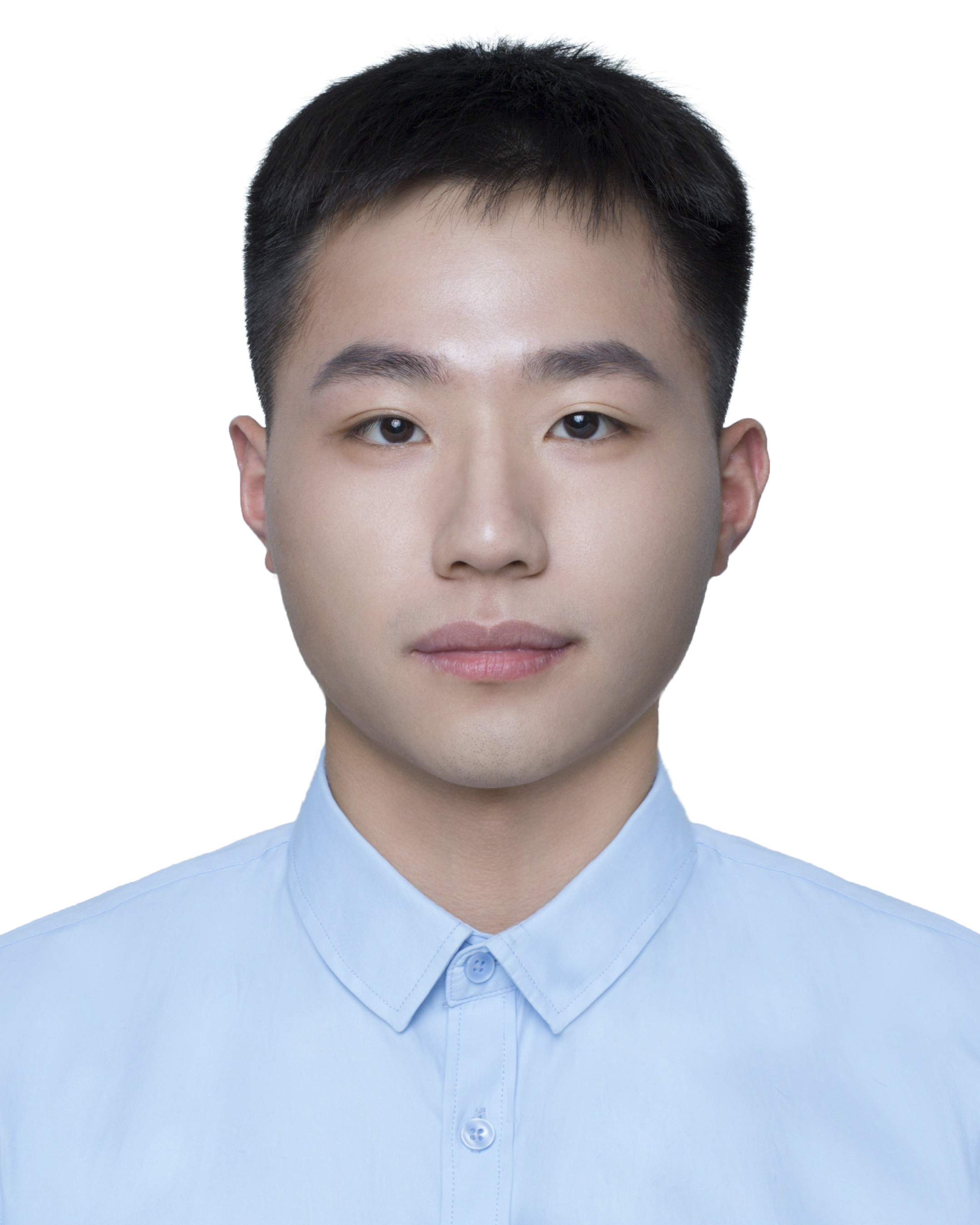}}]{Ting-Wei Zhou}
	received the B.S. degree from the College of Science, Zhejiang University of Technology, Hangzhou, China, in 2023. He is currently pursuing the Ph.D. degree with the School of Mathematical Sciences, University of Electronic Science and Technology of China, Chengdu, China.
	
	His research interests include high-order tensor modeling, computer	vision, and unsupervised learning.
\end{IEEEbiography}

\begin{IEEEbiography}[{\includegraphics[width=1in,height=1.25in,clip]{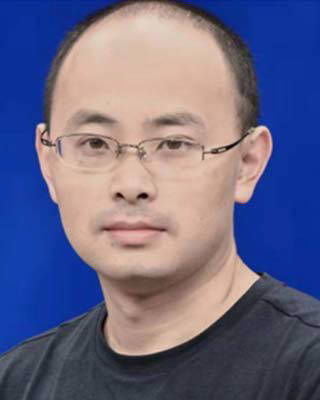}}]{Xi-Le Zhao}	received the M.S. and Ph.D. degrees	from the University of Electronic Science and Technology of China (UESTC), Chengdu, China, in 2009 and 2012. He worked as a post-doc with Prof. M. Ng at Hong Kong Baptist University from 2013 to	2014. He worked as a visiting scholar with Prof. J.	Bioucas Dias at University of Lisbon from 2016 to 2017.
	
He is currently a Professor with the School	of Mathematical Sciences, UESTC. His research interests include image processing, machine learning, and scientific computing. More information can be found on his homepage \href{https://zhaoxile.github.io/}{https://zhaoxile.github.io/}.
\end{IEEEbiography}

\begin{IEEEbiography}[{\includegraphics[width=1in,height=1.25in,clip,keepaspectratio]{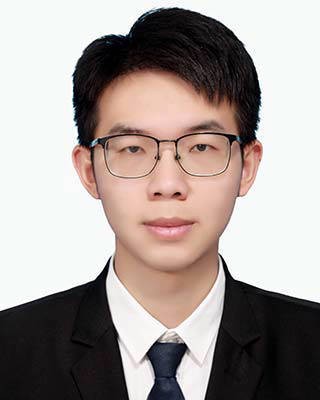}}]{Sheng Liu}
	received the B.S. degree in information	and computer science from the Northwest	A\&F University, Yangling, China, in 2022. He is currently working toward the Ph.D. degree with the School of Mathematical Sciences, University of Electronic Science and Technology of China	(UESTC), Chengdu, China. 
	
	His research interests	include tensor modeling and algorithms for high-order data recovery.
\end{IEEEbiography}

\begin{IEEEbiography}[{\includegraphics[width=1in,height=1.25in,clip,keepaspectratio]{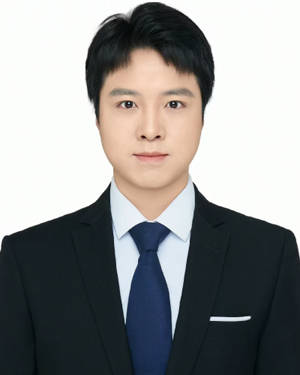}}]{Wei-Hao Wu}
	received the B.S. degree from the School of Mathematical Sciences, University of Electronic	Science and Technology of China, Chengdu,	China, in 2021, where he is currently pursuing the Ph.D. degree.
	
	His research interests include model-based tensor modeling and unsupervised learning for low-level visual tasks, such as inpainting and denoising.
\end{IEEEbiography}

\begin{IEEEbiography}[{\includegraphics[width=1in,height=1.25in,clip]{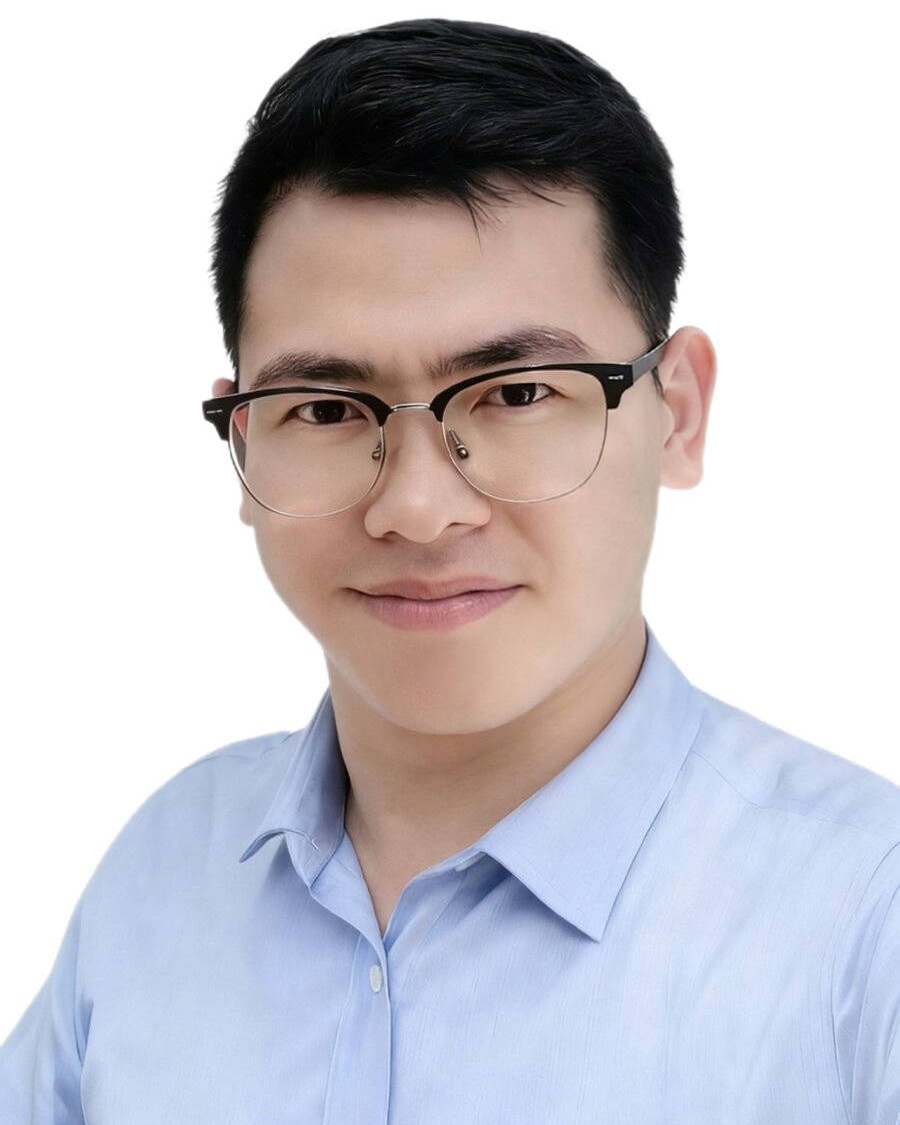}}]{Yu-Bang Zheng}
	received the B.S. degree from the Institute of Statistics and Applied Mathematics, Anhui University of Finance and Eco
	nomics, Bengbu, China, in 2017, and the Ph.D. degree from the School of Mathematical Sciences, University of Electronic Science and Technology of China, Chengdu, China, in 2022. From 2021 to 2022, he was a Student Trainee
	with the Tensor Learning Team, RIKEN Center for	Advanced Intelligence Project, Tokyo, Japan. He is currently an Assistant Professor with the School of Information Science and Technology, Southwest Jiaotong University,
	Chengdu. 
	
	His current research interests include tensor modeling and computing, machine learning, and high-dimensional data processing.
\end{IEEEbiography}

\begin{IEEEbiography}[{\includegraphics[width=1in,height=1.25in,clip]{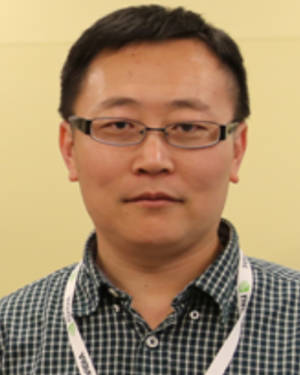}}]{Deyu Meng}
received the B.Sc., M.Sc., and Ph.D. degrees from Xi’an Jiaotong University, Xi’an, China, in 2001, 2004, and 2008, respectively. 

He is currently a Professor with the School of Mathematics and Statistics, Xi’an Jiaotong University,
and an Adjunct Professor with the Faculty of Information Technology, Macau University of Science and Technology, Macau, China. His research interests include model-based deep learning, variational networks, and meta-learning. 

Dr. Meng currently serves as an Associate Editor for \emph{IEEE Transactions on Pattern Analysis and Machine
Intelligence, Science China-Information Sciences, and Frontiers of Computer Science.}
\end{IEEEbiography}

\vfill

\end{document}